\documentclass{article}
\pdfoutput=1

\usepackage[final, nonatbib]{neurips_2023}

\usepackage[utf8]{inputenc} 
\usepackage[T1]{fontenc}    
\usepackage{hyperref}       
\usepackage{url}            
\usepackage{booktabs}       
\usepackage{amsfonts}       
\usepackage{nicefrac}       
\usepackage{microtype}      
\usepackage{xcolor}         

\usepackage{xspace}
\usepackage{amsmath}
\usepackage{amssymb}
\usepackage{graphicx}
\usepackage{enumitem}
\usepackage{bm}
\hypersetup{colorlinks=true}
\usepackage{subcaption}
\usepackage{colortbl}
\usepackage{wrapfig}
\usepackage[font=small]{caption}

\usepackage[capitalize]{cleveref}
\crefname{section}{Sec.}{Secs.}
\Crefname{section}{Section}{Sections}
\Crefname{table}{Table}{Tables}
\crefname{table}{Tab.}{Tabs.}
\graphicspath{{imgs/}}
\setlist[itemize]{itemsep=0pt,topsep=0pt,partopsep=0pt,parsep=0pt,leftmargin=4mm}
\usepackage{array}
\newcolumntype{C}{>{$}c<{$}} 

\title{Differentiable Blocks World:\\Qualitative 3D Decomposition by Rendering Primitives}

\author{%
  Tom Monnier$^1$\quad Jake Austin$^2$\quad Angjoo Kanazawa$^2$\quad Alexei A. Efros$^2$\quad 
  Mathieu Aubry$^1$\\\\
  $^1$LIGM, Ecole des Ponts, Univ Gustave Eiffel \quad\quad\quad $^2$UC Berkeley\\
}

\makeatletter
\DeclareRobustCommand\onedot{\futurelet\@let@token\@onedot}
\def\@onedot{\ifx\@let@token.\else.\null\fi\xspace}
\def\eg{\emph{e.g}\onedot} 
\def\ie{\emph{i.e}\onedot}

\def\wrt{w.r.t\onedot}

\def\etal{\emph{et al}\onedot}
\makeatother

\definecolor{mygreen}{RGB}{0, 175, 0}
\definecolor{myblue}{RGB}{0, 102, 204}
\definecolor{lightblue}{RGB}{232, 244, 248}
\definecolor{myred}{RGB}{220, 60, 60}
\definecolor{myorange}{RGB}{255, 128, 0}
\definecolor{greenyellow}{RGB}{197,228,10}

\def\RR{{\rm I\!R}}

\def\bA{{\mathbb A}}  \def\bG{{\mathbb G}}   
\def\bB{{\mathbb B}}     
     
   \def\bP{{\mathbb P}}

\def\cC{{\mathcal C}}   \def\cO{{\mathcal O}}

  \def\cL{{\mathcal L}}

\newcommand{\img}{\mathbf{I}}
\newcommand{\txt}{\mathbf{T}}
\newcommand{\cam}{\mathbf{c}}
\newcommand{\rec}{\hat{\mathbf{I}}}
\newcommand{\bg}{\text{bg}}
\newcommand{\gr}{\text{gr}}
\newcommand{\pmesh}{\bP}
\newcommand{\bmesh}{\bB}
\newcommand{\fmesh}{\bG}

\newcommand{\point}{\mathbf{x}}

\newcommand{\shape}{\mathbf{e}}
\newcommand{\scale}{\mathbf{s}}
\newcommand{\rot}{\mathbf{r}}
\newcommand{\pose}{\mathbf{p}}
\newcommand{\trans}{\mathbf{t}}

\newcommand{\impsq}{\Psi}
\newcommand{\paramsq}{\Phi}
\newcommand{\occsqk}{\cO^{\text{3D}}_{k}}
\newcommand{\temp}{\tau}

\newcommand{\pointset}{\Omega}

\newcommand{\lrec}{\cL_{\text{render}}}
\newcommand{\lperc}{\cL_{\text{perc}}}
\newcommand{\lpix}{\cL_{\text{MSE}}}
\newcommand{\lpars}{\cL_{\text{parsi}}}
\newcommand{\ltv}{\cL_{\text{TV}}}
\newcommand{\ltvs}{\cL_{\text{tv}}}
\newcommand{\lover}{\cL_{\text{over}}}
\newcommand{\ltot}{\cL}

\newcommand{\wperc}{\lambda_{\text{perc}}}

\newcommand{\wpars}{\lambda_{\text{parsi}}}
\newcommand{\wtv}{\lambda_{\text{TV}}}
\newcommand{\wover}{\lambda_{\text{over}}}

\newcommand{\pixi}{\mathbf{u}}
\newcommand{\facej}{j}
\newcommand{\dist}{\Delta_\facej(\pixi)}
\newcommand{\agg}{\cC}

\newcommand{\occ}{\cO^{\text{2D}}}
\newcommand{\prob}{\mathbf{O}}
\newcommand{\col}{\mathbf{C}}

\DeclareMathOperator*{\sigmoid}{sigmoid}
\DeclareMathOperator*{\rotfn}{rot}

\begin{document}

\maketitle

\begin{abstract} \looseness=-1
  Given a set of calibrated images of a scene, we present an approach that produces a simple, 
  compact, and actionable 3D world representation by means of 3D primitives.  While many 
  approaches focus on recovering high-fidelity 3D scenes, we focus on parsing a scene into 
  \textit{mid-level} 3D representations made of a small set of textured primitives. Such 
  representations are interpretable, easy to manipulate and suited for physics-based 
  simulations.  Moreover, unlike existing primitive decomposition methods that rely on 
  3D input data, our approach operates directly on images through differentiable rendering.  
  Specifically, we model primitives as textured superquadric meshes and optimize their 
  parameters from scratch with an image rendering loss.  We highlight the importance of 
  modeling transparency for each primitive, which is critical for optimization and also 
  enables handling varying numbers of primitives. We show that the resulting textured 
  primitives faithfully reconstruct the input images and accurately model the visible 3D 
  points, while providing amodal shape completions of unseen object regions.  We compare our 
  approach to the state of the art on diverse scenes from DTU, and demonstrate its robustness 
  on real-life captures from BlendedMVS and Nerfstudio. We also showcase how our results can 
  be used to effortlessly edit a scene or perform physical simulations. Code and video 
  results are available at 
  \href{https://www.tmonnier.com/DBW}{\texttt{www.tmonnier.com/DBW}}.
\end{abstract}

\section{Introduction}

Recent multi-view modeling approaches, building on Neural Radiance 
Fields~\cite{mildenhall2020nerf}, capture scenes with astonishing accuracy by optimizing a 
dense occupancy and color model. However, they do not incorporate any notion of objects, 
they are not easily interpretable for a human user or a standard 3D modeling software, and 
they are not useful for physical understanding of the scene. In fact, even though these 
approaches can achieve a high-quality 3D reconstruction, the recovered content is nothing but 
a soup of colorful particles! In contrast, we propose an approach that recovers textured 
primitives, which are compact, actionable, and interpretable.

More concretely, our method takes as input a collection of calibrated images of a scene, and 
optimizes a set of primitive meshes parametrized by 
superquadrics~\cite{barr1981superquadrics} and their UV textures to minimize a rendering 
loss.  The approach we present is robust enough to work directly from a random 
initialization. One of its key components is the optimization of a transparency parameter for 
each primitive, which helps in dealing with occlusions as well as handling
varying number of primitives.
This notably requires adapting standard differentiable 
renderers to deal with transparency. We also show the benefits of using a perceptual loss, a 
total variation regularization on the textures and a parsimony loss favoring the use of a
minimal number of primitives.

\begin{figure}[t]
  \centering
  \begin{subfigure}{0.3\textwidth}
    \centering
    \addtolength{\tabcolsep}{-5pt}
    \renewcommand{\arraystretch}{0.8}
    \begin{tabular}{@{}ccc@{}}
      \small Input &\small Output\\
      \includegraphics[trim=400 100 400 100, clip, height=2.2cm]{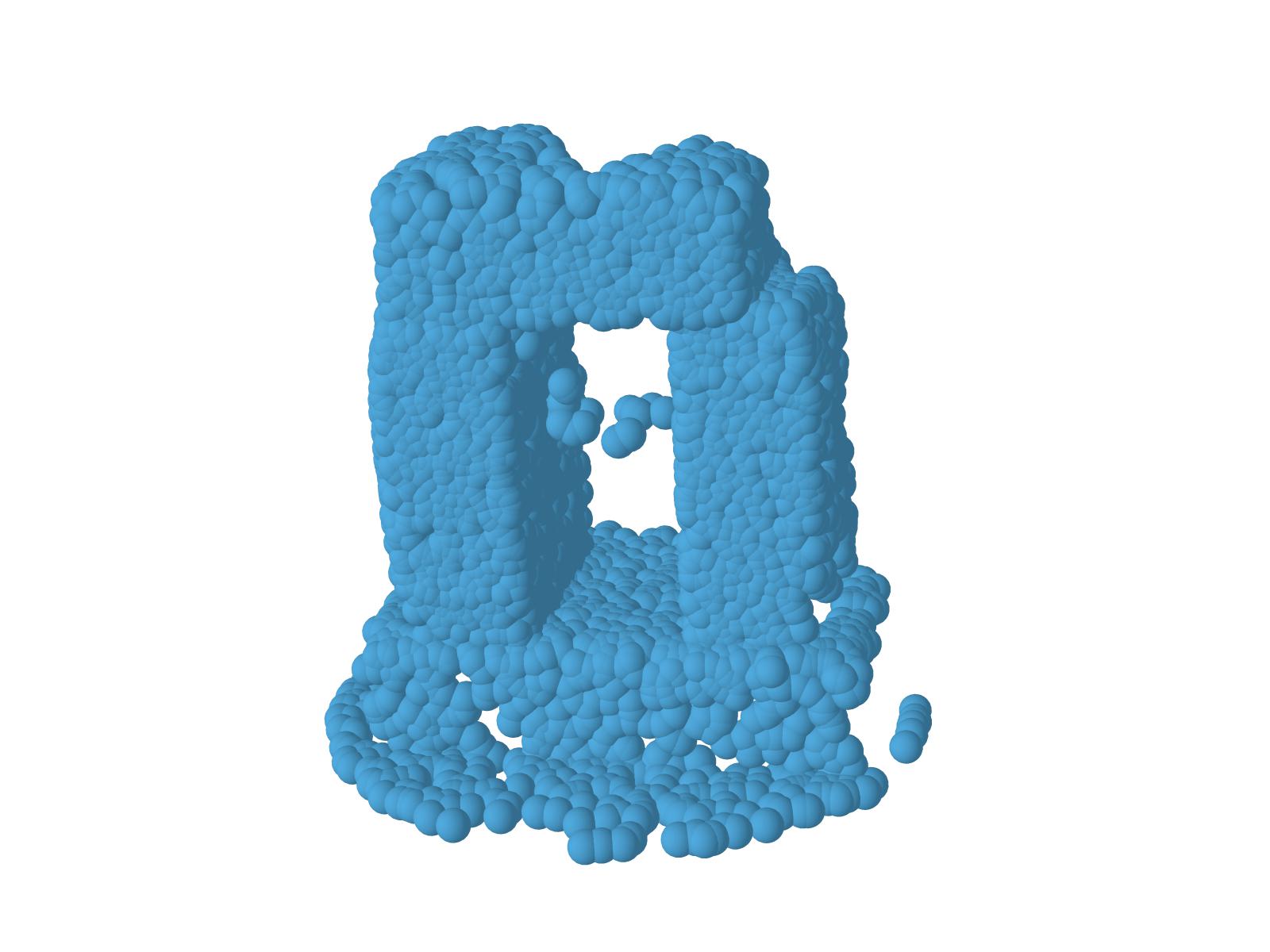} &
      \includegraphics[trim=400 100 450 100, clip, height=2.2cm]{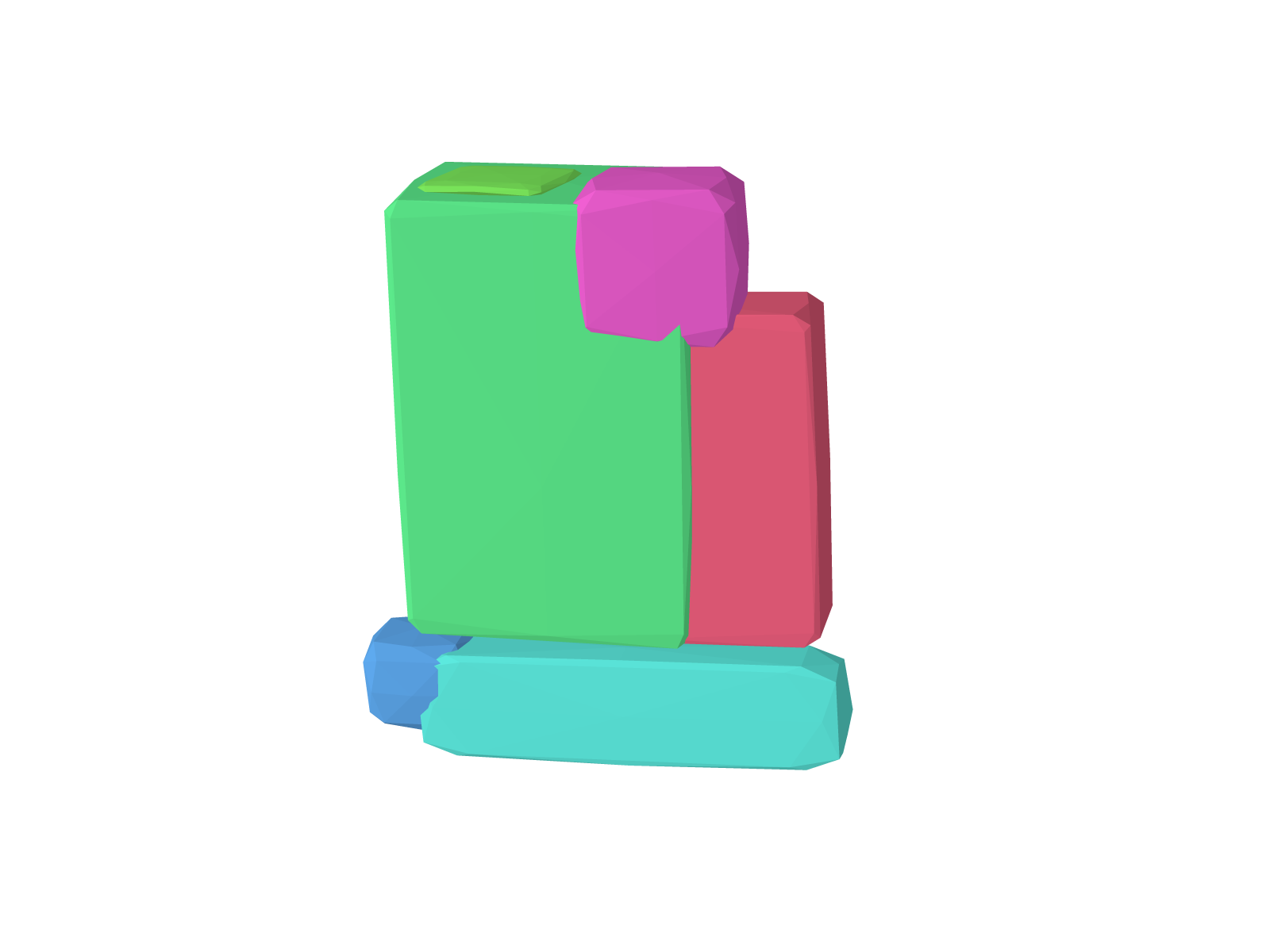}\\
    \end{tabular}%
    \vspace{-.4em}
    \caption{Prior works, \eg,~\cite{liu2022robust}}
    \label{fig:teaser_prior}
  \end{subfigure}\hfill
  {\vrule width 0.5pt}\hfill
  \begin{subfigure}{0.65\textwidth}
    \centering
    \addtolength{\tabcolsep}{-4pt}
    \renewcommand{\arraystretch}{0.8}
    \begin{tabular}{@{}ccc@{}}
      \small Input & \multicolumn{2}{c}{\small Optimized textured 3D primitives} \\ 
      \includegraphics[trim=0 100 0 100, clip, height=2.2cm]{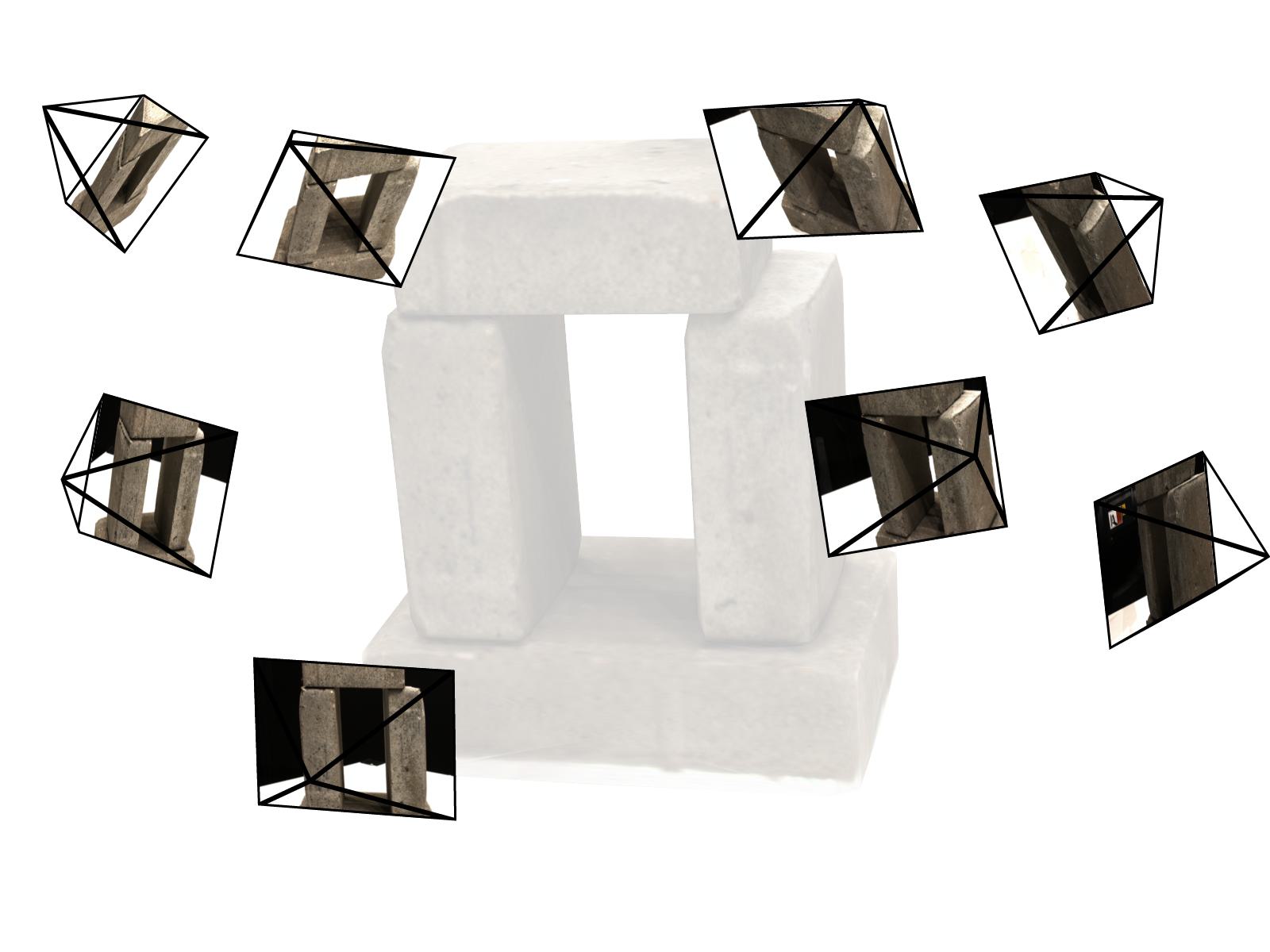} &
      \includegraphics[trim=0 100 0 100, clip, height=2.2cm]{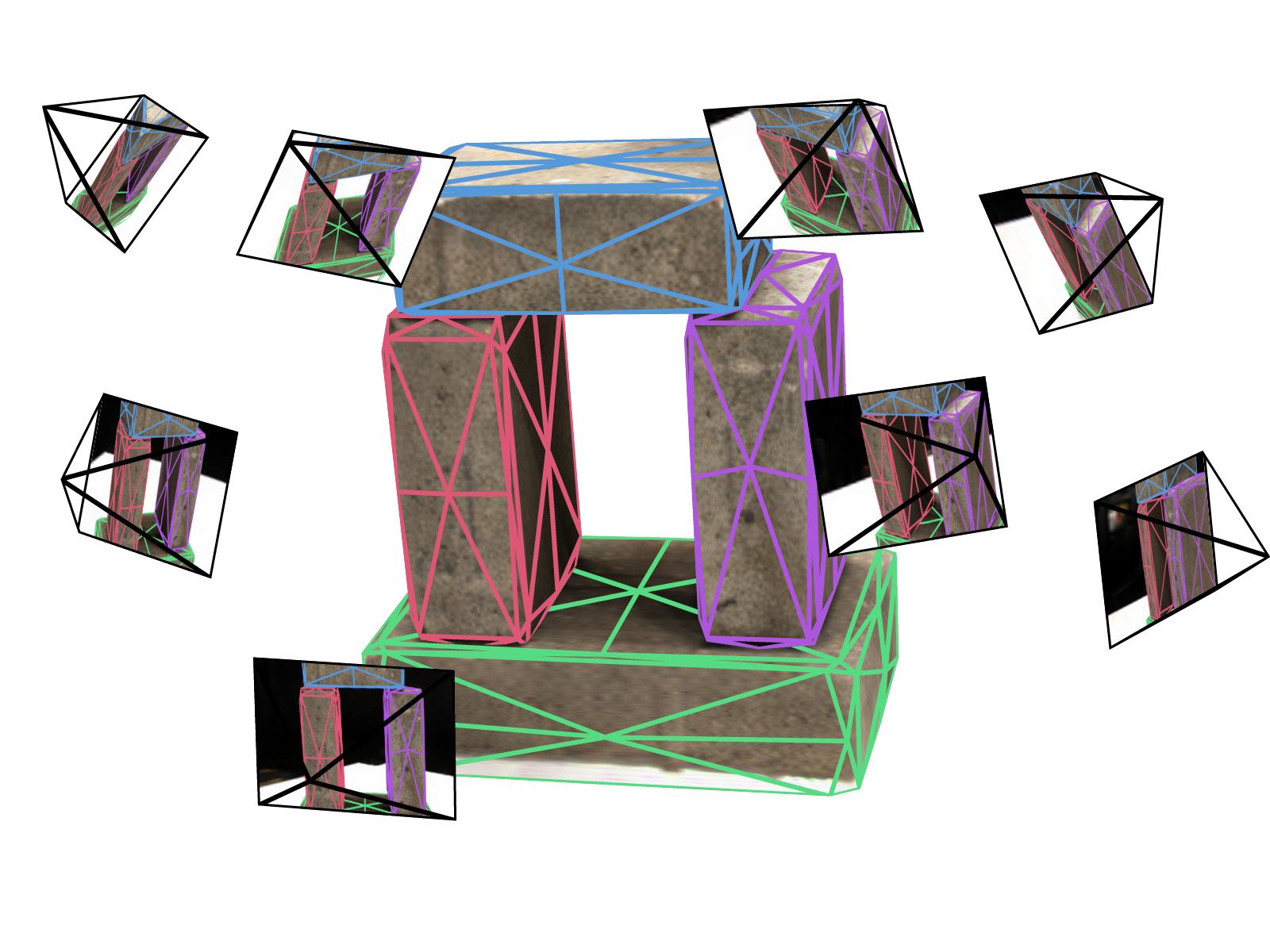} &
      \includegraphics[trim=400 100 450 100, clip, height=2.2cm]{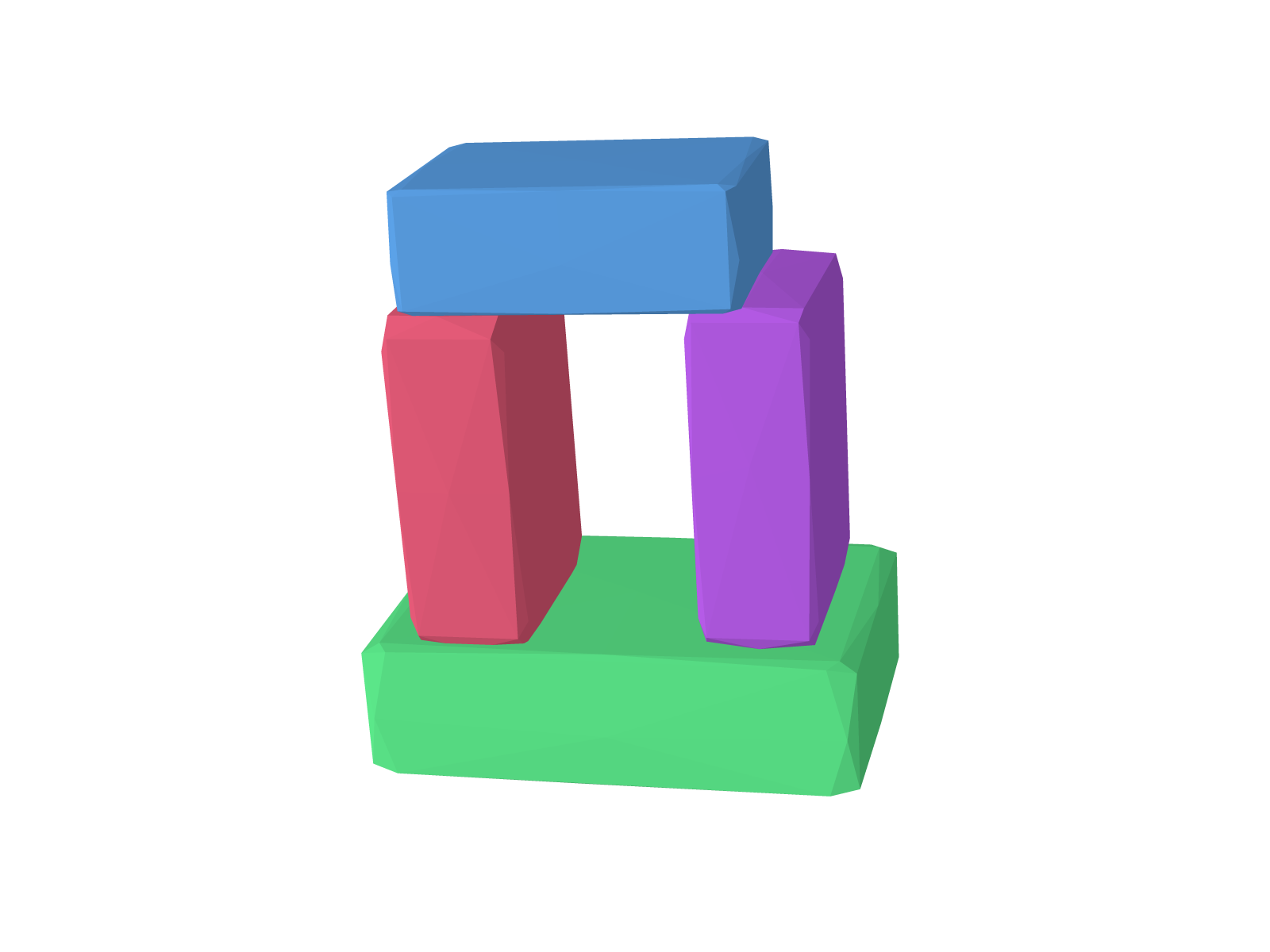}\\
    \end{tabular}%
    \vspace{-.4em}
    \caption{\textbf{Our work}}
    \label{fig:teaser_ours}
  \end{subfigure}
  
  \vspace{.5em}

  \begin{subfigure}{\textwidth}
    \centering
    \includegraphics[trim=200 0 100 50, clip, width=0.24\linewidth]{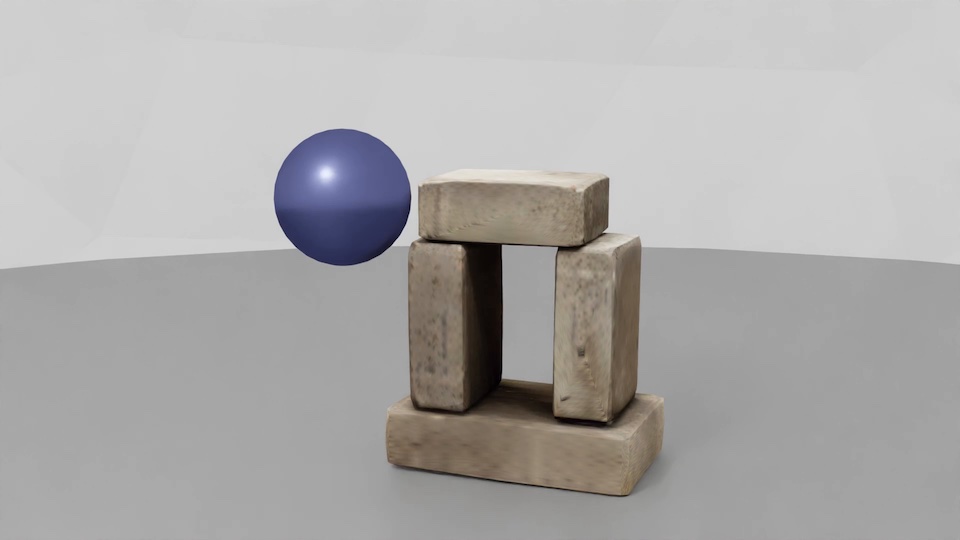}
    \includegraphics[trim=200 0 100 50, clip, width=0.24\linewidth]{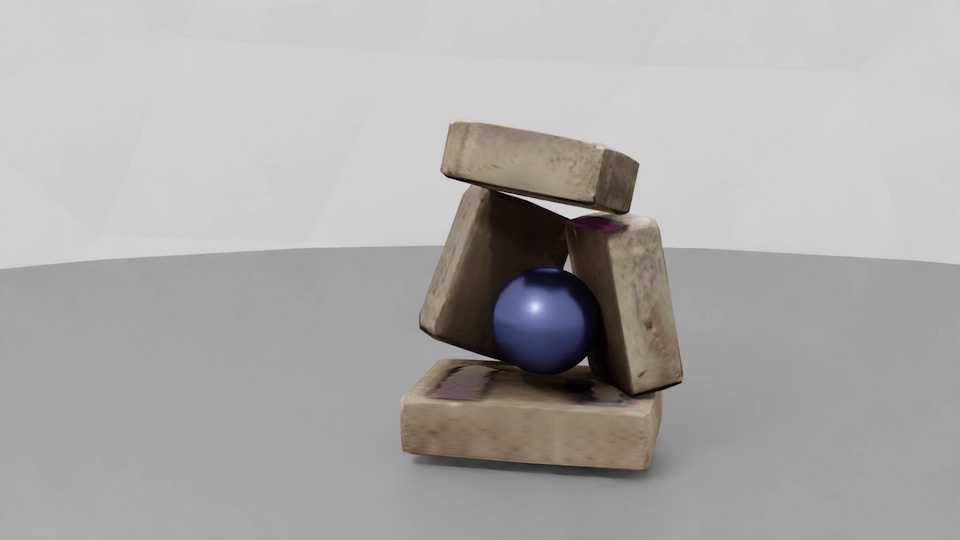}
    \includegraphics[trim=200 0 100 50, clip, width=0.24\linewidth]{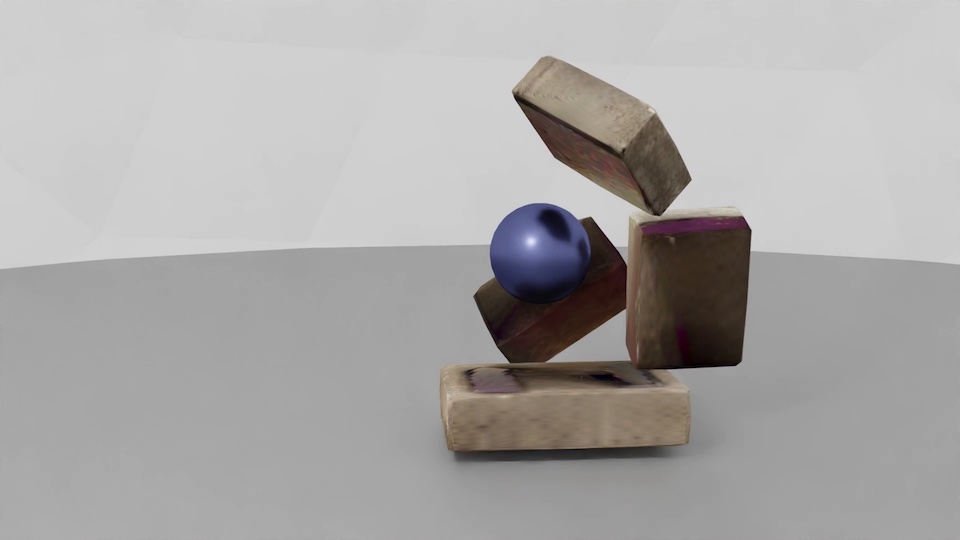}
    \includegraphics[trim=200 0 100 50, clip, width=0.24\linewidth]{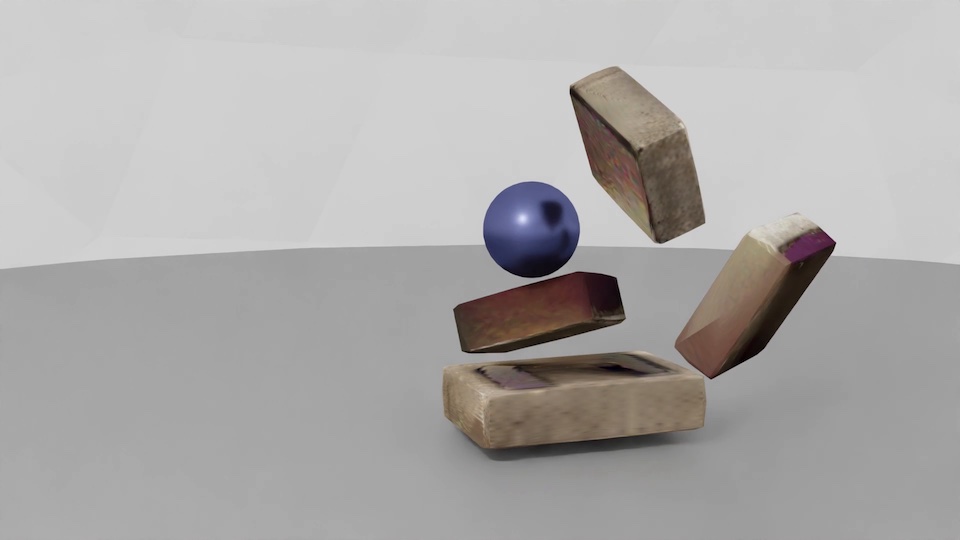}
    \caption{Application: physics-based simulations}
    \label{fig:teaser_sim}
  \end{subfigure}
  \vspace{-.2in}
  \caption{\textbf{Differentiable Blocks World.}  \textbf{(a)} Prior works fit primitives to 
    point clouds and typically fail for real data where ground-truth point clouds are 
    extremely noisy and incomplete. \textbf{(b)} We propose using calibrated multi-view 
    images instead and simultaneously tackle 3D decomposition and 3D reconstruction by 
  rendering learnable textured primitives in a differentiable manner. \textbf{(c)} Such a 
textured decomposition is highly compact and user-friendly: it enables us to do physics-based 
simulations, \eg, throwing a ball at the discovered primitives.}
  \label{fig:teaser}
\end{figure}

\looseness=-1 Our scene representation harks back to the classical Blocks World 
ideas~\cite{roberts1963machine}. An important difference is that the Blocks World-inspired 
approaches are typically bottom-up, leveraging low-level image features, such as 
edges~\cite{roberts1963machine},
super-pixels~\cite{gupta2010blocks}, or more recently learned 
features~\cite{xiao2012localizing, kluger2021cuboids}, to infer 3D blocks. In 
contrast, we perform a direct top-down optimization of 3D primitives and texture using a 
rendering loss, starting from a random initialization in the spirit of analysis-by-synthesis. 
Unlike related works that fit
primitives to 3D point clouds~\cite{binford1971visual, barr1981superquadrics, 
tulsiani2017learning,  li2019supervised, wu2022primitivebased, liu2022robust, 
loiseau2023learnable} (\Cref{fig:teaser_prior}),
our approach, dubbed \emph{Differentiable Blocks World} (or DBW), does not require any 3D 
reconstruction  \emph{a priori} but instead operates directly on a set of calibrated input 
images, leveraging photometric consistency across different views (\Cref{fig:teaser_ours}).  
This makes our approach more robust since methods based on 3D are very sensitive to noise in 
the reconstructions and have difficulties dealing with incomplete objects. Our setting is 
similar to existing NeRF-like approaches, but our model is able to recover a significantly 
more interpretable and parsimonious representation.  In particular, such an interpretable 
decomposition allows us to easily play with the discovered scene, \eg, by performing 
physics-based simulations (\Cref{fig:teaser_sim}). Code and video results are available on 
our project webpage: \href{https://www.tmonnier.com/DBW}{\texttt{www.tmonnier.com/DBW}}.

\section{Related Work}

\paragraph{Scene decomposition into 3D primitives.}

The goal of understanding a scene by decomposing it into a set of geometric primitives can be 
traced back to the very fist computer vision thesis by Larry Roberts on \textit{Blocks 
World}~\cite{roberts1963machine} in
1963. In it, Roberts shows a complete scene understanding system for a simple closed world of 
textureless polyhedral shapes by using a generic library of polyhedral block components.  In 
the 1970s, Binford proposes the use of Generalized Cylinders as general 
primitives~\cite{binford1971visual}, later refined by Biederman into the 
recognition-by-components theory~\cite{biederman1987recognition}.  But applying these ideas 
to real-world image data has proved
rather difficult. 

A large family of methods does not consider images at all, instead focusing on finding 
primitives
in 3D data.  Building upon the classical idea of RANSAC~\cite{fischler1981random}, works 
like~\cite{bolles1981ransacbased, chaperon2001extracting, schnabel2007efficient, 
schnabel2009completion, li2011globfit, nan2017polyfit, ramamonjisoa2022monteboxfinder} accurately
extract various primitive shapes (\eg, planes, spheres and cylinders 
for~\cite{schnabel2007efficient, schnabel2009completion, li2011globfit}) from a point cloud.
In particular, MonteBoxFinder~\cite{ramamonjisoa2022monteboxfinder} is a recent RANSAC-based 
system that robustly extracts cuboids from noisy point clouds by selecting the best
proposals through Monte Carlo Tree Search. To avoid the need for RANSAC hyperparameter tuning
while retaining robustness, Liu~\etal~\cite{liu2022robust} introduce a probabilistic 
framework dubbed EMS that recovers superquadrics~\cite{barr1981superquadrics}.
Other methods
instead leverage neural learning advances to robustly predict primitive decomposition from a 
collection of shapes (\eg, ShapeNet~\cite{chang2015shapenet}), in the form of 
cuboids~\cite{tulsiani2017learning}, superquadrics~\cite{paschalidou2019superquadrics, 
paschalidou2020learning, wu2022primitivebased}, shapes from a small 
dictionary~\cite{li2019supervised, le2021cpfn} or learnable prototypical 
shapes~\cite{deprelle2019learning, paschalidou2021neural, loiseau2023learnable}. However, 
they are typically limited to shapes of known categories and require perfect 3D data. More 
generally, the decomposition results of all 3D-based methods highly depend on the quality of 
the 3D input, which is always noisy and incomplete for real scenes.  For a complete survey of 
3D decomposition methods, we refer the reader to~\cite{kaiser2019survey}.

More recently, there has been a renewed effort to fit 3D primitives to various image 
representations,
such as depth maps, segmentation predictions or low-level image features. Depth-based 
approaches~\cite{jiang2013linear, fouhey2013datadriven, lin2013holistic, geiger2015joint, 
kluger2021cuboids} naturally associate a 3D point cloud to each image which is then used for 
primitive fitting. However, the resulting point cloud is highly incomplete,
ambiguous and sometimes inaccurately predicted, thus limiting the decomposition quality.  
Building upon the single-image scene layout estimation~\cite{hoiem2005geometric, 
hoiem2007recovering}, works like~\cite{gupta2010blocks, lee2010estimating} compute cuboids 
that best match the predicted surface orientations.
Finally, Façade~\cite{debevec1996modeling}, the classic image-based rendering work, leverages 
user annotations across multiple images with known camera viewpoints to render a scene with 
textured 3D primitives.
In this work, we \textit{do not} rely on 3D, depth, segmentation, low-level features, or user 
annotations to compute the 3D decomposition. Instead, taking inspiration from 
Façade~\cite{debevec1996modeling} and recent multi-view modeling 
advances~\cite{tulsiani2017multiview, niemeyer2020differentiable, mildenhall2020nerf}, 
our approach only requires calibrated views of the scene 
and directly optimizes textured primitives through photometric consistency in an
end-to-end fashion. That is, we solve the 3D decomposition and multi-view stereo problems 
simultaneously.

\paragraph{Multi-view stereo.} Our work can be seen as an end-to-end primitive-based approach 
to multi-view stereo (MVS), whose goal is to output a 3D reconstruction from multiple images 
taken from known camera viewpoints. We refer the reader to~\cite{hartley2003multiple, 
furukawa2015multiview} for an exhaustive review of classical methods. Recent MVS works can be 
broadly split into two groups.

Modular multi-step approaches typically rely on several processing steps to extract the 
final geometry from the images. Most methods~\cite{zheng2014patchmatch, 
galliani2015massively, yao2018mvsnet, yao2019recurrent, zhang2020visibilityaware, 
gu2020cascade, sinha2020deltas}, including the widely used 
COLMAP~\cite{schonberger2016pixelwise}, first estimate depth maps for each image (through 
keypoint matching~\cite{schonberger2016pixelwise} or neural network 
predictions~\cite{yao2018mvsnet, yao2019recurrent, zhang2020visibilityaware, gu2020cascade, 
sinha2020deltas}), then apply a depth fusion step to generate a textured point cloud.  
Finally, a mesh can be obtained with a meshing algorithm~\cite{kazhdan2013screened, 
labatut2007efficient}. Other multi-step approaches directly rely on point 
clouds~\cite{furukawa2007accurate, labatut2007efficient} or voxel 
grids~\cite{seitz1997photorealistic, kutulakos1999theory, ji2017surfacenet, murez2020atlas}.  
Note that, although works like~\cite{ji2017surfacenet, murez2020atlas} leverage end-to-end 
trainable networks to regress the geometry, we consider them as multi-step methods as they 
still rely on a training phase requiring 3D supervision before being applied to unknown sets 
of multi-view images. Extracting geometry through multiple steps involves careful tuning of 
each stage, thus increasing the pipeline complexity.

End-to-end approaches directly optimize a 3D scene representation using photometric 
consistency across different views along with other constraints in an
optimization framework. Recent 
methods use neural networks to implicitly represent the 3D scene, in the form of occupancy 
fields~\cite{niemeyer2020differentiable}, signed distance functions~\cite{yariv2020multiview} 
or radiance fields, as introduced in NeRF~\cite{mildenhall2020nerf}. Several works incorporate
surface constraints in neural volumetric rendering to further improve the scene 
geometry~\cite{oechsle2021unisurf, 
yariv2021volume, wang2021neus, darmon2022improving, fu2022geoneus},
with a quality approaching that of traditional MVS methods. Other 
methods~\cite{gao2020learning, zhang2021ners, goel2022differentiable, munkberg2022extracting} 
instead propose to leverage recent advances in mesh-based differentiable 
rendering~\cite{loper2014opendr, katoNeural3DMesh2018, liuSoftRasterizerDifferentiable2019, 
chenLearningPredict3D2019, raviAccelerating3DDeep2020, laine2020modular} to explicitly 
optimize textured meshes. Compared to implicit 3D representations, meshes are highly 
interpretable and are straightforward to use in computer graphic pipelines, thus 
enabling effortless scene editing and simulation~\cite{munkberg2022extracting}.
However, all the above approaches represent the scene as a single mesh, making it ill-suited 
for manipulation and editing.  We instead propose to discover the primitives that make up
the scene, resulting in an interpretable and actionable representation. A concurrent work 
PartNeRF~\cite{tertikas2023partnerf} introduces parts in NeRFs. However, only 
synthetic scenes with a single object are studied and the discovered parts mostly correspond 
to regions in the 3D space rather than interpretable geometric primitives.

\section{Differentiable Blocks World}

Given a set of $N$ views $\img_{1:N}$ of a scene associated with camera poses $\cam_{1:N}$, 
our goal is to decompose the 3D scene into geometric primitives that best explain the images.  
We explicitly model the scene as a set of transparent superquadric meshes, whose parameters, 
texture and number are optimized to maximize photoconsistency through differentiable 
rendering.  Note that compared to recent advances in neural volumetric 
representations~\cite{niemeyer2020differentiable, mildenhall2020nerf, yu2021pixelnerf}, we 
\textit{do not} use any neural network and directly optimize meshes, which are 
straightforward to use in computer graphic pipelines.

\textbf{Notations.} We use bold lowercase for vectors (\eg, $\mathbf{a}$), bold uppercase for 
images (\eg, $\mathbf{A}$), double-struck uppercase for meshes (\eg, $\bA$)
and write $a_{1:N}$ the ordered set $\{a_1,\ldots,a_n\}$.

\subsection{Parametrizing a World of Blocks}\label{sec:scene}

\begin{figure}
  \centering
  \begin{subfigure}{0.9\textwidth}
    \centering
    \includegraphics[width=\linewidth]{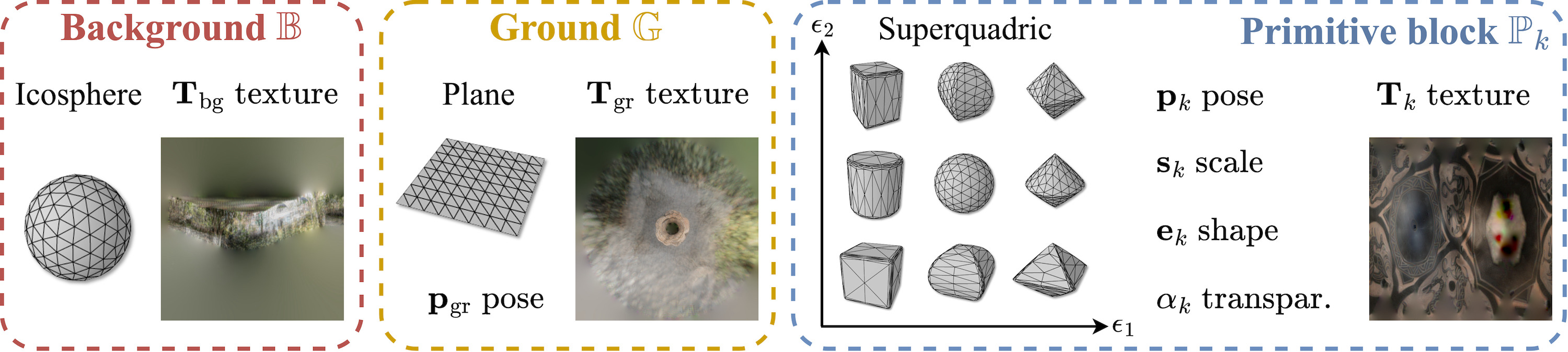}
  \end{subfigure}

  \vspace{.9em}

  \begin{subfigure}{\textwidth}
    \centering
    \addtolength{\tabcolsep}{-5pt}
    \renewcommand{\arraystretch}{0.8}
\resizebox{\linewidth}{!}{%
    \begin{tabular}{@{}cccccccc@{}}
      \multicolumn{2}{c}{\small Input (subset)} &\small Init &\small Iter 200 &\small Iter 1k 
      &\small Iter 10k &\small Final &\small Output \\
      
      \includegraphics[width=0.0403\linewidth]{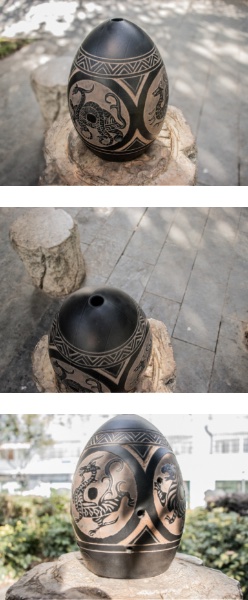} & 
      \includegraphics[width=0.13\linewidth]{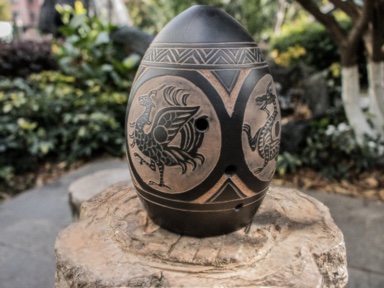} & 
      \includegraphics[width=0.13\linewidth]{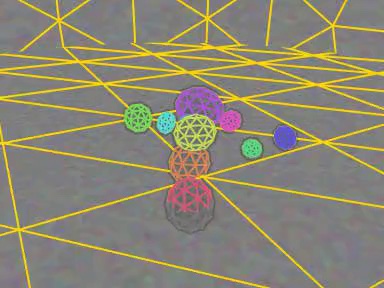} & 
      \includegraphics[width=0.13\linewidth]{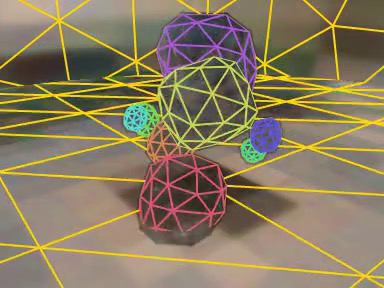} & 
      \includegraphics[width=0.13\linewidth]{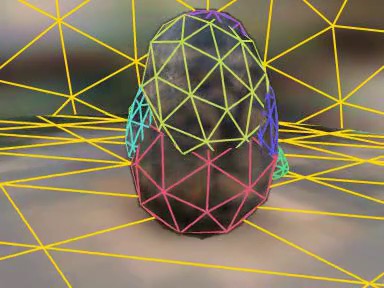} & 
      \includegraphics[width=0.13\linewidth]{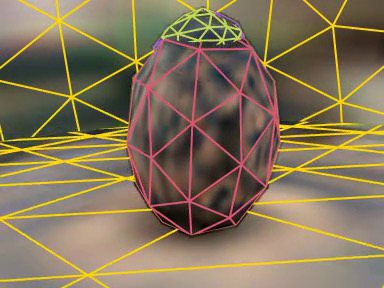} & 
      \includegraphics[width=0.13\linewidth]{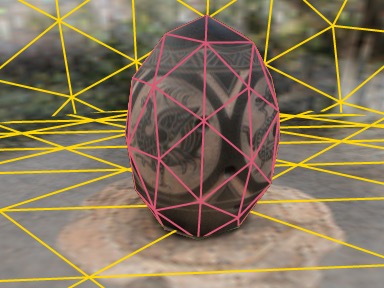} &
      \includegraphics[width=0.13\linewidth]{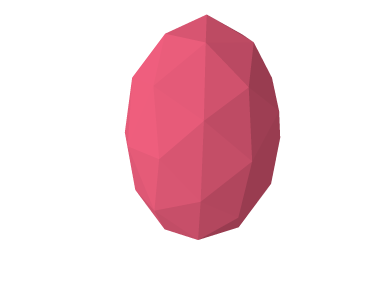}\\

      \includegraphics[width=0.0403\linewidth]{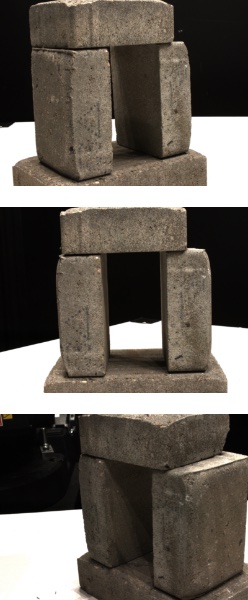} & 
      \includegraphics[width=0.13\linewidth]{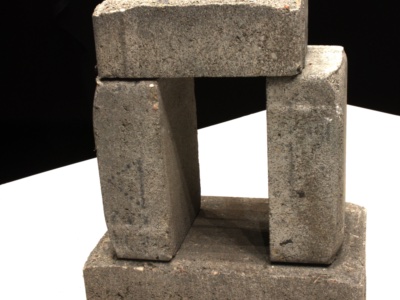} & 
      \includegraphics[width=0.13\linewidth]{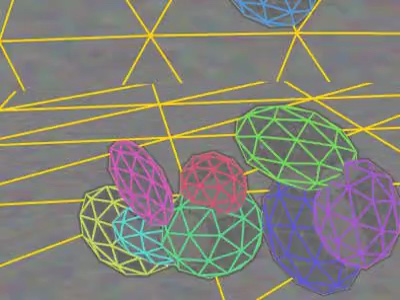} & 
      \includegraphics[width=0.13\linewidth]{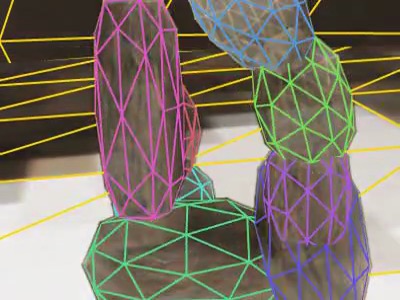} & 
      \includegraphics[width=0.13\linewidth]{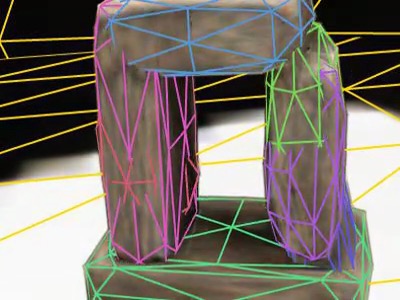} & 
      \includegraphics[width=0.13\linewidth]{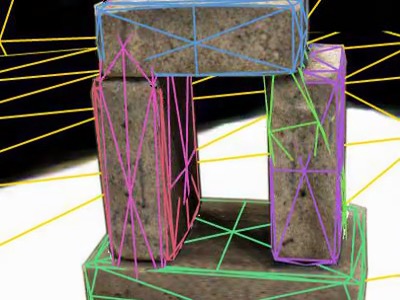} & 
      \includegraphics[width=0.13\linewidth]{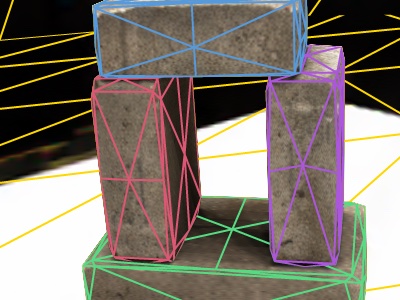} &
      \includegraphics[width=0.13\linewidth]{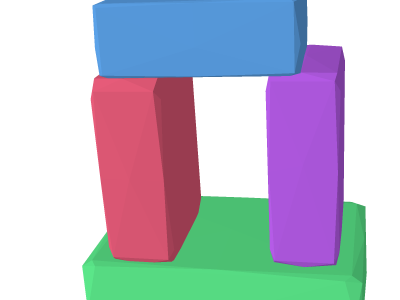} \\

      \includegraphics[width=0.0403\linewidth]{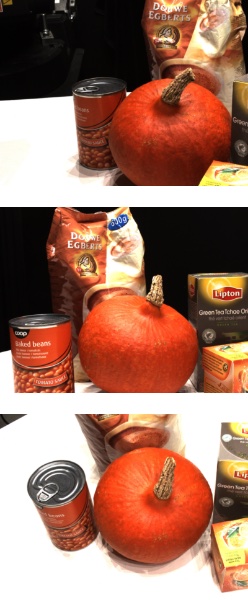} & 
      \includegraphics[width=0.13\linewidth]{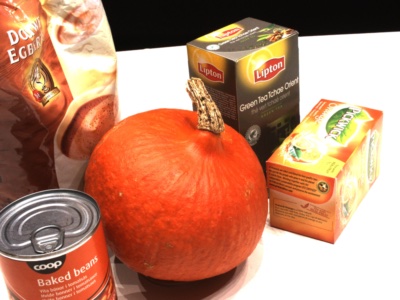} & 
      \includegraphics[width=0.13\linewidth]{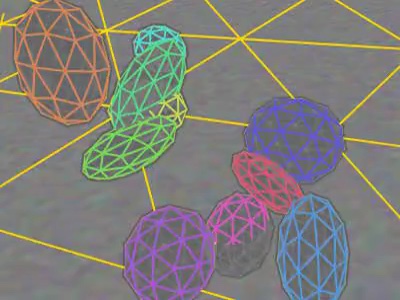} & 
      \includegraphics[width=0.13\linewidth]{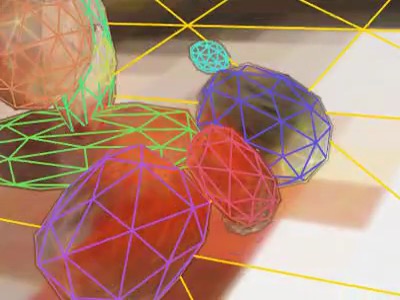} & 
      \includegraphics[width=0.13\linewidth]{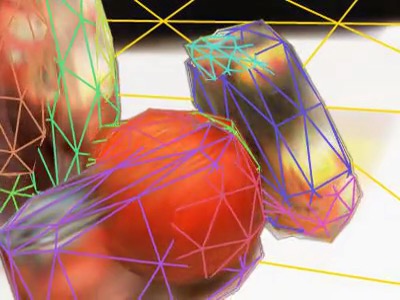} & 
      \includegraphics[width=0.13\linewidth]{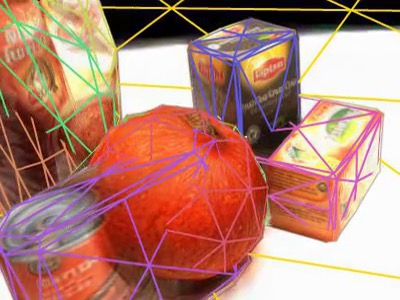} & 
      \includegraphics[width=0.13\linewidth]{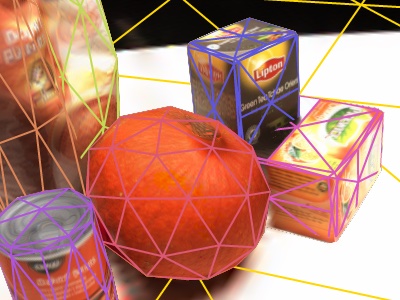} &
      \includegraphics[width=0.13\linewidth]{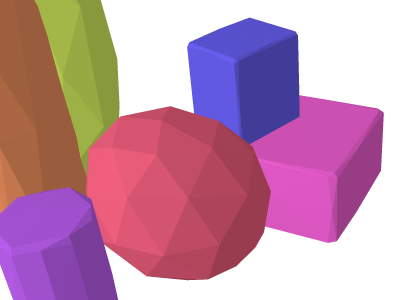} \\

    \end{tabular}
    }
  \end{subfigure}
  \vspace{-.2em}
  \caption{\textbf{Overview}. \textbf{(top)} We model the world as an explicit set of learnable 
  textured meshes that are assembled together in the 3D space. \textbf{(bottom)} Starting from a random
  initialization, we optimize such a representation through differentiable rendering by photometric
  consistency across the different views.}
  \label{fig:overview}
\end{figure}

We propose to represent the world scene as an explicit set of textured meshes positioned in 
the 3D space.~\Cref{fig:overview} summarizes our modeling and the parameters updated (top)
during the optimization (bottom). Specifically, we model each scene as a union of primitive 
meshes: (i) an icosphere $\bmesh$ modeling a background dome and centered on the scene, (ii) 
a plane $\fmesh$ modeling the ground, and (iii) $K$ primitive blocks $\pmesh_{1:K}$ in the 
form of superquadric meshes, where $K$ is fixed and refers to a maximum number of blocks.  
Unless mentioned otherwise, we arbitrarly use $K=10$. We write the resulting scene mesh 
$\bmesh \cup \fmesh \cup \pmesh_1 \cup \ldots \cup \pmesh_K$.

The goal of the background dome is to model things far from the cameras that can be well 
approximated with a planar surface at infinity. In practice, we consider an icosphere with a 
fixed location and a fixed scale that is much greater than the scene scale.  On the contrary, 
the goal of the planar ground and the blocks is to model the scene close to the cameras.  We 
thus introduce rigid transformations modeling locations that will be updated during the 
optimization. Specifically, we use the 6D rotation parametrization 
of~\cite{zhou2019continuity} and associate to each block $k$ a pose $\pose_k = \{\rot_k, 
\trans_k\}\in\RR^9$ such that every point of the block $\point \in \RR^3$ is transformed into 
world space by $\point_{\textrm{world}} = \rotfn(\rot_k) \point + \trans_k$, where $\trans_k 
\in \RR^3$, $\rot_k \in \RR^6$ and $\rotfn$ maps a 6D vector to a rotation 
matrix~\cite{zhou2019continuity}.  Similarly, we associate a rigid transformation $\pose_\gr 
= \{\rot_{\gr}, \trans_{\gr}\}$ to the ground plane. We next describe how we model variable 
number of blocks via transparency values and the parametrization of blocks' shape and texture.

\noindent\textbf{Block existence through transparency.} Modeling a variable number of 
primitives is a difficult task as it involves optimizing over a discrete random variable. 
Recent works tackle the problem using reinforcement learning~\cite{tulsiani2017learning},
probabilistic approximations~\cite{paschalidou2019superquadrics} or greedy 
algorithms~\cite{monnierUnsupervisedLayeredImage2021}, which often yield complex optimization 
strategies. In this work, we instead propose to handle variable number of primitive blocks by 
modeling meshes that are \textit{transparent}. Specifically, we associate to each block $k$ a 
learnable transparency value $\alpha_k$, parametrized with a sigmoid, that can be pushed 
towards zero to change the effective number of blocks. Such transparencies are not only
used in our rendering process to softly model the blocks existence and occlusions
(\Cref{sec:render}), but also in regularization terms during our optimization, \eg, to 
encourage parsimony in the number of blocks used (\Cref{sec:optim}).

\noindent\textbf{Superquadric block shape.} We model blocks with
superquadric meshes. Introduced by
Barr in 1981~\cite{barr1981superquadrics} and revived recently 
by~\cite{paschalidou2019superquadrics},
superquadrics define a family of parametric surfaces that exhibits a strong expressiveness 
with a small number of continuous parameters, thus making a good candidate for primitive 
fitting by gradient descent.
More concretely, we derive a superquadric mesh from a unit icosphere. For each vertex of the 
icosphere, its spherical coordinates $\eta \in [-\frac{\pi}{2}, \frac{\pi}{2}]$ and $\omega 
\in [-\pi, \pi]$ are mapped to the superquadric surface through the
parametric equation~\cite{barr1981superquadrics}:%
\begin{equation}
  \paramsq(\eta, \omega) =
  \begin{bmatrix}
    s_{1}\cos^{\epsilon_{1}}\eta \cos^{\epsilon_{2}}\omega \\
    s_{2}\sin^{\epsilon_{1}}\eta\\
    s_{3}\cos^{\epsilon_{1}}\eta \sin^{\epsilon_{2}}\omega \\
  \end{bmatrix},
  \label{eq:parametric_eq}
\end{equation}%
where $\scale = \{s_1, s_2, s_3\}\in\RR^3$ represents an anisoptropic scaling
and $\shape = \{\epsilon_{1}, \epsilon_{2}\}\in\RR^2$ defines the shape of the superquadric.  
Both $\scale$ and $\shape$ are updated during the optimization process. Note that by design, 
each vertex of the icosphere is mapped continuously to a vertex of the superquadric mesh, so 
the icosphere connectivity - and thus the icosphere faces - is transferred to the 
superquadric mesh.

\noindent\textbf{Texturing model.} We use texture mapping to model scene appearance.  
Concretely, we optimize $K + 2$ texture images $\{\txt_{\bg}, \txt_{\gr}, \txt_{1:K}\}$ which 
are UV-mapped onto each mesh triangle using pre-defined UV mappings. Textures for the 
background and the ground are trivially obtained using respectively spherical coordinates of 
the icosphere and a simple plane projection. For a given block $k$, each vertex of the 
superquadric mesh is associated to a vertex of the icosphere. Therefore, we can map the 
texture image $\txt_k$ onto the superquadric by first mapping it to the icosphere using a 
fixed UV map computed with spherical coordinates, then mapping the icosphere triangles to the 
superquadric ones (see supplementary material for details).


\subsection{Differentiable Rendering}\label{sec:render}

In order to optimize our scene parameters to best explain the images, we propose to leverage 
recent mesh-based differentiable renderers~\cite{liuSoftRasterizerDifferentiable2019, 
chenLearningPredict3D2019, raviAccelerating3DDeep2020}. Similar to them, our differentiable 
rendering corresponds to the soft rasterization of the mesh faces followed by a blending 
function. In contrast to existing mesh-based differentiable renderers, we introduce the 
ability to account for transparency. Intuitively, our differentiable rendering can be 
interpreted as an alpha compositing of the transparent colored faces of the mesh. In the 
following, we write pixel-wise multiplication with $\odot$ and the division of image-sized 
tensors corresponds to pixel-wise division.

\noindent\textbf{Soft rasterization.} 
Given a 2D pixel location $\pixi$, we model the influence of 
the face $\facej$ projected onto the image plane with the 2D occupancy function 
of~\cite{chenLearningPredict3D2019} that we modify to incorporate the transparency 
value $\alpha_{k_j}$ associated to this face. Specifically, we write the occupancy function 
as:%
\begin{equation}
  \occ_\facej(\pixi) = \alpha_{k_j} \exp\Big(\min\Big(\frac{\dist}{\sigma},\;0\Big)\Big)~,
\end{equation}%
where $\sigma$ is a scalar hyperparameter modeling the extent of the soft mask of the face
and $\dist$ is the signed Euclidean distance between pixel $\pixi$ and projected face 
$\facej$, such that $\dist < 0$ if pixel $\pixi$ is outside face $\facej$ and $\dist \geq 0$ 
otherwise.  We consider the faces belonging to the background and the ground to be opaque, 
\ie, use a transparency of 1 for all their faces in the occupancy function.

\noindent\textbf{Blending through alpha compositing.} For each pixel, we find all projected 
faces with an occupancy greater than a small threshold at this pixel location, and sort them 
by increasing depth. Denoting by $L$ the maximum number of faces per pixel, we build 
image-sized tensors for occupancy $\prob_\ell$ and color $\col_\ell$ by associating to each 
pixel the $\ell$-th intersecting face attributes. The color is obtained through barycentric 
coordinates, using clipped barycentric coordinates for locations outside the face. Different 
to most differentiable renderers and as
advocated by~\cite{monnierShareThyNeighbors2022}, we directly interpret these tensors as an 
ordered set of RGBA image layers and blend them through traditional alpha
compositing~\cite{porterCompositingDigitalImages1984}:%
\begin{equation}
  \agg(\prob_{1:L}, \col_{1:L}) = \sum_{\ell = 1}^L \Big(\prod_{p<\ell}^L(1 - \prob_p)\Big) 
  \odot \prob_\ell \odot \col_\ell~.
\end{equation}%
We found this simple alpha composition to behave better during optimization than the original 
blending function used in~\cite{liuSoftRasterizerDifferentiable2019, 
chenLearningPredict3D2019, raviAccelerating3DDeep2020}. This is notably in line with recent 
advances in differentiable rendering like NeRF~\cite{mildenhall2020nerf} which can be 
interpreted as alpha compositing points along the rays.


\subsection{Optimizing a Differentiable Blocks World}\label{sec:optim}

We optimize our scene parameters by minimizing a rendering loss across batches of images 
using gradient descent. Specifically, for each image $\img$, we build the scene mesh as 
described in~\Cref{sec:scene} and use the associated camera pose to render an image 
$\rec$ using the rendering process detailed in~\Cref{sec:render}. We optimize an objective
function defined as:%
\begin{equation}
  \ltot = \lrec + \wpars \lpars + \wtv \ltv + \wover \lover~,
\end{equation}
where $\lrec$ is a rendering loss between $\img$ and $\rec$, $\wpars, \wtv, \wover$ are
scalar hyperparameters and $\lpars, \ltv, \lover$ are regularization terms respectively 
encouraging parsimony in the use of primitives, favoring smoothness in the texture maps and
penalizing the overlap between primitives. Our rendering loss is composed of a pixel-wise MSE
loss $\lpix$ and a perceptual LPIPS
loss~\cite{zhangUnreasonableEffectivenessDeep2018} $\lperc$ such that $\lrec = \lpix + \wperc 
\lperc$. In all experiments, we use $\wpars = 0.01, \wperc =
\wtv = 0.1$ and $\wover = 1$.~\Cref{fig:overview} (bottom) shows the evolution of our 
renderings throughout the optimization.

\noindent\textbf{Encouraging parsimony and texture smoothness.} We found that regularization 
terms were critical to obtain meaningful results.  In particular, the raw model typically 
uses the maximum number of blocks available to reconstruct the scene, thus over-decomposing 
the scene.  To adapt the number of blocks per scene and encourage parsimony, we use the 
transparency values as a proxy for the number of blocks used and penalize the loss by $\lpars 
= \sum_k \nicefrac{\sqrt{\alpha_k}}{K}$.  We also use a total variation (TV) 
penalization~\cite{rudinTotalVariation1994} on the texture maps to encourage uniform 
textures. Given a texture map $\txt$ of size $U \times V$ and denoting by $\txt[u, v] \in 
\RR^3$ the RGB values of the pixel at location $(u, v)$, we define:%
\begin{equation}
  \ltvs(\txt) = \frac{1}{UV}\sum_{u, v} \Big(\big\|\txt[u+1, v] - \txt[u, v]\big\|_2^2 + 
  \big\|\txt[u, v+1] - \txt[u, v]\big\|_2^2\Big)~,
\end{equation}
and write $\ltv = \ltvs(\txt_\bg) + \ltvs(\txt_\gr) + \sum_k \ltvs(\txt_k)$ the final 
penalization.

\noindent\textbf{Penalizing overlapping blocks.} We introduce a regularization term 
encouraging primitives to not overlap. Because penalizing volumetric intersections of 
superquadrics is difficult and computationally expensive, we instead propose to use a Monte 
Carlo alternative, by sampling 3D points in the scene and penalizing points belonging to more 
than $\lambda$ blocks, in a fashion similar
to~\cite{paschalidou2021neural}. Following~\cite{paschalidou2021neural}, $\lambda$ is set to
$1.95$ so that blocks could slightly overlap around their surface thus avoiding unrealistic 
floating blocks.  More specifically, considering a block $k$ and a 3D point $\point$, we 
define a soft 3D occupancy function $\occsqk$ as:
\begin{equation}
  \occsqk(\point) = \alpha_k \sigmoid\Big(\frac{1 - \impsq_k(\point)}{\temp}\Big)~,
\end{equation}%
where $\temp$ is a temperature hyperparameter and $\impsq_k$ is the superquadric 
inside-outside function~\cite{barr1981superquadrics} associated to the block $k$, such that 
$\impsq_k(\point) \leq 1$ if $\point$ lies inside the superquadric and $\impsq_k(\point) > 1$ 
otherwise.  Given a set of $M$ 3D points $\pointset$, our final regularization 
term can be written as:
\begin{equation}
  \lover = \frac{1}{M}\sum_{\point \in \pointset} \max \Big(\sum_{k=1}^{K} \occsqk(\point),\;
  \lambda\Big)~.
\end{equation}%
Note that in practice, for better efficiency and accuracy, we only sample points in the 
region where blocks are located, which can be identified using the block poses $\pose_{1:K}$.

\noindent\textbf{Optimization details.} We found that two elements were key to avoid bad 
local minima during optimization. First, while transparent meshes enable
differentiability \wrt the number of primitives, we observed a failure mode where two semi 
opaque meshes model the same 3D region. To prevent this behavior, we propose to inject 
gaussian noise before the sigmoid in the transparencies $\alpha_{1:K}$ to create 
stochasticity when values are not close to the sigmoid saturation, and thus encourage values 
that are close binary. Second, another failure mode we observed is one where the planar 
ground is modeling the entire scene. We avoid this by leveraging a two-stage curriculum 
learning scheme, where texture maps are downscaled by 8 during the first stage. We 
empirically validate these two contributions in~\Cref{sec:analysis}. We provide other 
implementation details in the supplementary material.

\section{Experiments}


\subsection{DTU Benchmark}\label{sec:dtu}

\begin{table}
  \centering
  \addtolength{\tabcolsep}{-2.9pt}
  \caption{\looseness=-1 \textbf{Quantitative results on DTU~\cite{jensen2014large}.} We use
    the official DTU evaluation to report Chamfer Distance (CD) between 3D reconstruction and 
    ground-truth, \textbf{best} results are highlighted. We also highlight the average number 
    of primitives found (\#P) in green (smaller than 10) or
    red (larger than 10). Our performances correspond to a single random run (random) and a 
    run automatically selected among 5 runs using the minimal rendering loss (auto). We 
  augment the best concurrent methods with a filtering step removing the ground from the 3D 
input.}
  \vspace{.05in}
\resizebox{\linewidth}{!}{%
  \begin{tabular}{lccccccccccccc}
  \toprule
  
  &  & \multicolumn{10}{c}{Chamfer Distance (CD) per scene} & Mean & Mean \\
  \cmidrule(lr){3-12}

  Method & Input & \tt S24 &\tt S31 &\tt S40 &\tt S45 &\tt S55 &\tt S59 &\tt S63 &\tt S75 &
  \tt S83 &\tt S105 & CD & \#P\\
  \midrule

  EMS~\cite{liu2022robust} & NeuS-mesh &
  8.42 & 8.53 & 7.84 & 6.98 & 7.2 & 8.57 & 7.77 & 8.69 & 4.74 & 9.11 & 7.78 &
  \cellcolor{mygreen!20} 9.6  \\

  EMS~\cite{liu2022robust} & 3D GT&
  6.77 & 5.93 & 3.36 & 6.91 & 6.52 & 3.50 & 4.72 & 7.08 & 7.25 & 6.10 & 5.82 & 
  \cellcolor{mygreen!20} 7.4  \\

  MBF~\cite{ramamonjisoa2022monteboxfinder} & NeuS-mesh &
  3.97 & 4.28 & 3.56 & 4.76 & 3.33 & 3.92 & 3.63 & 5.58 & 5.3 & 6.07 & 4.44 &
  \cellcolor{red!25} 53.5 \\

  MBF~\cite{ramamonjisoa2022monteboxfinder} & 3D GT &
  3.73 & 4.79 & 4.31 & 3.95 & 3.26 & 4.00 & 3.66 & 3.92 & 3.97 &\bf 4.25 & 3.98 &
  \cellcolor{red!25} 16.4 \\

  \bf Ours (random) & Image &
  5.41 &\bf 3.13 & 1.57 & 4.93 & 3.08 & 3.66 &\bf 3.40 & 2.78 & 3.94 & 4.85 & 3.67 &
  \cellcolor{mygreen!20} \bf 4.6\\

  \bf Ours (auto) & Image &
  \bf 3.25 &\bf 3.13 &\bf 1.16 &\bf 3.02 &\bf 2.98 &\bf 2.32 &\bf 3.40 &\bf 2.78 &\bf 3.43 & 
  5.21 &\bf 3.07 & \cellcolor{mygreen!20} 5.0\\

  \midrule

  EMS~\cite{liu2022robust} + filter & 3D GT &
  6.32 & 4.11 & 2.98 & 4.94 & 4.26 & 3.03 & 3.60 & 5.44 & 3.24 & 4.43 & 4.23 &
  \cellcolor{mygreen!20} 8.3 \\

  MBF~\cite{ramamonjisoa2022monteboxfinder} + filter & 3D GT &
  \bf 3.35 &\bf 2.95 &\bf 2.61 &\bf 2.19 &\bf 2.53 &\bf 2.47 &\bf 1.97 &\bf 2.60 &\bf 2.60 & 
  \bf 3.27 &\bf 2.65 & \cellcolor{red!25} 29.9 \\

  \bottomrule
  \end{tabular}
}

  \label{tab:dtu}
\end{table}

\noindent\textbf{Benchmark details.} DTU~\cite{jensen2014large} is an MVS dataset containing 
80 forward-facing scenes captured in a controlled indoor setting, where the 3D ground-truth 
points are obtained through a structured light scanner. We evaluate on 10 scenes 
(\texttt{S24}, \texttt{S31}, \texttt{S40}, \texttt{S45}, \texttt{S55}, \texttt{S59}, 
\texttt{S63}, \texttt{S75}, \texttt{S83}, \texttt{S105}) that have different geometries and a 
3D decomposition that is relatively intuitive. We use standard processing 
practices~\cite{yariv2020multiview,
yariv2021volume, darmon2022improving}, resize the images to $400 \times 300$ and run our 
model with $K = 10$ on all available views for each scene (49 or 64 depending on the scenes).  
We use the official evaluation presented in~\cite{jensen2014large}, which computes the 
Chamfer distance between the ground-truth points and points sampled from the 3D 
reconstruction, filtered out if not in the neighborhood of the ground-truth points. We 
evaluate two state-of-the-art methods for 3D decomposition,
EMS~\cite{liu2022robust} and MonteboxFinder (MBF)~\cite{ramamonjisoa2022monteboxfinder}, by 
applying them to the 3D ground-truth point clouds. We also evaluate them in a setup 
comparable to ours, where the state-of-the-art MVS method NeuS~\cite{wang2021neus} is first 
applied to the multi-view images to extract a mesh, which is then used as input to the 3D 
decomposition methods. We refer to this input as ``NeuS-mesh''.

\noindent\textbf{Results.} We compare our Chamfer distance performances to these 
state-of-the-art 3D decomposition methods in~\Cref{tab:dtu}. For each method, we report the 
input used and highlight the average number of discovered primitives \#P in green (smaller 
than 10) or red (larger than 10). Intuitively, overly large numbers of primitives lead to 
less intuitive and manipulative scene representations. Our performances correspond to a 
single random run (random) and a run automatically selected among 5 runs using the minimal 
rendering loss (auto). We augment the best concurrent methods with a filtering step using 
RANSAC to remove the planar ground from the 3D input.  Overall, we obtain results that are 
much more satisfactory than prior works. On the one hand, EMS outputs a reasonable number of 
primitives but has a high Chamfer distance reflecting bad 3D reconstructions. On the other 
hand, MBF yields a lower Chamfer distance (even better than ours with the filtering step) but 
it outputs a significantly higher number of primitives, thus reflecting over-decompositions.

\begin{figure}
  \centering
  \addtolength{\tabcolsep}{-4pt}
  \resizebox{\linewidth}{!}{%
  \begin{tabular}{@{}ccccc|c@{}}
    \small Input view &\small  GT point cloud &\small EMS~\cite{liu2022robust} & 
    \small MBF~\cite{ramamonjisoa2022monteboxfinder} &\small\bf Ours &\small \bf Ours rendering\\

    \includegraphics[width=0.158\linewidth]{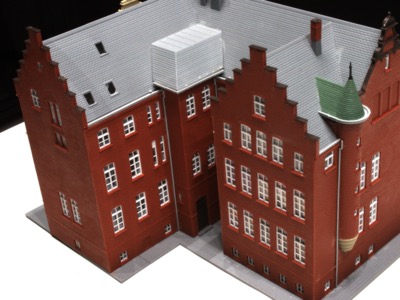} & 
    \includegraphics[width=0.158\linewidth]{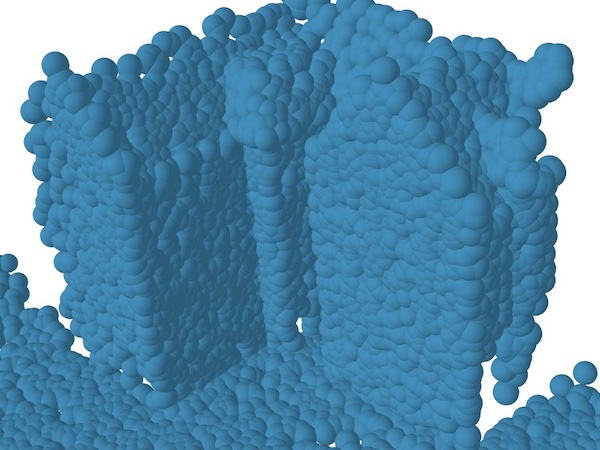} & 
    \includegraphics[width=0.158\linewidth]{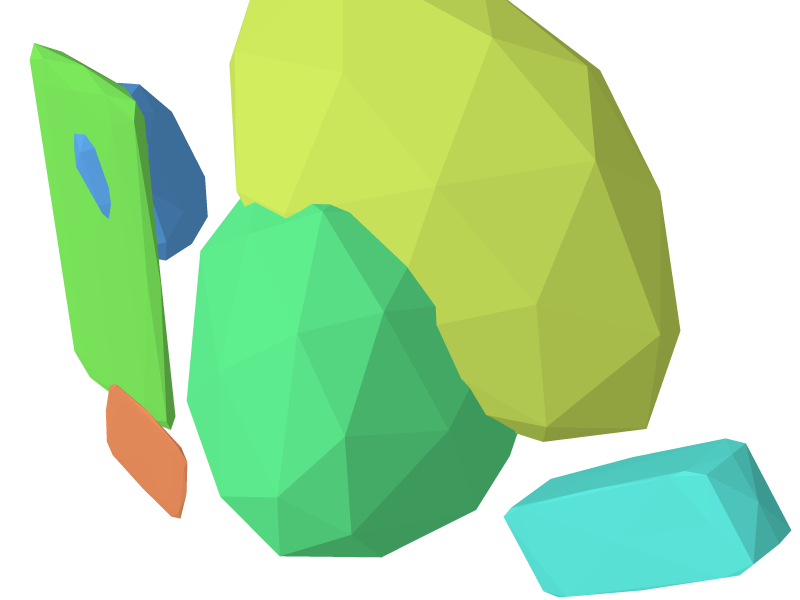} & 
    \includegraphics[width=0.158\linewidth]{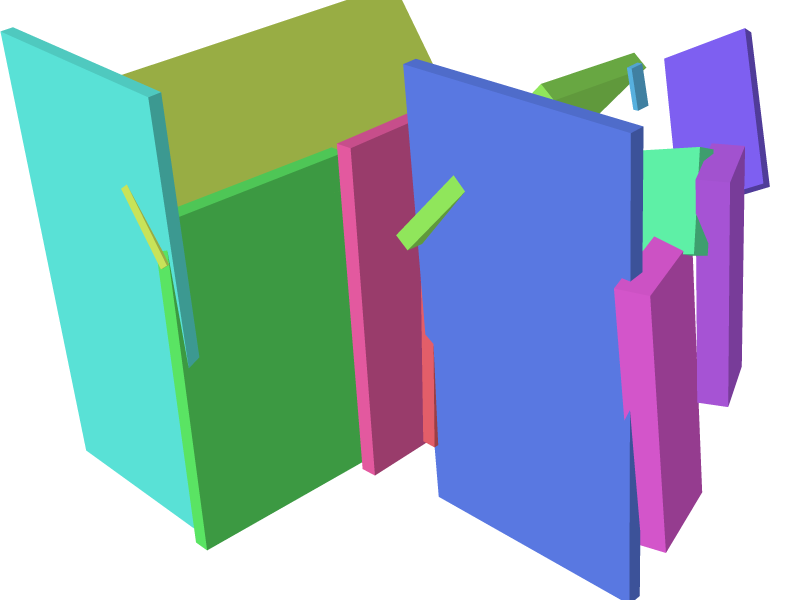} & 
    \includegraphics[width=0.158\linewidth]{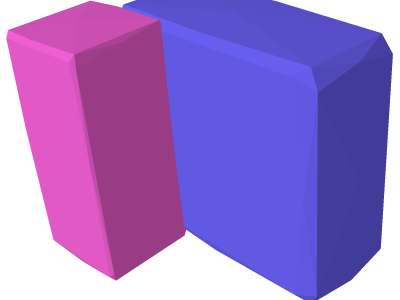}&
    \includegraphics[width=0.158\linewidth]{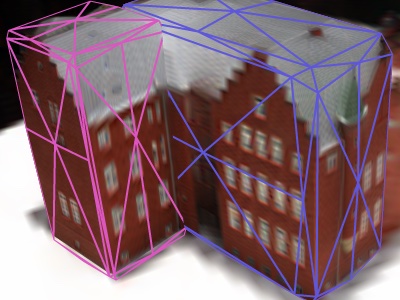}\\

    \includegraphics[width=0.158\linewidth]{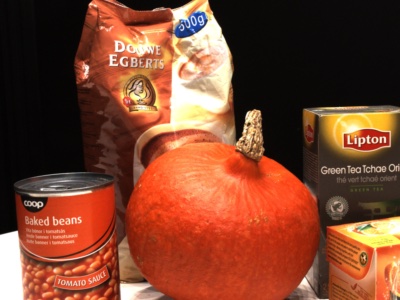} & 
    \includegraphics[width=0.158\linewidth]{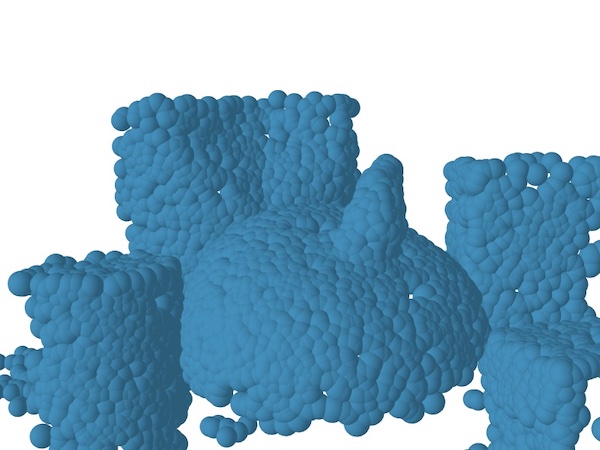} & 
    \includegraphics[width=0.158\linewidth]{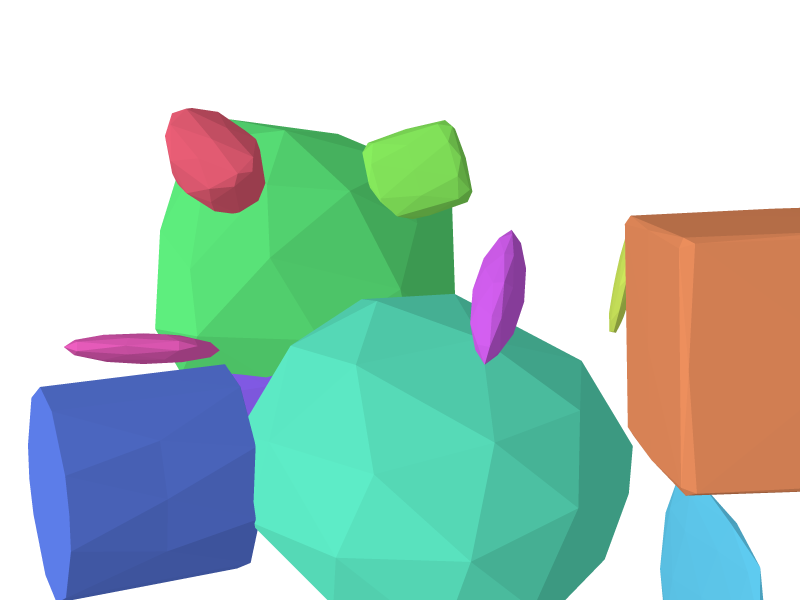} & 
    \includegraphics[width=0.158\linewidth]{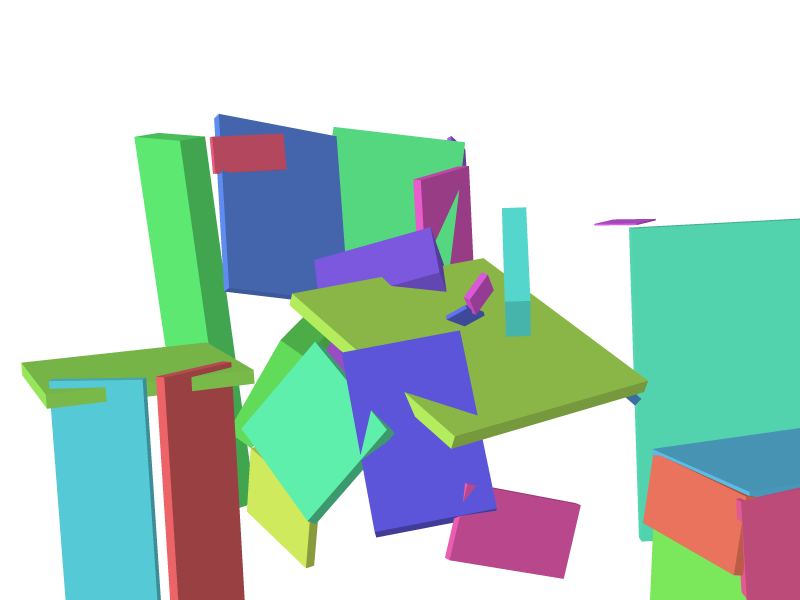} & 
    \includegraphics[width=0.158\linewidth]{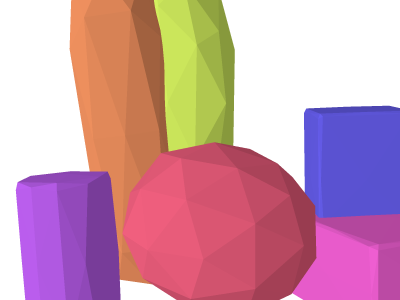}&
    \includegraphics[width=0.158\linewidth]{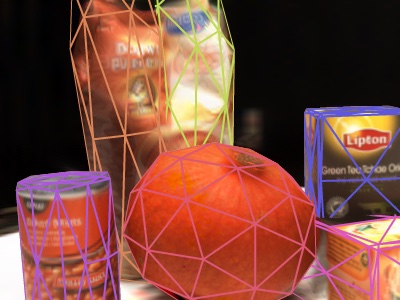}\\

    \includegraphics[width=0.158\linewidth]{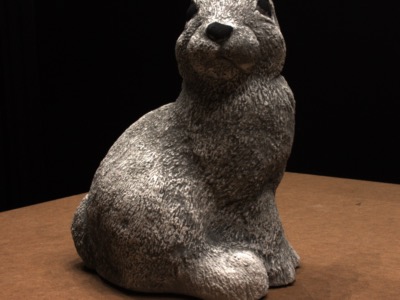} & 
    \includegraphics[width=0.158\linewidth]{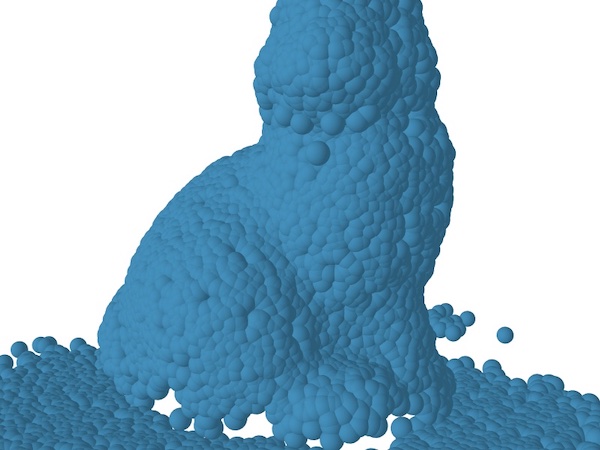} & 
    \includegraphics[width=0.158\linewidth]{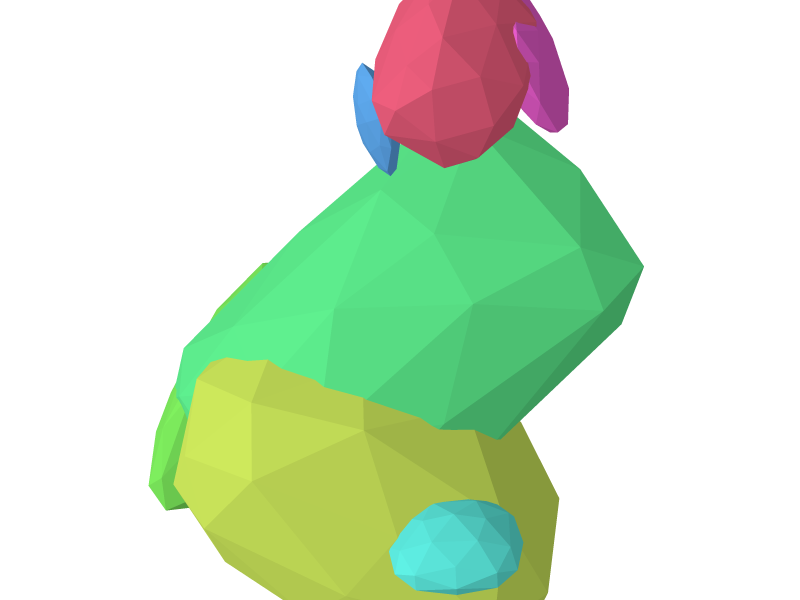} & 
    \includegraphics[width=0.158\linewidth]{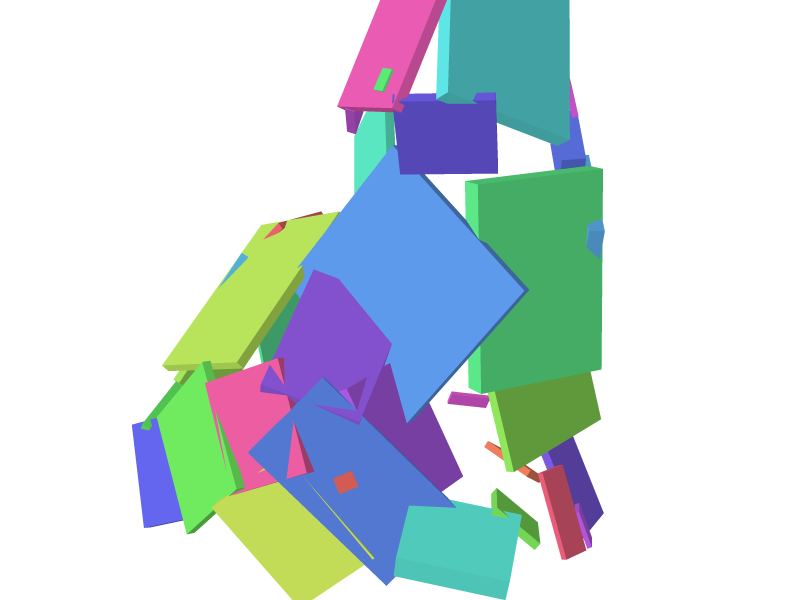} & 
    \includegraphics[width=0.158\linewidth]{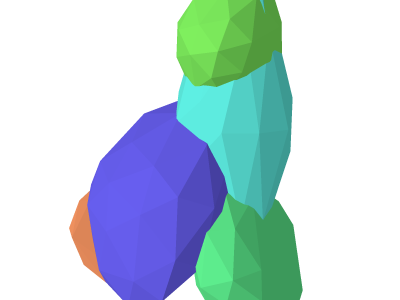}&
    \includegraphics[width=0.158\linewidth]{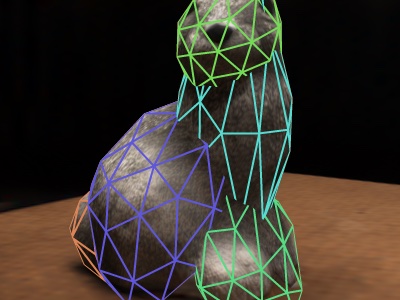}\\

    \includegraphics[width=0.158\linewidth]{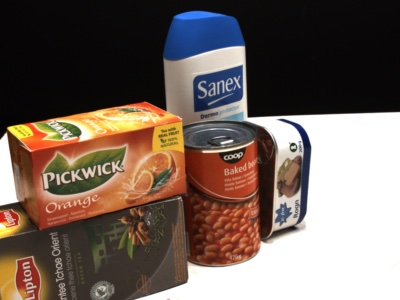} & 
    \includegraphics[width=0.158\linewidth]{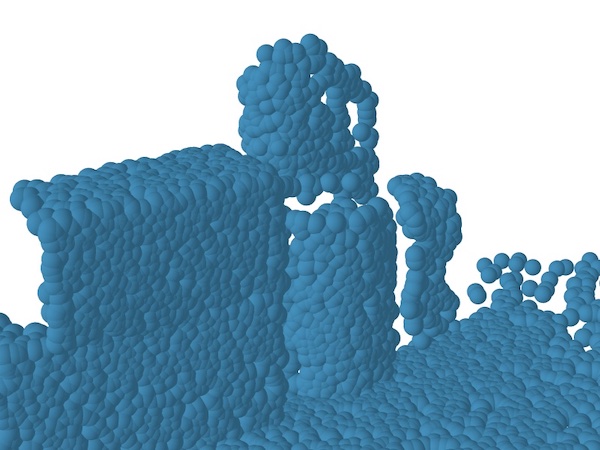} & 
    \includegraphics[width=0.158\linewidth]{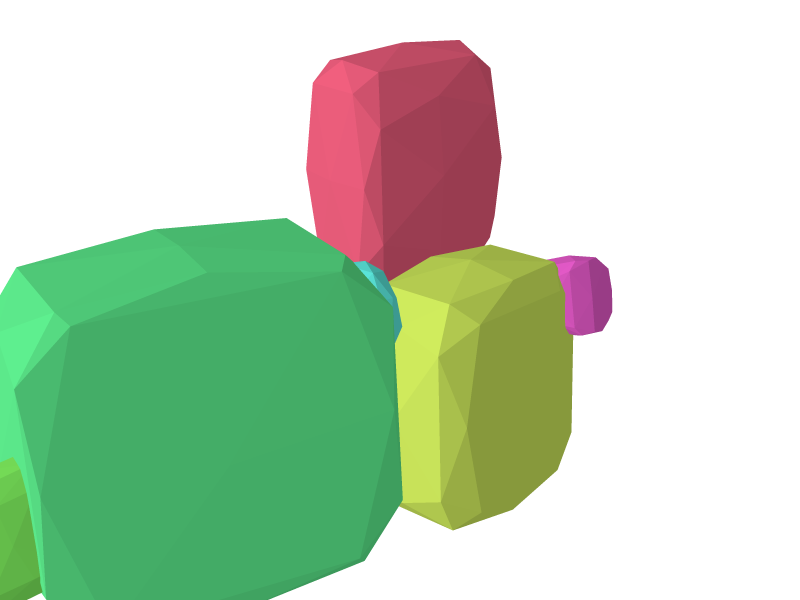} & 
    \includegraphics[width=0.158\linewidth]{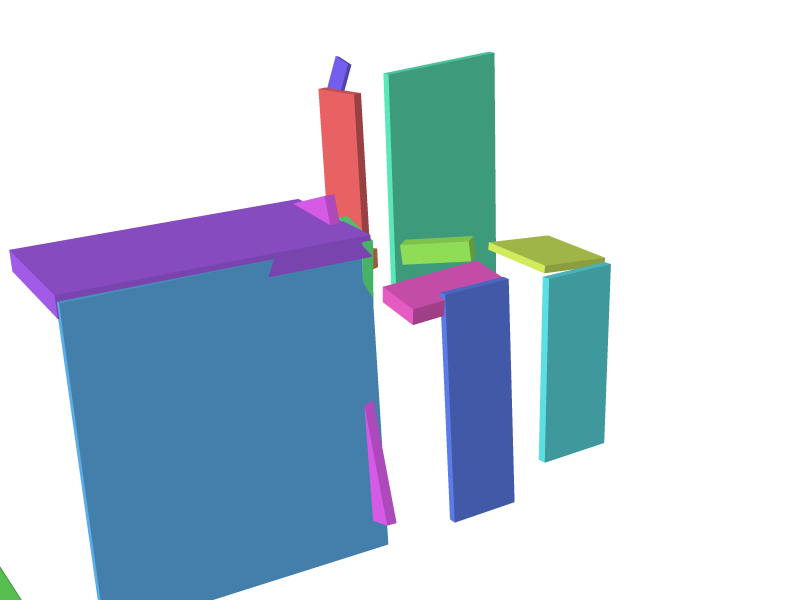} & 
    \includegraphics[width=0.158\linewidth]{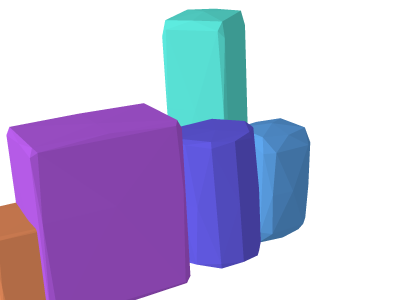}&
    \includegraphics[width=0.158\linewidth]{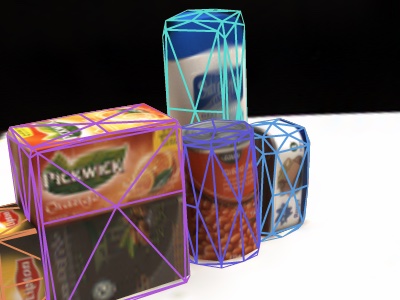}\\

    \includegraphics[width=0.158\linewidth]{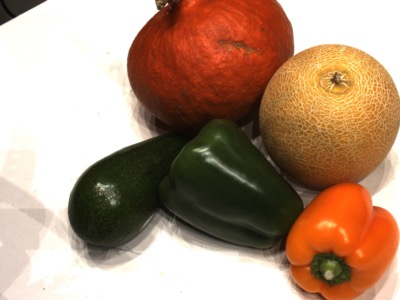} & 
    \includegraphics[width=0.158\linewidth]{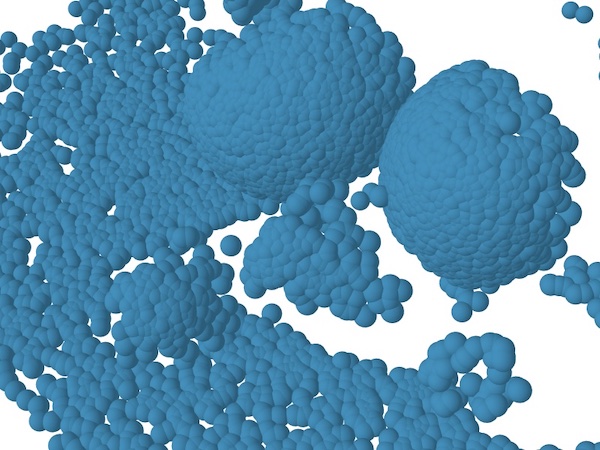} & 
    \includegraphics[width=0.158\linewidth]{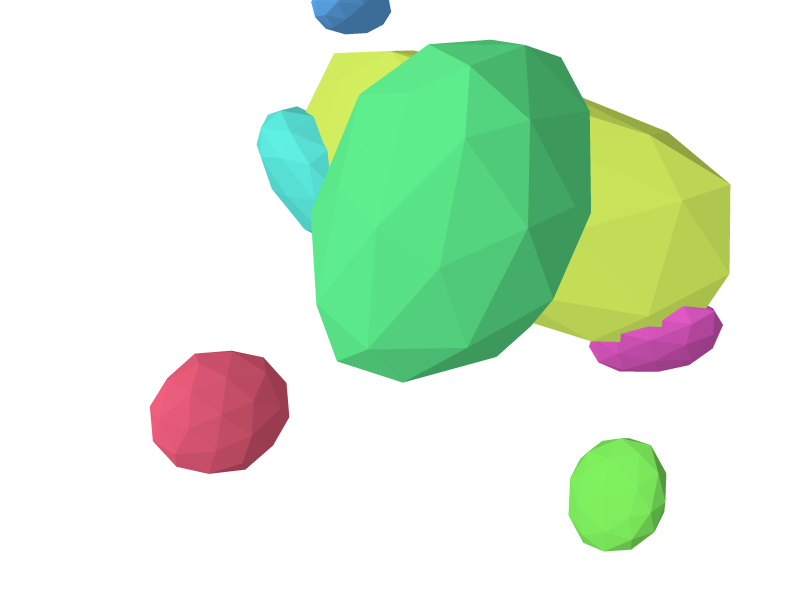} & 
    \includegraphics[width=0.158\linewidth]{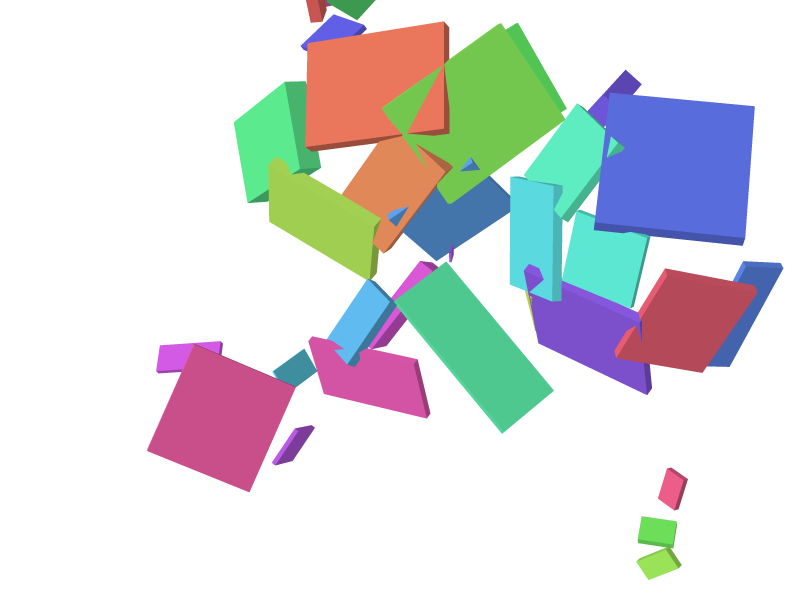} & 
    \includegraphics[width=0.158\linewidth]{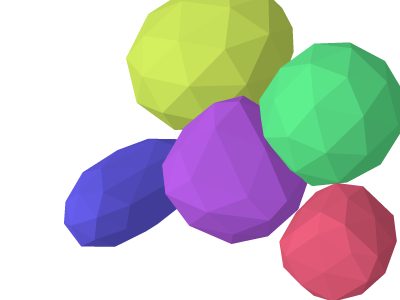}&
    \includegraphics[width=0.158\linewidth]{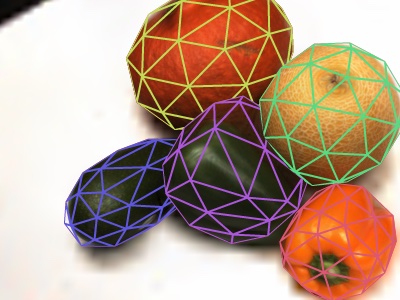}\\

    \includegraphics[width=0.158\linewidth]{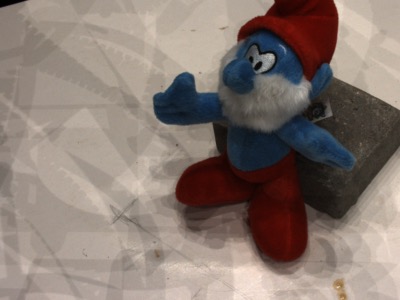} & 
    \includegraphics[width=0.158\linewidth]{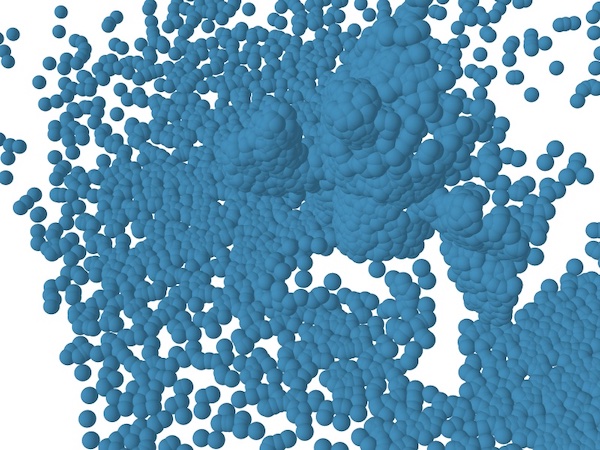} & 
    \includegraphics[width=0.158\linewidth]{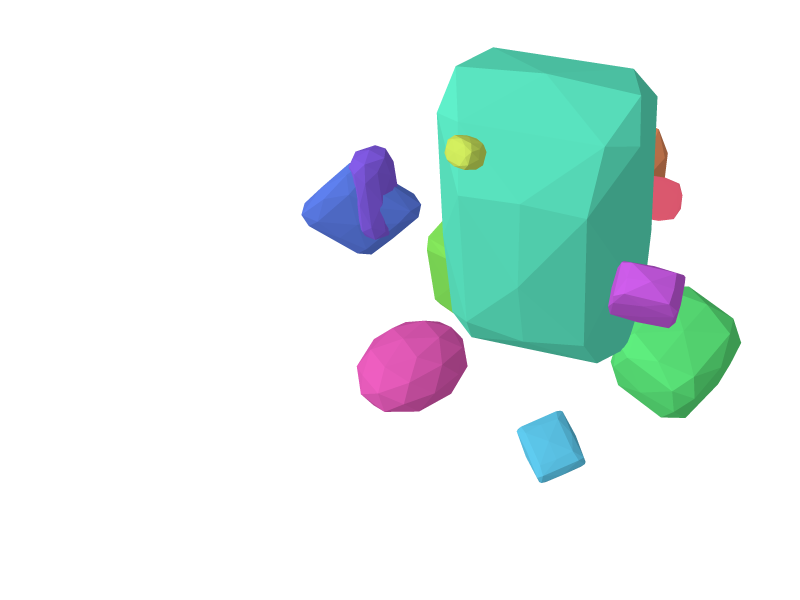} & 
    \includegraphics[width=0.158\linewidth]{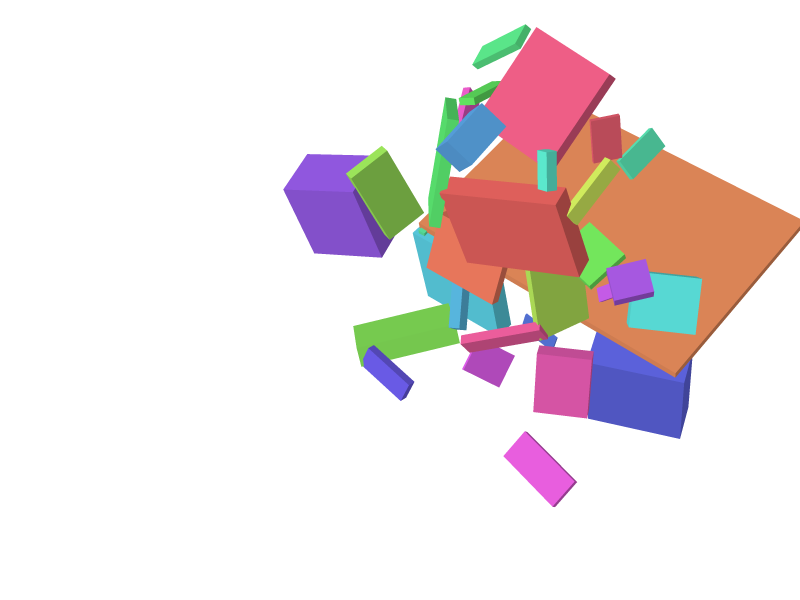} & 
    \includegraphics[width=0.158\linewidth]{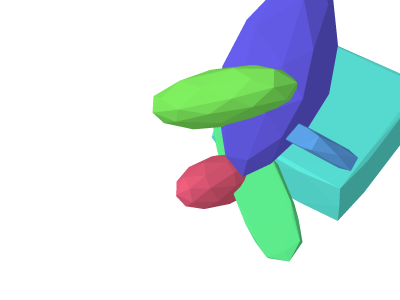}&
    \includegraphics[width=0.158\linewidth]{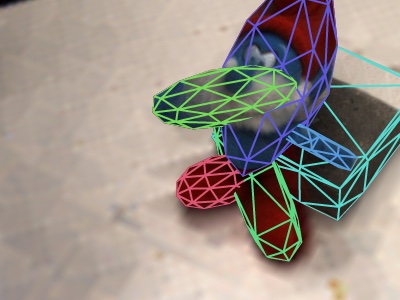}\\

  \end{tabular}
  }
  \vspace{-.2em}
  \caption{\textbf{Qualitative comparisons on DTU~\cite{jensen2014large}.} We compare our 
  model to state-of-the-art methods (augmented with a preprocessing step to remove the 3D ground) 
  which, unlike ours, find primitives in the ground-truth 
point cloud that is noisy and incomplete. Additionally, our approach is the only one able
to capture the scene appearance (last column). 
}
  \label{fig:dtu_comparison}
\end{figure}

\looseness=-1 Our approach is qualitatively compared in~\Cref{fig:dtu_comparison} to the best 
EMS and MBF models, which correspond to the ones applied on the 3D ground truth and augmented 
with the filtering step. Because the point clouds are noisy and incomplete (see 360$^{\circ}$ 
renderings in our supplementary material), EMS and MBF struggle
to find reasonable 3D decompositions: EMS misses some important parts, while MBF 
over-decomposes the 3D into piecewise planar surfaces. On the contrary, our model is able to 
output meaningful 3D decompositions with varying numbers of primitives and very different 
shapes.  Besides, ours is the only approach that recovers the scene appearance (last column).  
Also note that it produces a complete 3D scene, despite being only optimized on 
forward-facing views.

\subsection{Real-Life Data and Applications}\label{sec:real_data}

\begin{figure}
  \centering
  \addtolength{\tabcolsep}{-5pt}
  \renewcommand{\arraystretch}{0.8}
  \resizebox{\linewidth}{!}{%
  \begin{tabular}{@{}cccccccc@{}}
    \multicolumn{2}{c}{\small Input (subset)} &\small Rendering &\small Output &
    \multicolumn{2}{c}{\small Novel views rendering} & \multicolumn{2}{c}{\small Novel views 
    output} \\
    
    \includegraphics[width=0.0403\linewidth]{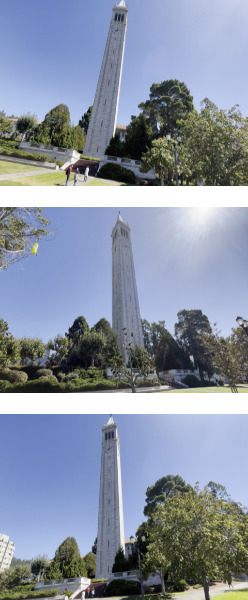} & 
    \includegraphics[height=0.0975\linewidth,width=0.13\linewidth]{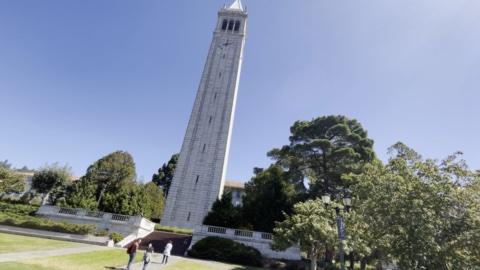}
    &
    \includegraphics[height=0.0975\linewidth,width=0.13\linewidth]{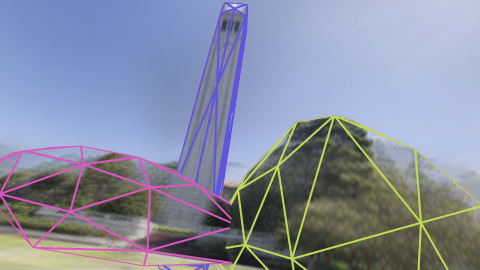} 
    &
    \includegraphics[height=0.0975\linewidth,width=0.13\linewidth]{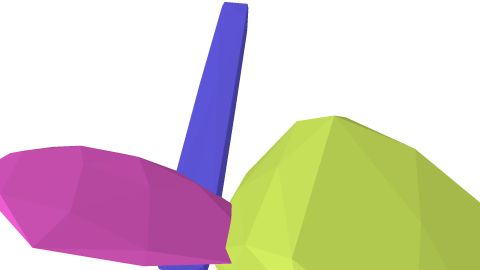} 
    &
    \includegraphics[height=0.0975\linewidth,width=0.13\linewidth]{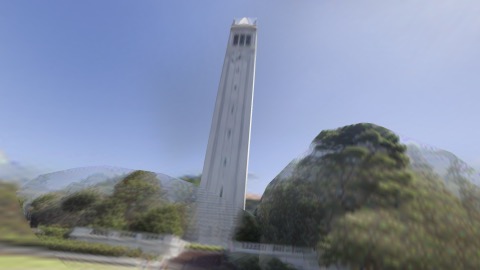} 
    &
    \includegraphics[height=0.0975\linewidth,width=0.13\linewidth]{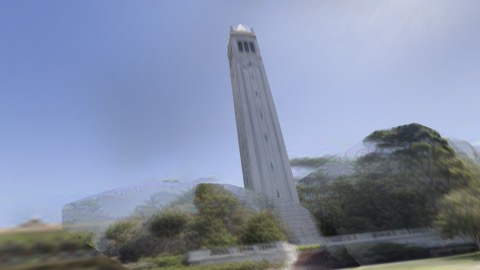} 
    &
    \includegraphics[height=0.0975\linewidth,width=0.13\linewidth]{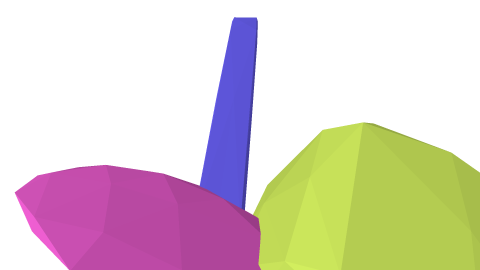} 
    &
    \includegraphics[height=0.0975\linewidth,width=0.13\linewidth]{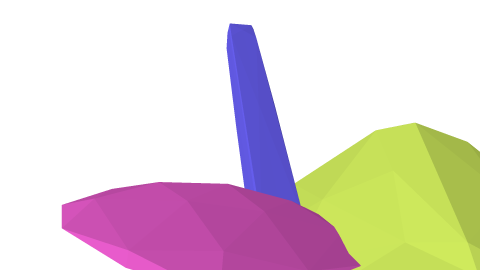}\\

    \includegraphics[width=0.0403\linewidth]{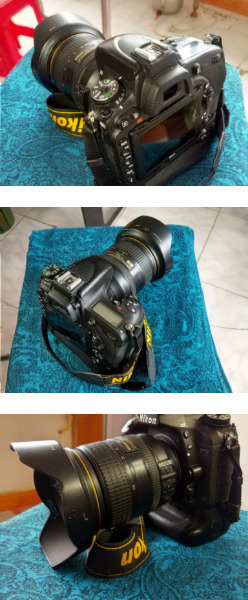} & 
    \includegraphics[width=0.13\linewidth]{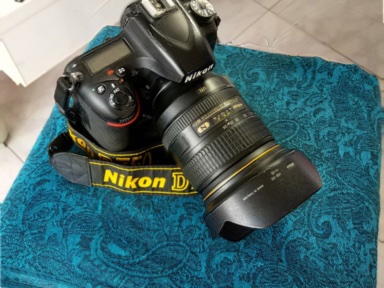} & 
    \includegraphics[width=0.13\linewidth]{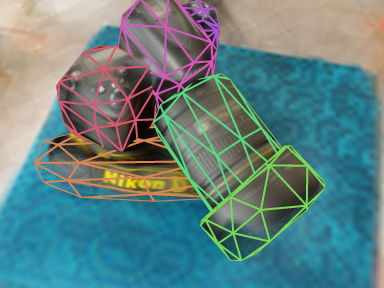} & 
    \includegraphics[width=0.13\linewidth]{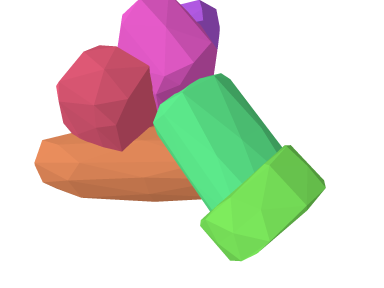} & 
    \includegraphics[width=0.13\linewidth]{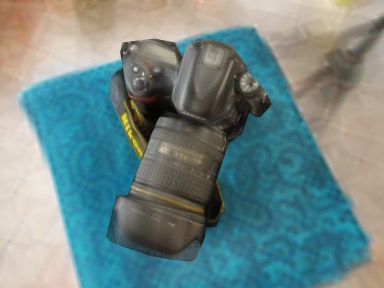} & 
    \includegraphics[width=0.13\linewidth]{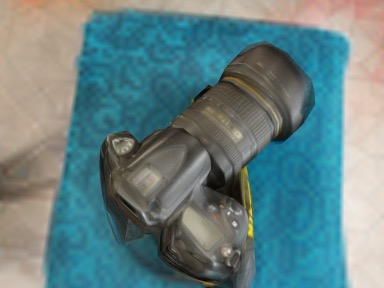} & 
    \includegraphics[width=0.13\linewidth]{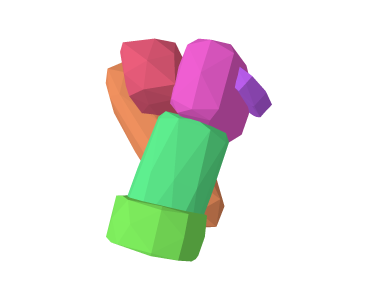} & 
    \includegraphics[width=0.13\linewidth]{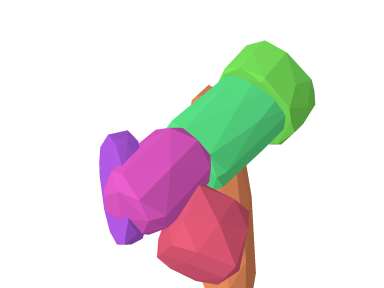}\\

    \includegraphics[width=0.0403\linewidth]{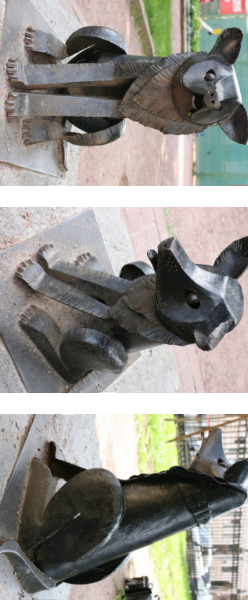} & 
    \includegraphics[width=0.13\linewidth]{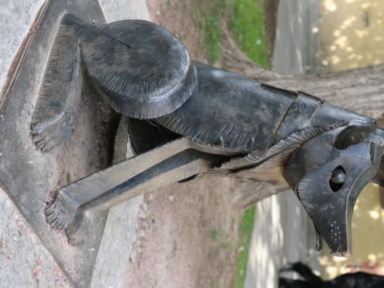} & 
    \includegraphics[width=0.13\linewidth]{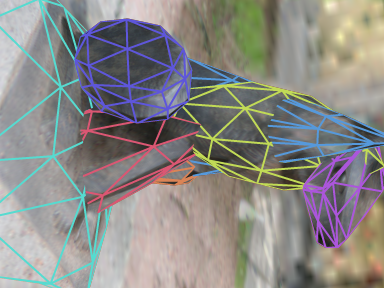} & 
    \includegraphics[width=0.13\linewidth]{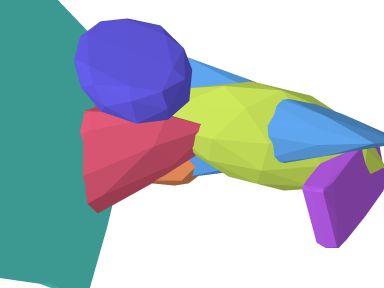} & 
    \includegraphics[width=0.13\linewidth]{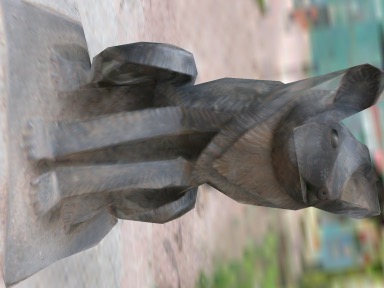} & 
    \includegraphics[width=0.13\linewidth]{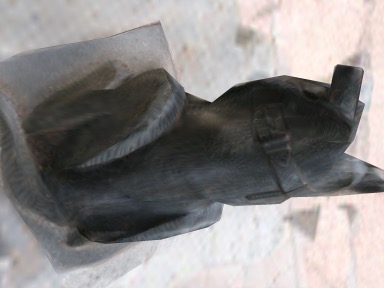} & 
    \includegraphics[width=0.13\linewidth]{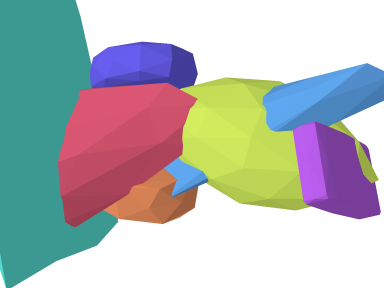} & 
    \includegraphics[width=0.13\linewidth]{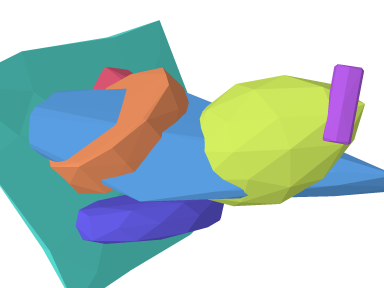}\\

    \includegraphics[width=0.0403\linewidth]{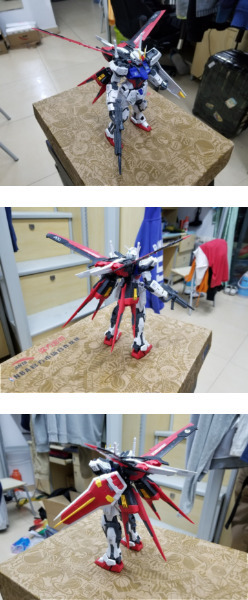} & 
    \includegraphics[width=0.13\linewidth]{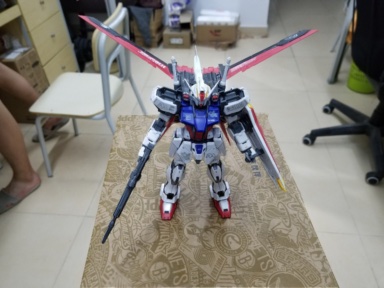} & 
    \includegraphics[width=0.13\linewidth]{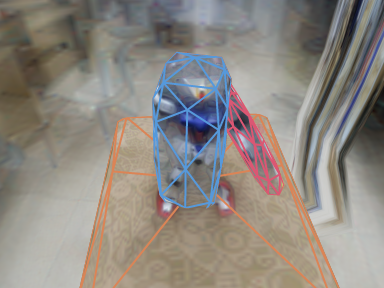} & 
    \includegraphics[width=0.13\linewidth]{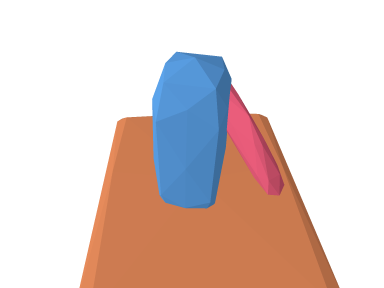} & 
    \includegraphics[width=0.13\linewidth]{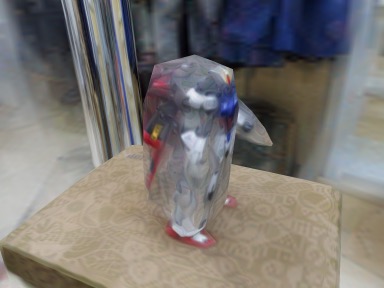} & 
    \includegraphics[width=0.13\linewidth]{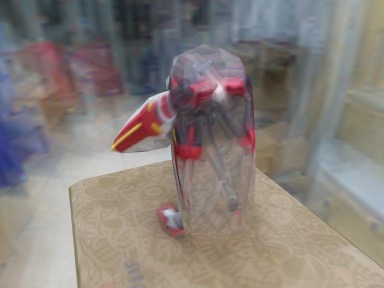} & 
    \includegraphics[width=0.13\linewidth]{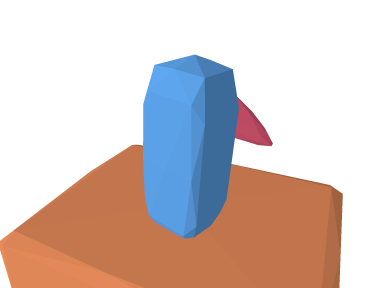} & 
    \includegraphics[width=0.13\linewidth]{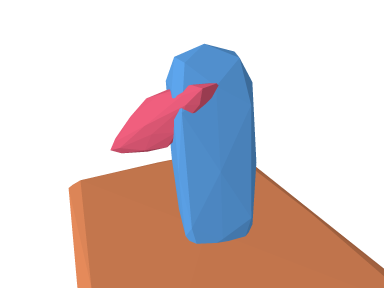}\\
    \specialrule{.5pt}{2pt}{4pt}

    \includegraphics[width=0.0403\linewidth]{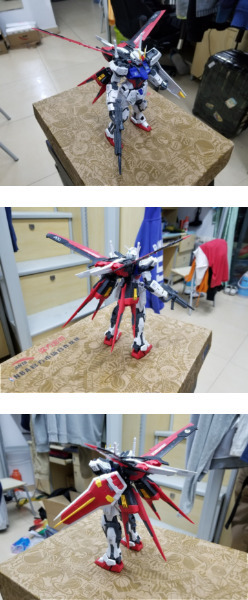} & 
    \includegraphics[width=0.13\linewidth]{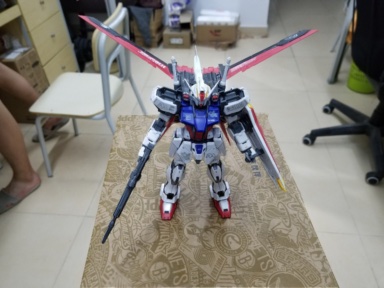} & 
    \includegraphics[width=0.13\linewidth]{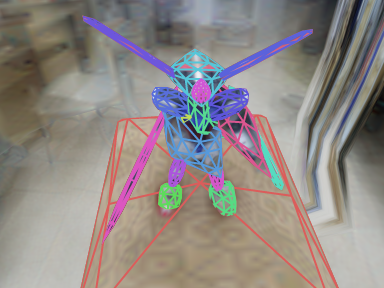} & 
    \includegraphics[width=0.13\linewidth]{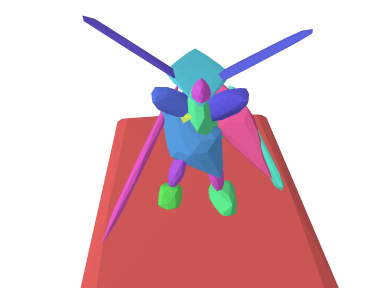} & 
    \includegraphics[width=0.13\linewidth]{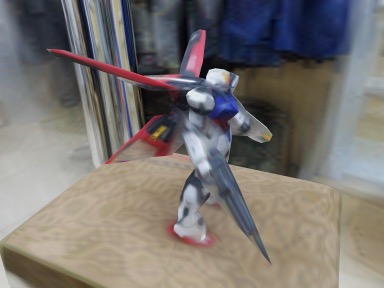} & 
    \includegraphics[width=0.13\linewidth]{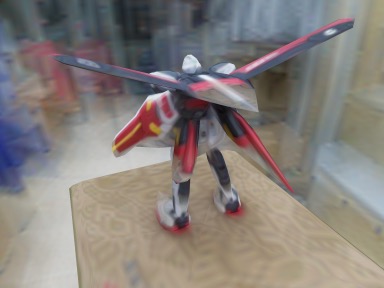} & 
    \includegraphics[width=0.13\linewidth]{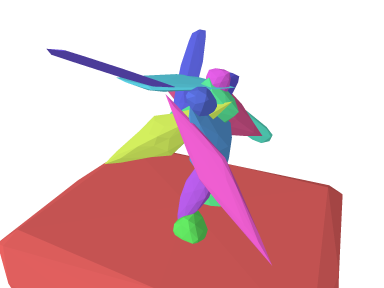} & 
    \includegraphics[width=0.13\linewidth]{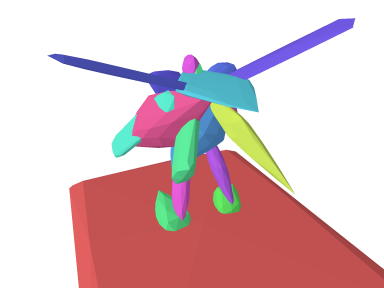}\\
  \end{tabular}
  }
  \vspace{-.2em}
  \caption{\textbf{Qualitative results on real-life data.} We run our default model ($K = 
    10$) on scenes from Nerfstudio~\cite{tancik2023nerfstudio} (first row) and  BlendedMVS~\cite{yao2020blendedmvs} (all other rows). The last row corresponds to results where the 
maximum number of primitives is increased to $K = 50$, yielding 17 effective primitives found.}
  \label{fig:real_data}
\end{figure}

We present qualitative results on real-life captures in~\Cref{fig:real_data}. The first row 
corresponds to the \textit{Campanile} scene from Nerfstudio 
repository~\cite{tancik2023nerfstudio} and the last four rows correspond to BlendedMVS 
scenes~\cite{yao2020blendedmvs} that were selected in~\cite{yariv2021volume}. We adapt their 
camera conventions to ours and resize the images to roughly $400 \times 300$. From left to 
right, we show a subset of the input views, a rendering overlaid with the primitive edges, 
the primitives, as well as two novel view synthesis results. For each scene, we run our model 
5 times and automatically select the results with the minimal rendering loss. We set the 
maximum number of primitives to $K = 10$, except the last row where it is increased to $K = 
50$ due to the scene complexity. These results show that despite its simplicity, our approach 
is surprisingly robust. Our method is still able to compute 3D decompositions that capture 
both appearances and meaningful geometry on a variety of scene types. In addition, increasing 
the maximum number of primitives $K$ allows us to easily adapt the decomposition granularity 
(last row).

\begin{figure}
  \centering
  \addtolength{\tabcolsep}{-5pt}
  \renewcommand{\arraystretch}{0.8}
  \resizebox{\linewidth}{!}{%
  \begin{tabular}{@{}ccccc@{}}
    \small Input (subset) & \multicolumn{2}{c}{\small Amodal view synthesis - 
    Nerfacto~\cite{tancik2023nerfstudio} }
    & \multicolumn{2}{c}{\small Amodal view synthesis - \textbf{Ours}} \\
    \includegraphics[width=0.19\linewidth]{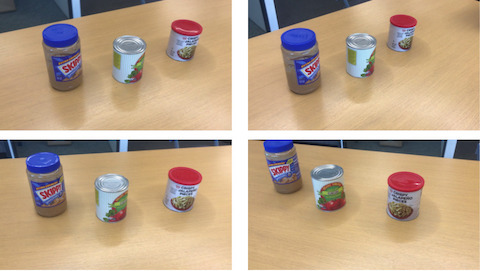}
    &\includegraphics[width=0.19\linewidth]{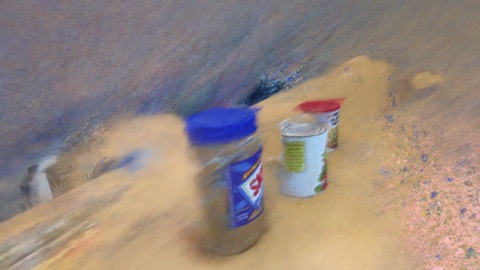}
    &\includegraphics[width=0.19\linewidth]{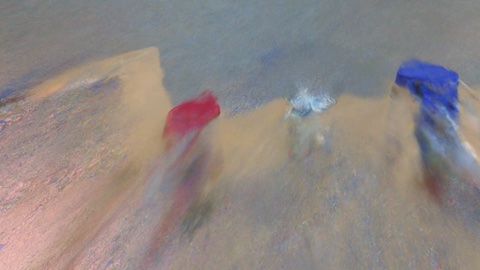}
    &\includegraphics[width=0.19\linewidth]{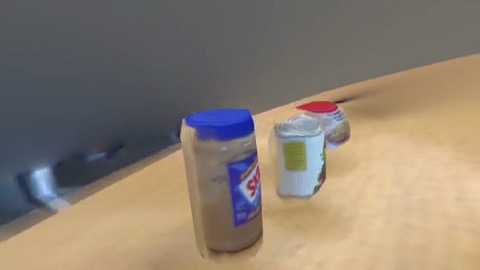}
    &\includegraphics[width=0.19\linewidth]{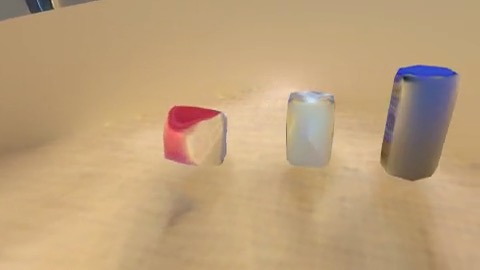}\\

    \specialrule{.5pt}{2pt}{4pt}

    \small Input (subset) & \multicolumn{2}{c}{\small Scene editing - Removing ears }
    & \multicolumn{2}{c}{\small Scene editing - Moving arm} \\
    \includegraphics[width=0.19\linewidth]{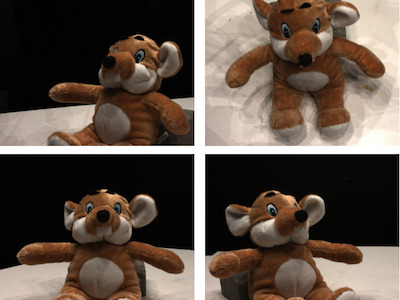}
    &\includegraphics[width=0.19\linewidth]{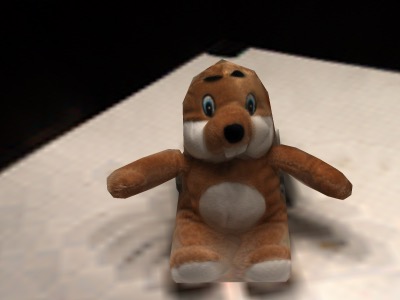}
    &\includegraphics[width=0.19\linewidth]{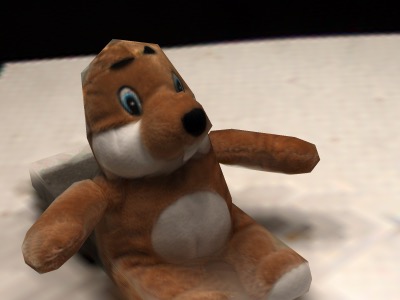}
    &\includegraphics[width=0.19\linewidth]{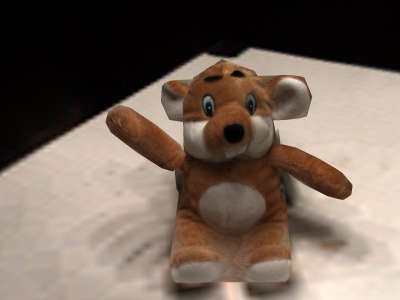}
    &\includegraphics[width=0.19\linewidth]{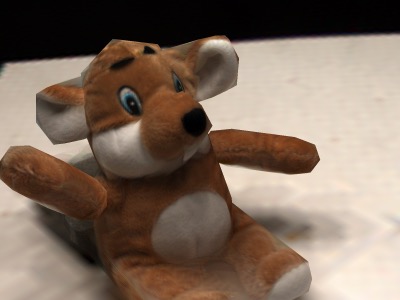}\\

    \specialrule{.5pt}{2pt}{4pt}
  \end{tabular}
  }

  \centering
  \addtolength{\tabcolsep}{-0.2pt}
  \renewcommand{\arraystretch}{0.7}
  \resizebox{\linewidth}{!}{%
  \begin{tabular}{@{}cccc@{}}
    \includegraphics[trim=100 0 0 100, clip, width=0.24\linewidth]{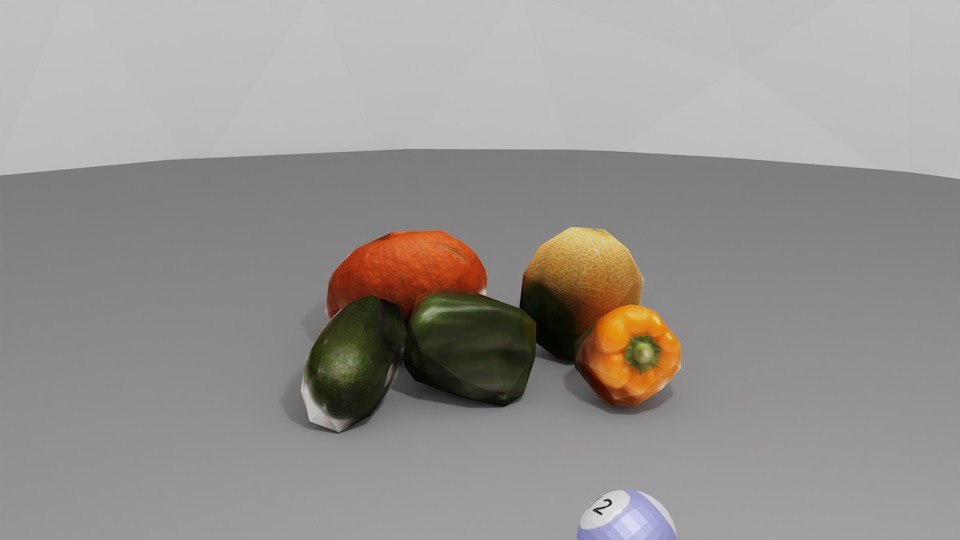}&
    \includegraphics[trim=100 0 0 100, clip, width=0.24\linewidth]{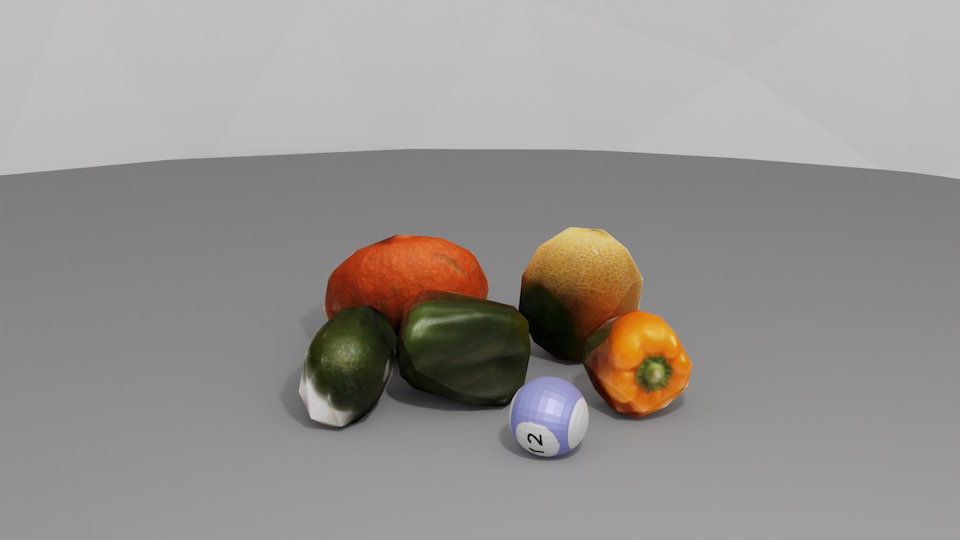}&
    \includegraphics[trim=100 0 0 100, clip, width=0.24\linewidth]{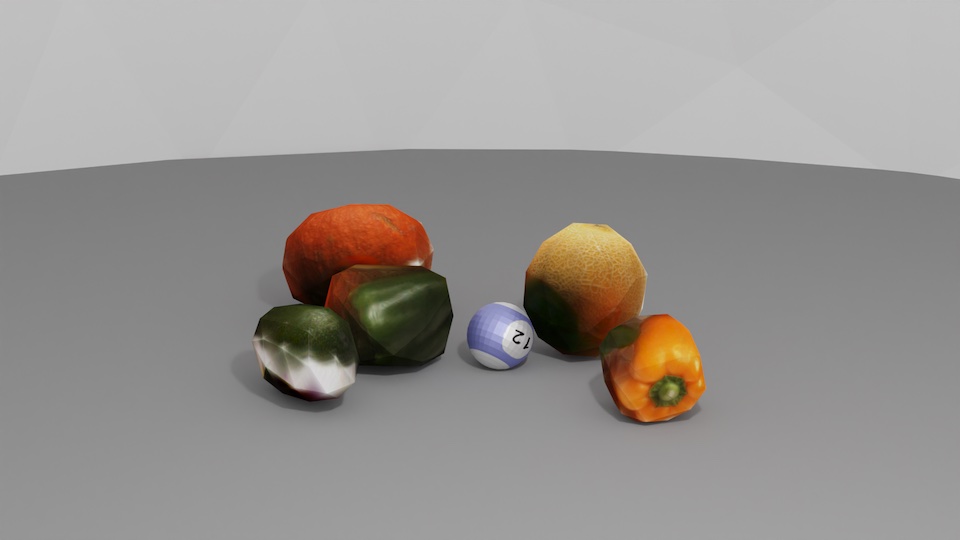}&
    \includegraphics[trim=100 0 0 100, clip,width=0.24\linewidth]{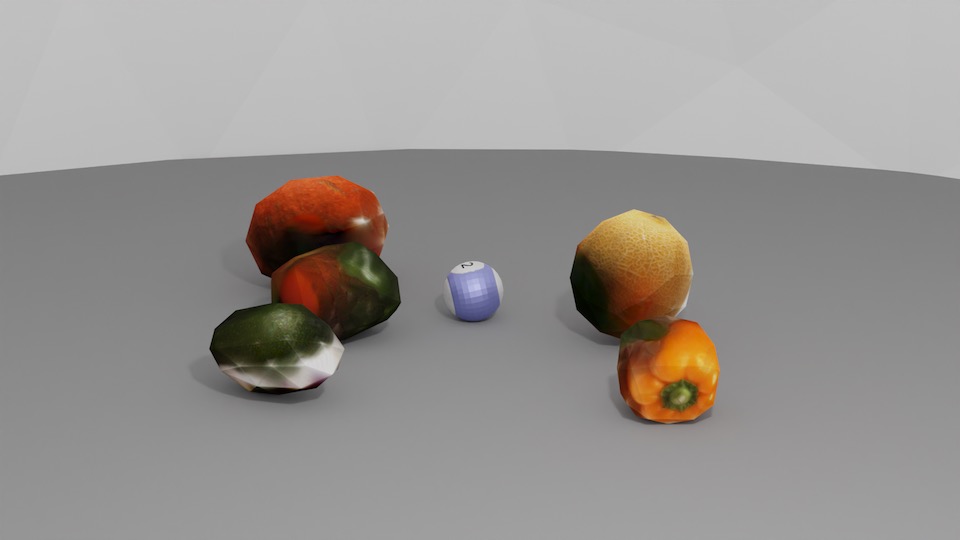}\\

    \includegraphics[trim=100 50 0 0, clip, width=0.24\linewidth]{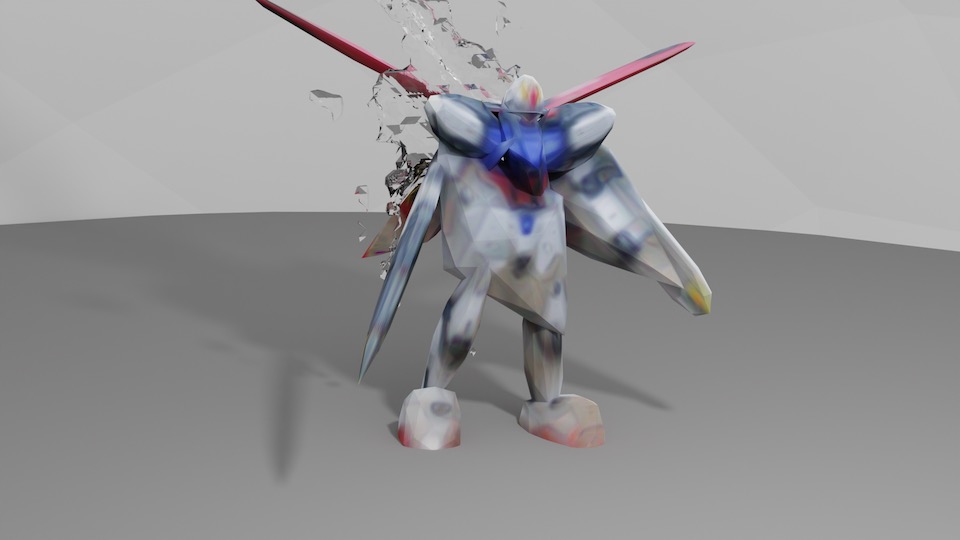}&
    \includegraphics[trim=100 50 0 0, clip, width=0.24\linewidth]{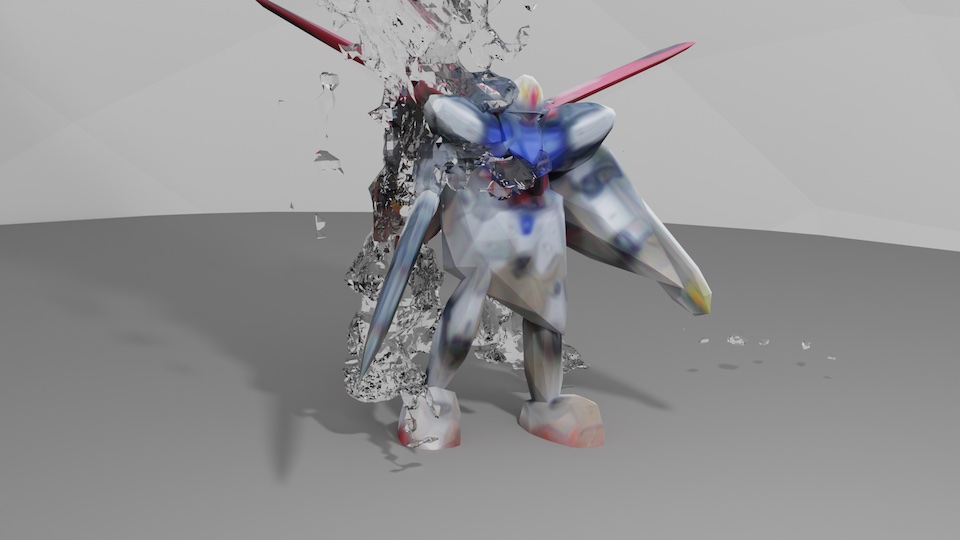}&
    \includegraphics[trim=100 50 0 0, clip, width=0.24\linewidth]{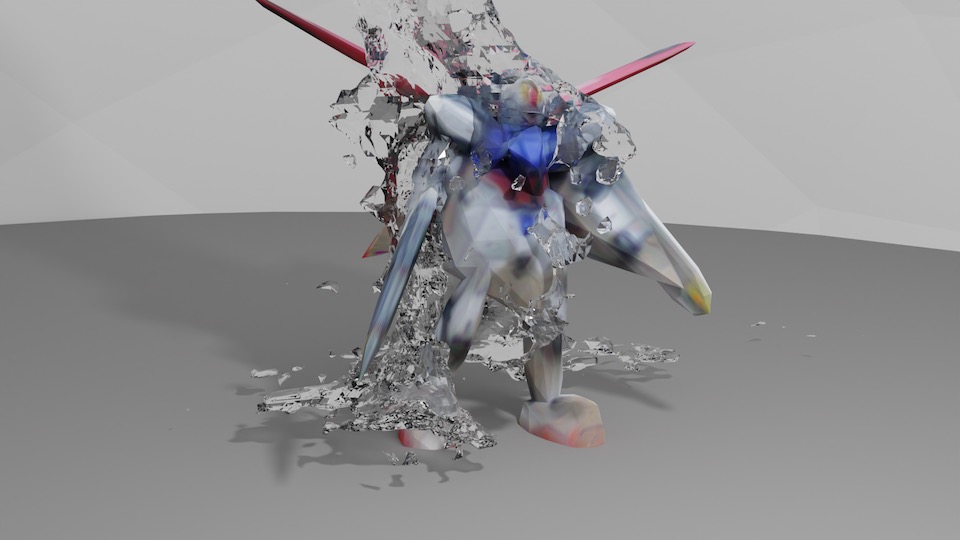}&
    \includegraphics[trim=100 50 0 0, clip,width=0.24\linewidth]{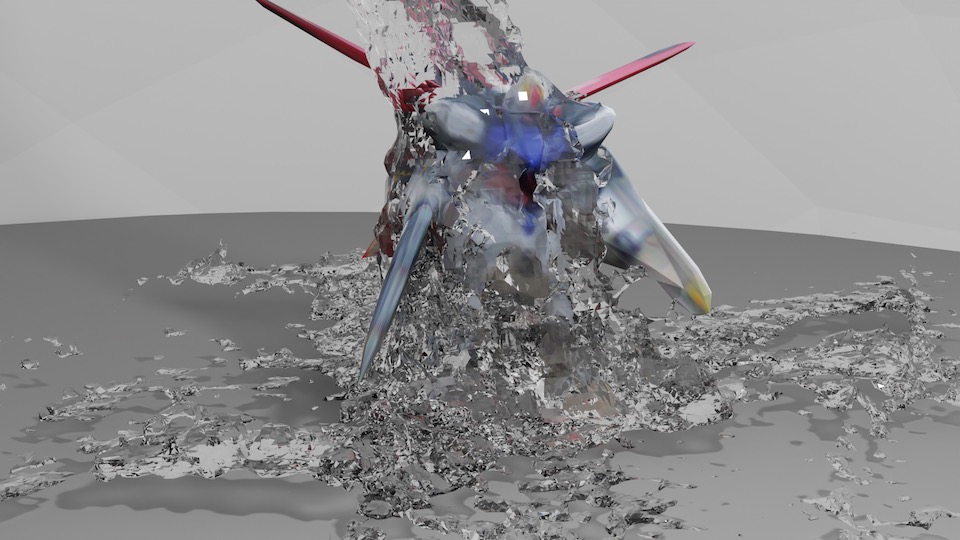}\\
  \end{tabular}
  }
  \caption{\textbf{Applications.} \textbf{(top)} Given a set of views constrained to limited 
    viewpoint variations,
  we compare amodal view synthesis results using Nerfacto~\cite{tancik2023nerfstudio} and our 
  approach.  \textbf{(middle)} After optimization, we can easily modify the rendered scene by
  editing the different parts. \textbf{(bottom)} Our primitive-based representation enables 
  straightforward physics-based simulations, such as throwing a ball at the objects or 
  pouring water on the scene.
  }
  \label{fig:applications}
\end{figure}

\looseness=-1 In~\Cref{fig:applications}, we demonstrate other advantages of our approach.  
First, compared to NeRF-based approaches like Nerfacto~\cite{tancik2023nerfstudio} which only 
reconstruct visible regions, our method performs amodal scene completion (first row). Second, 
our textured primitive decomposition allows to easily edit the 3D scene (second row). Finally, 
our optimized primitive meshes can be directly imported into standard computer graphics 
softwares like Blender to perform physics-based simulations (bottom).

\subsection{Analysis}\label{sec:analysis}

\begin{table}
  \small
  \addtolength{\tabcolsep}{-1pt}
  \caption{\textbf{Ablation study on DTU~\cite{jensen2014large}.} We report metrics averaged 
    over five runs: number of
    primitives (\#P), Chamfer Distance (CD) and rendering metrics (PSNR in
    dB and SSIM, LPIPS in \%).
  \textbf{Best} and \underline{second best} are highlighted, \#P variability is emphasized in 
green (smaller than 5) and red (larger than 5).}
  \centering
  \label{tab:ablation_detailed}
  \vspace{.05in}

\begin{tabular}{lCCCCC}
  \toprule
  
  Method & \text{\#P}\downarrow &\text{CD}\downarrow & \text{PSNR}\uparrow & 
  \text{SSIM}\uparrow & \text{LPIPS}\downarrow\\
  \midrule

  Complete model & \cellcolor{mygreen!20} 4.60 \pm 0.23 &
  \bf 3.63 \pm 0.23 &\underline{20.5 \pm 0.2} &\underline{73.5 \pm 0.6} &\underline{23.9 \pm 
  0.5} \\

  \enspace w/o $\lpars$ &\cellcolor{red!25} 8.86 \pm 0.27 &
  \underline{3.65 \pm 0.22} &\bf 20.6 \pm 0.1 &\bf 73.7 \pm 0.4 &\bf 23.2 \pm 0.4 \\

  \enspace w/o $\lover$ &\cellcolor{mygreen!20} 4.38 \pm 0.19 &
  3.80 \pm 0.30 & 20.4 \pm 0.3 & 73.2 \pm 0.7 & 24.1 \pm 0.7 \\

  \enspace w/o curriculum &\cellcolor{mygreen!20} 4.66 \pm 0.30 &
  3.99 \pm 0.17 & 20.4 \pm 0.2 & 72.7 \pm 0.5 & 24.5 \pm 0.4 \\

  \enspace w/o noise in $\alpha_{1:K}$ &\cellcolor{mygreen!20} \underline{3.60 \pm 0.21} &
  4.13 \pm 0.28 & 20.0 \pm 0.2 & 72.0 \pm 0.6 & 25.6 \pm 0.6 \\

  \enspace w/o $\ltv$ &\cellcolor{mygreen!20} 4.04 \pm 0.18 &
  4.58 \pm 0.42 & 19.7 \pm 0.3 & 70.8 \pm 1.3 & 26.5 \pm 1.2  \\
  
  \enspace w/o $\lperc$ &\cellcolor{mygreen!20} \bf 3.22 \pm 0.17 &
  4.80 \pm 0.20 & 19.7 \pm 0.1 & 72.7 \pm 0.3 & 40.0 \pm 0.4  \\

  \bottomrule
  \end{tabular}
\end{table}

\paragraph{Ablation study.}
In~\Cref{tab:ablation_detailed}, we assess the key components of our model by removing one 
component at a time and computing the performance averaged over the 10 DTU scenes. We report 
the final number of primitives, Chamfer distance and rendering metrics.  We highlight the 
varying number of primitives in green (smaller than 5) and red (larger than 5). Results are 
averaged over five runs, we report the means and standard deviations. Overall, each component 
except $\lpars$ consistently improves the quality of the 3D reconstruction and the 
renderings.  $\lpars$ successfully limits the number of primitives (and thus, primitive 
duplication and over-decomposition) at a very small quality cost.

\paragraph{Influence of \textnormal{$K$} and \textnormal{$\wpars$}.} In~\Cref{tab:analysis}, 
we evaluate the impact of two key hyperparameters of our approach, namely the maximum number 
of primitives $K$ and the weight of the parsimony regularization $\wpars$. Results are averaged
over the 10 DTU scenes for 5 random seeds. First, we can observe that increasing $K$ slightly 
improves 
the reconstruction and rendering performances at the cost of a higher effective number of 
primitives. Second, these results show that $\wpars$ directly influences the effective 
number of primitives found. When $\wpars = 0.1$, this strong regularization limits the 
reconstruction to roughly one primitive, which dramatically decreases the performances. When 
$\wpars$ is smaller, the effective number of primitives increases without significant 
improvements in the reconstruction quality.

\begin{table}

  \caption{\textbf{Effect of hyperparameters on DTU~\cite{jensen2014large}.} We evaluate 
  the influence of two key hyperparameters of our model: the maximum number of primitives $K$ 
\textbf{(left)} and the parsimony regularization $\wpars$ \textbf{(right)}.}
  \label{tab:analysis}
  \vspace{.05in}
  \begin{subtable}{0.55\linewidth}
    \centering
    \addtolength{\tabcolsep}{-2pt}
    \small
    \begin{tabular}{@{}lccccc@{}} \toprule
      Method & \fontsize{8pt}{8pt}\selectfont \#P$\downarrow$ &
      \fontsize{8pt}{8pt}\selectfont CD$\downarrow$ &
      \fontsize{8pt}{8pt}\selectfont PSNR$\uparrow$ &
      \fontsize{8pt}{8pt}\selectfont SSIM$\uparrow$ &
      \fontsize{8pt}{8pt}\selectfont LPIPS$\downarrow$\\
      \midrule
      $K=10$ (default) & 4.60 & 3.63 & 20.5 & 73.5 & 23.9\\
      $K = 25$ & 7.00 & 3.58 & 21.0 & 74.6 & 22.5\\
      $K = 50$ & 9.26 & 3.52 & 20.9 & 74.7 & 22.8\\
      \bottomrule
    \end{tabular}
  \end{subtable}\quad\enspace
  \begin{subtable}{0.4\linewidth}
    \centering
    \addtolength{\tabcolsep}{-2pt}
    \small
    \begin{tabular}{@{}lcc@{}} \toprule
      Method & \fontsize{8pt}{8pt}\selectfont \#P$\downarrow$ &
      \fontsize{8pt}{8pt}\selectfont CD$\downarrow$\\
      \midrule
      $\wpars = 0.001$ & 7.44 & 3.61\\
      $\wpars = 0.01$ (default) & 4.60 & 3.63\\
      $\wpars = 0.1$ & 1.30 & 6.88\\
      \bottomrule
    \end{tabular}
  \end{subtable}
\end{table}

\begin{figure}
  \centering
    \begin{subfigure}{0.18\textwidth}
    \centering
    \includegraphics[width=.98\linewidth]{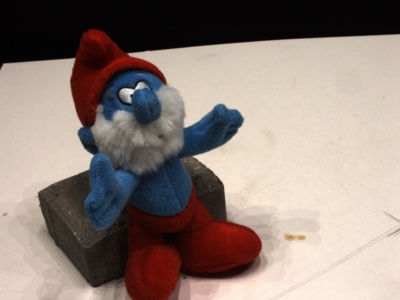}
    \includegraphics[width=.98\linewidth]{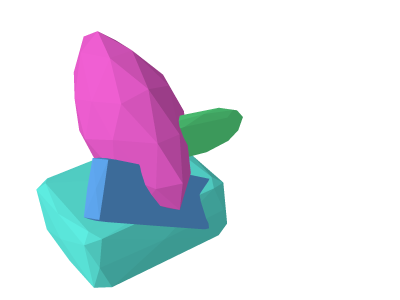}
    \caption{Missing parts}
    \label{fig:failure_a}
  \end{subfigure}\quad\enspace\,
  \begin{subfigure}{0.36\textwidth}
    \centering
    \includegraphics[width=.49\linewidth]{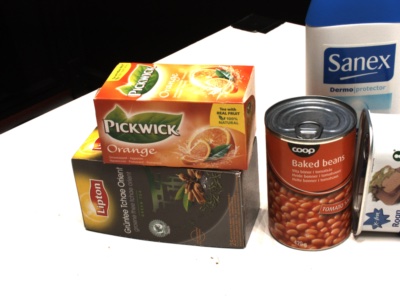}
    \includegraphics[width=0.49\linewidth]{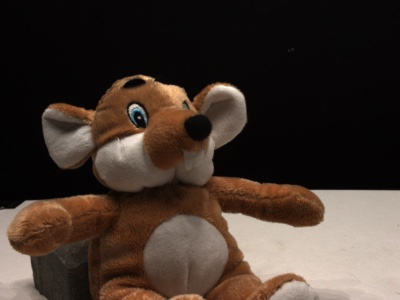}
    \includegraphics[width=.49\linewidth]{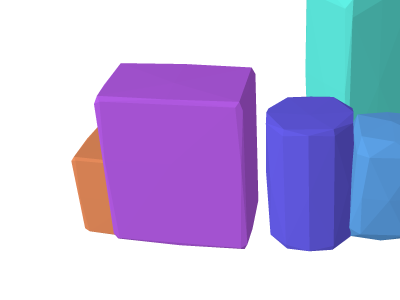}
    \includegraphics[width=0.49\linewidth]{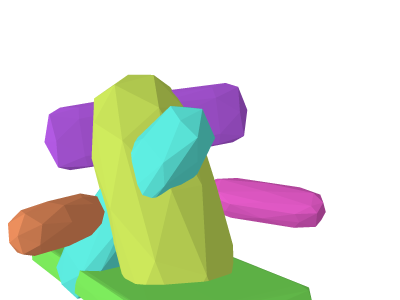}
    \caption{Unnatural decomposition}
    \label{fig:failure_b}
  \end{subfigure}\quad\enspace\,
  \begin{subfigure}{0.36\textwidth}
    \centering
    \includegraphics[width=.49\linewidth]{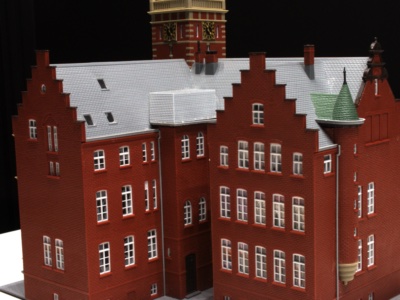}
    \includegraphics[width=.49\linewidth]{failures/house_inp.jpg}
    \includegraphics[width=.49\linewidth]{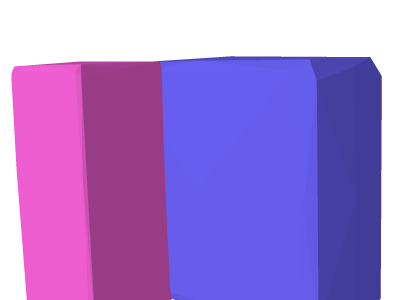}
    \includegraphics[width=.49\linewidth]{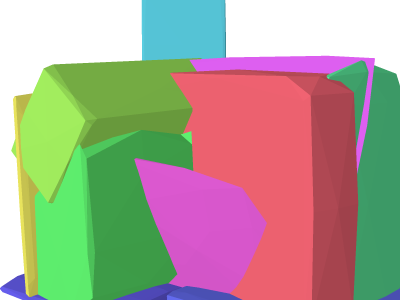}
    \caption{Parsimony/fidelity trade-off}
    \label{fig:failure_c}
  \end{subfigure}

  \caption{\textbf{Failure cases.} We show typical failure cases of our approach. All models 
  are optimized with $K=10$ except the rightmost model which is optimized with $K=50$.
  See text for details.}
  \label{fig:failures}
\end{figure}

\paragraph{Limitations and failure cases.} In~\Cref{fig:failures}, we show typical failure 
cases of our approach. First, for a random run, we may observe bad solutions where parts
of the geometry are not reconstructed (\Cref{fig:failure_a}). This is mainly caused 
by the absence of primitives in this region at initialization and our automatic selection 
among multiple runs alleviates the issue, yet this solution is computationally costly. 
Note that we also tried to apply a Gaussian kernel to blur the image and propagate
gradients farther, but it had little effect. Second, our reconstructions can yield 
unnatural decompositions as illustrated in~\Cref{fig:failure_b}, where tea boxes are wrongly 
split or a single primitive is modeling the bear nose and the rock behind. Finally, 
in~\Cref{fig:failure_c}, we show that increasing $K$ from 10 (left) to 50 (right) allows
us to trade-off parsimony for reconstruction fidelity. However, while this provides a form of 
control over the decomposition granularity, the ideal decomposition in this particular case
does not seem to be found: the former seems to slightly under-decompose the scene while 
the latter seems to over-decompose it.


\section{Conclusion}

\looseness=-1 We present an end-to-end approach that successfully computes a primitive-based 
3D reconstruction given a set of calibrated images. We show its applicability and robustness 
through various benchmarks, where our approach obtains better performances than methods 
leveraging 3D data.  We believe our work could be an important step towards more 
interpretable multi-view modeling.

\begin{ack}
  We thank Cyrus Vachha for help on the physics-based simulations; Antoine Guédon, Romain 
  Loiseau for visualization insights; Fran\c{c}ois Darmon, Romain Loiseau, Elliot Vincent for 
  manuscript feedback.  This work was supported in part by ANR project EnHerit 
  ANR-17-CE23-0008, gifts from Adobe and HPC resources from GENCI-IDRIS (2022-AD011011697R2, 
  2022-AD011013538). MA was supported by ERC project DISCOVER funded by the European Union’s 
  HorizonEurope Research and Innovation programme under grant agreement No. 101076028. Views 
  and opinions expressed are however those of the authors only and do not necessarily reflect 
  those of the European Union. Neither the European Union nor the granting authority can be 
  held responsible for them.

\end{ack}

{\small
\bibliographystyle{abbrv}
\bibliography{references}
}

\clearpage
\appendix

{\Large\bf\centering Supplementary Material for
  Differentiable Blocks World:\\ Qualitative 3D Decomposition by Rendering 
Primitives\\[1.5em]}

In this supplementary document, we provide additional results (\Cref{sec:extrares}),
details on the DTU benchmark (\Cref{sec:dataset}) as well as implementation 
details~(\Cref{sec:implem}), including design and optimization choices.

\section{Additional Results}\label{sec:extrares}

\paragraph{Videos for view synthesis, physical simulations and amodal completion.}
We present additional results in the form of videos at our project webpage: 
\href{https://www.tmonnier.com/DBW}{\texttt{www.tmonnier.com/DBW}}. Videos are separated in 
different sections depending on the experiment type. First, we provide view synthesis videos 
(rendered using a circular camera path), further outlining the quality of both our renderings 
and our primitive-based 3D reconstruction. Second, we include videos for physics-based 
simulations.  Such simulations were produced through Blender by simply uploading our output 
primitive meshes. Note that for modeling primitive-specific motions in Blender (\eg, in our 
teaser figure), primitives should not overlap at all, thus requiring a small preprocessing 
step to slightly move the primitives for a clear separation.  Because each primitive is its 
own mesh, this operation is easily performed within Blender.  Finally, we provide video
results where we perform scene editing and compare our amodal view synthesis results to the ones 
of Nerfacto introduced in Nerfstudio~\cite{tancik2023nerfstudio}. Models for amodal synthesis 
are optimized on a homemade indoor scene built from a forward-facing capture only. We use 
Nerfstudio for data processing and data convention.

%

\begin{wraptable}{rt}{0.5\linewidth}
  \vspace{-1.5em}
  \centering
  \small
  \addtolength{\tabcolsep}{-2.9pt}
  \caption{\textbf{PSNR comparison on DTU.}}
  \vspace{-.05in}

  \begin{tabular}{@{}lccccccc@{}}
  \toprule
  
  Method & \tt S24 &\tt S40 &\tt S55 &\tt S63 & \tt S83 &\tt S105 & Mean \\
  \midrule

  NeRF & 26.2 & 26.8 & 27.6&32.0 & 32.8& 32.1 & 29.6 \\
  Ours & 19.1 & 21.8 & 22.6 & 23.4 & 22.3 & 20.8 & 21.7\\
  \bottomrule
  \end{tabular}

  \label{tab:nerf}
\end{wraptable}

\paragraph{Rendering comparison with SOTA MVS.} For completeness, we provide the rendering 
performances of NeRF~\cite{mildenhall2020nerf}, a SOTA MVS method that does not predict
multiple parts. In~\Cref{tab:nerf}, we compare PSNR for our 
approach and NeRF using the results reported in~\cite{yariv2021volume} on the intersected 
set of 6 DTU scenes.

\section{DTU Benchmark}\label{sec:dataset}

\begin{figure}
  \centering
  \addtolength{\tabcolsep}{-5pt}
  \renewcommand{\arraystretch}{1.05}
  \resizebox{\linewidth}{!}{%
  \begin{tabular}{@{}ccccccc@{}}
    \small Input (subset) &\small View 1 &\small View 2 &
    \small View 3 & \small View 4 & \small View 5 \\
    
    \includegraphics[width=0.158\linewidth]{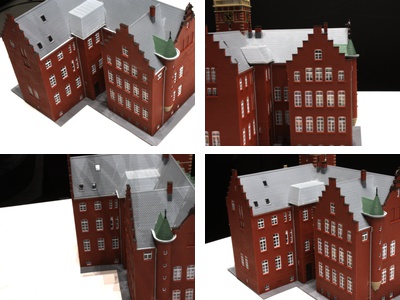} & 
    \includegraphics[width=0.158\linewidth]{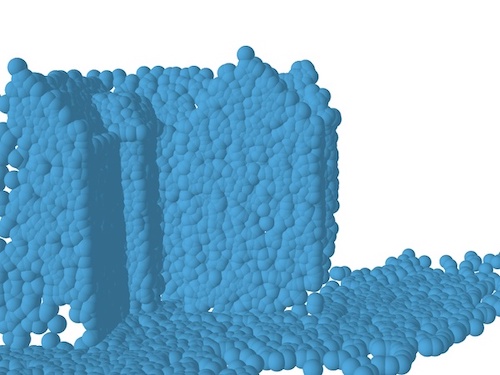} & 
    \includegraphics[width=0.158\linewidth]{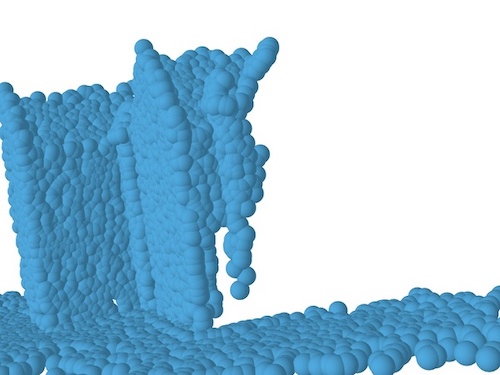} & 
    \includegraphics[width=0.158\linewidth]{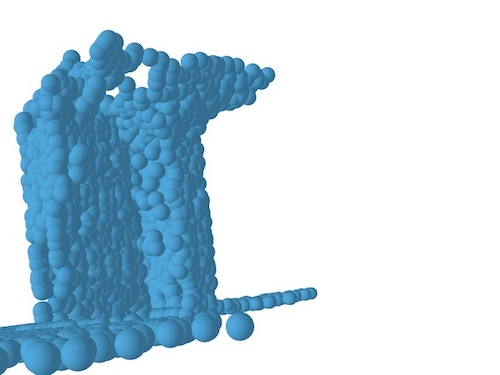} & 
    \includegraphics[width=0.158\linewidth]{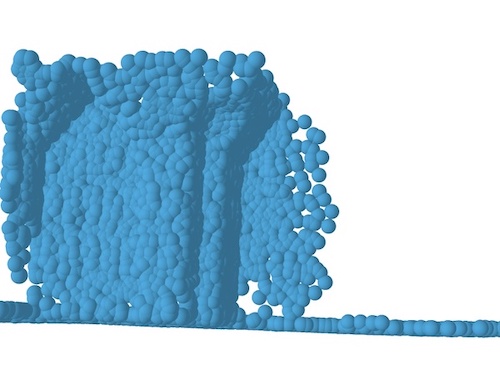} & 
    \includegraphics[width=0.158\linewidth]{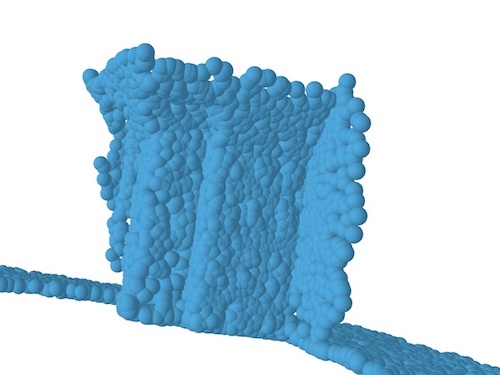} \\

    \includegraphics[width=0.158\linewidth]{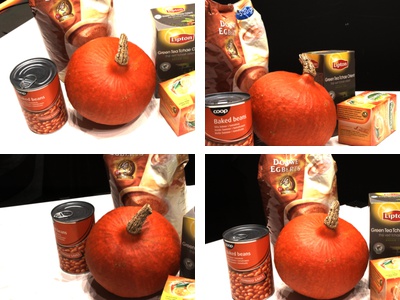} & 
    \includegraphics[width=0.158\linewidth]{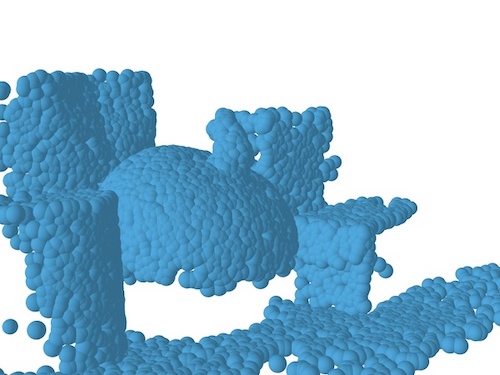} & 
    \includegraphics[width=0.158\linewidth]{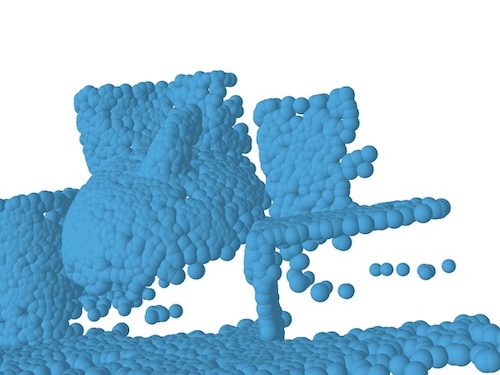} & 
    \includegraphics[width=0.158\linewidth]{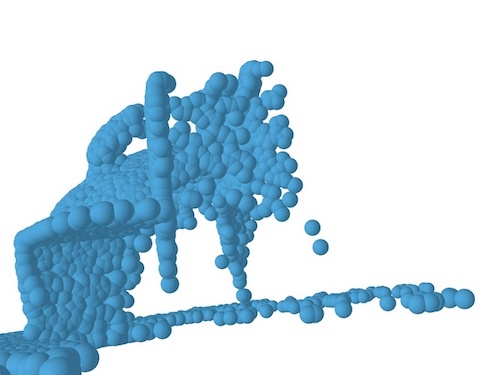} & 
    \includegraphics[width=0.158\linewidth]{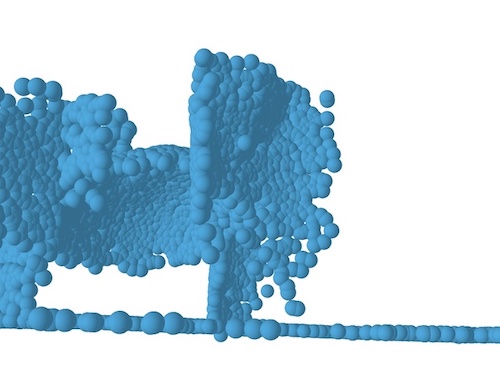} & 
    \includegraphics[width=0.158\linewidth]{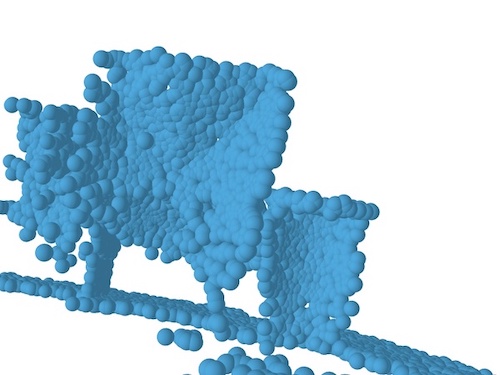} \\ 

    \includegraphics[width=0.158\linewidth]{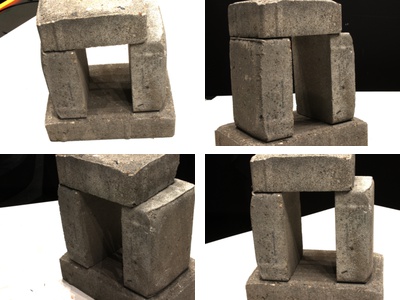} & 
    \includegraphics[width=0.158\linewidth]{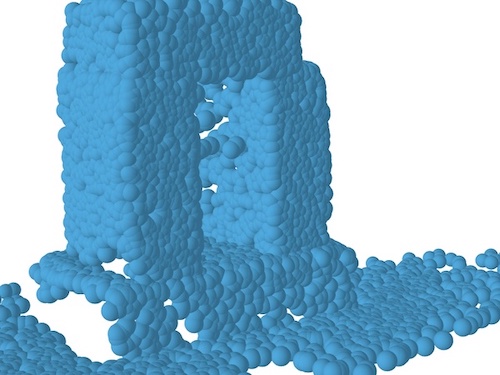} & 
    \includegraphics[width=0.158\linewidth]{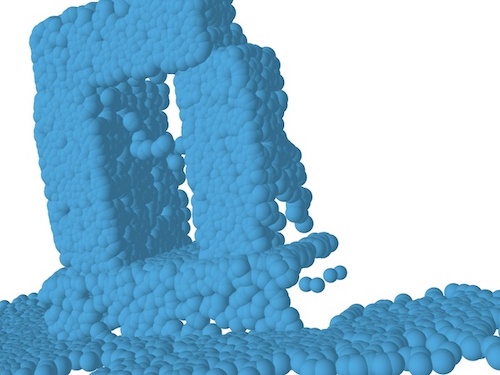} & 
    \includegraphics[width=0.158\linewidth]{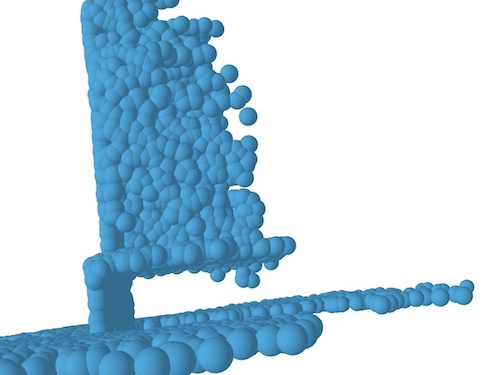} & 
    \includegraphics[width=0.158\linewidth]{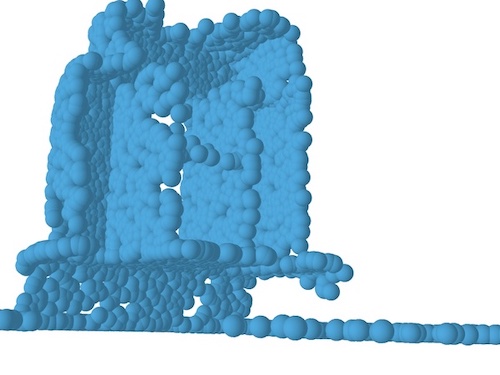} & 
    \includegraphics[width=0.158\linewidth]{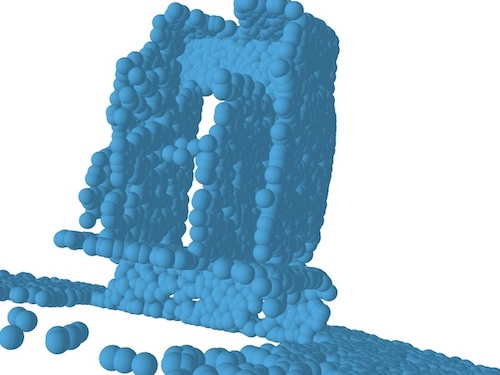} \\ 

    \includegraphics[width=0.158\linewidth]{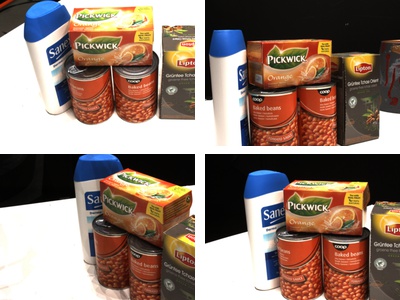} & 
    \includegraphics[width=0.158\linewidth]{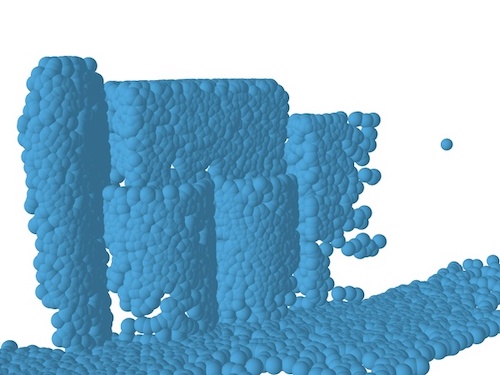} & 
    \includegraphics[width=0.158\linewidth]{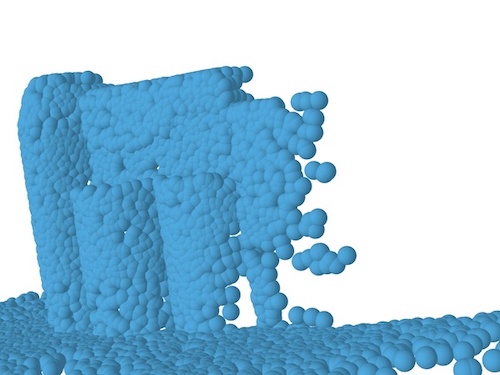} & 
    \includegraphics[width=0.158\linewidth]{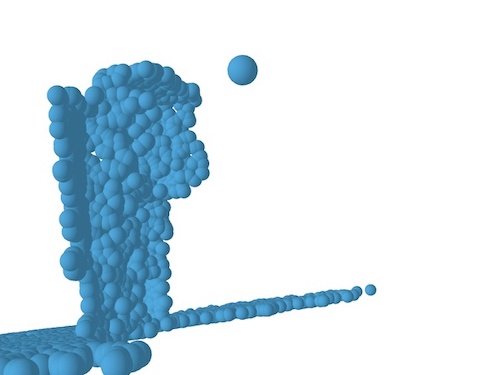} & 
    \includegraphics[width=0.158\linewidth]{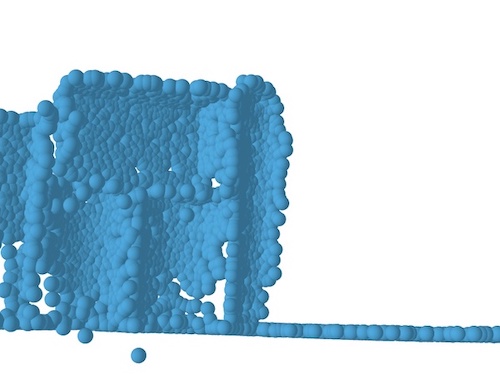} & 
    \includegraphics[width=0.158\linewidth]{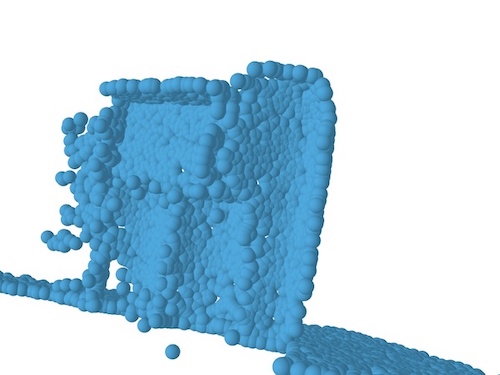} \\

    \includegraphics[width=0.158\linewidth]{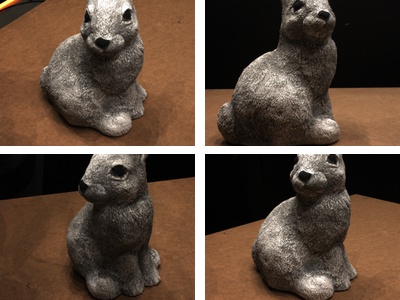} & 
    \includegraphics[width=0.158\linewidth]{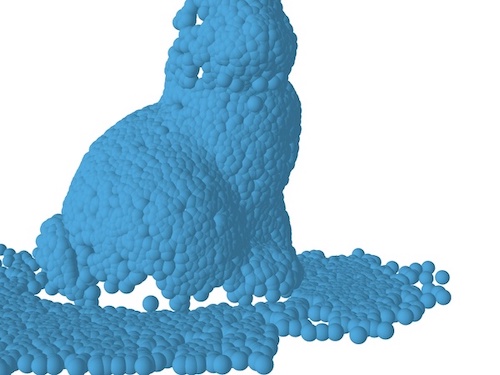} & 
    \includegraphics[width=0.158\linewidth]{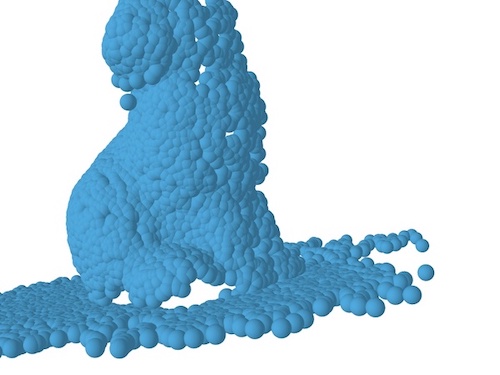} & 
    \includegraphics[width=0.158\linewidth]{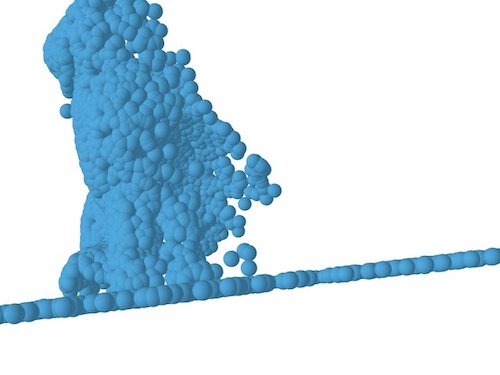} & 
    \includegraphics[width=0.158\linewidth]{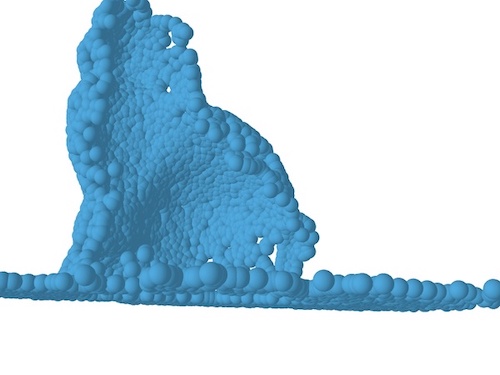} & 
    \includegraphics[width=0.158\linewidth]{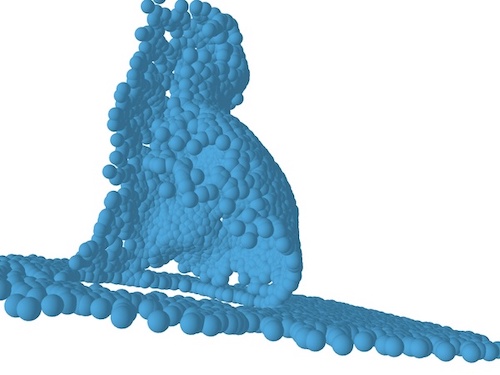} \\

    \includegraphics[width=0.158\linewidth]{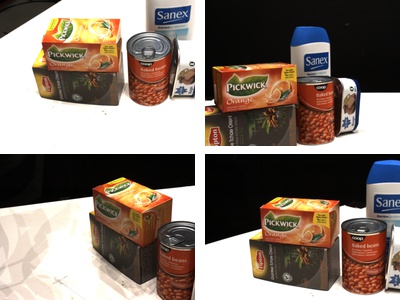} & 
    \includegraphics[width=0.158\linewidth]{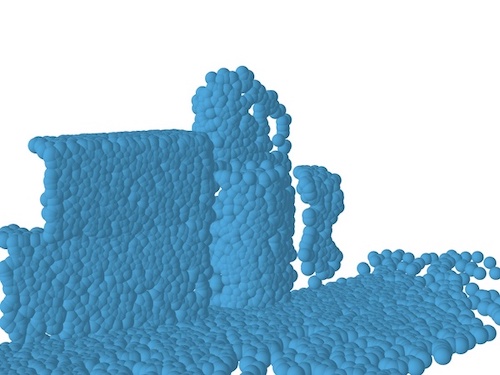} & 
    \includegraphics[width=0.158\linewidth]{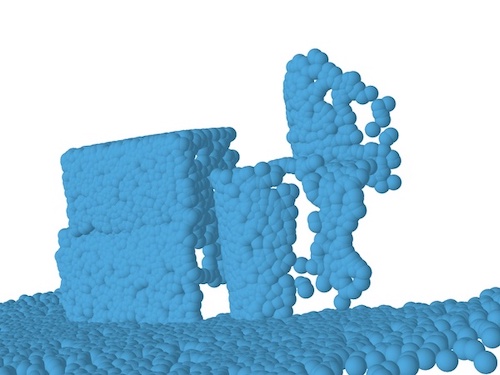} & 
    \includegraphics[width=0.158\linewidth]{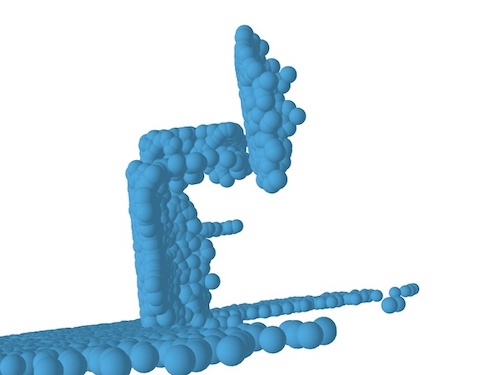} & 
    \includegraphics[width=0.158\linewidth]{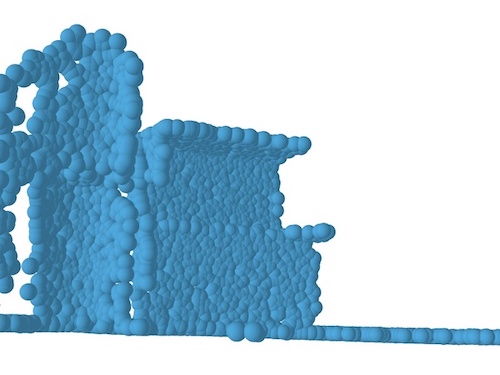} & 
    \includegraphics[width=0.158\linewidth]{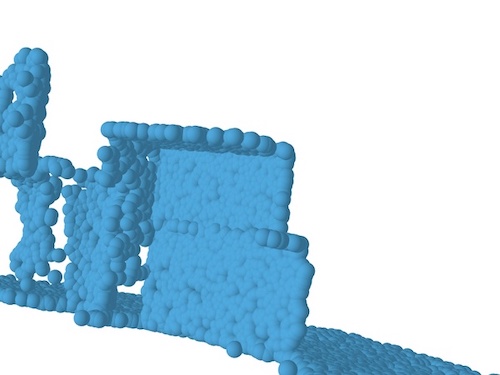} \\

    \includegraphics[width=0.158\linewidth]{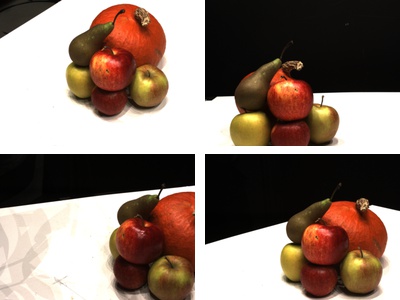} & 
    \includegraphics[width=0.158\linewidth]{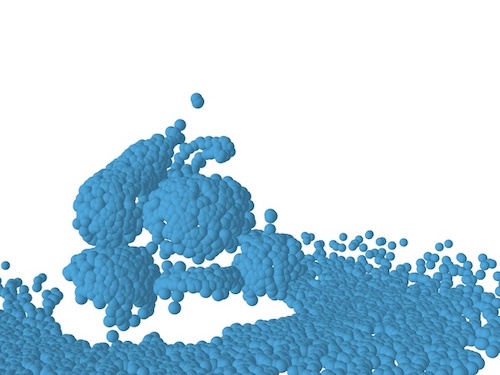} & 
    \includegraphics[width=0.158\linewidth]{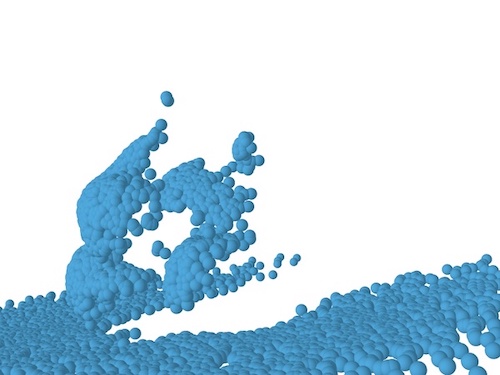} & 
    \includegraphics[width=0.158\linewidth]{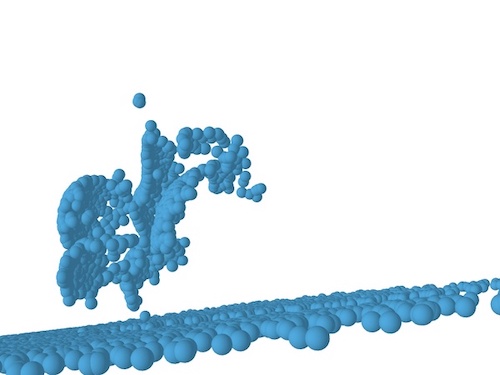} & 
    \includegraphics[width=0.158\linewidth]{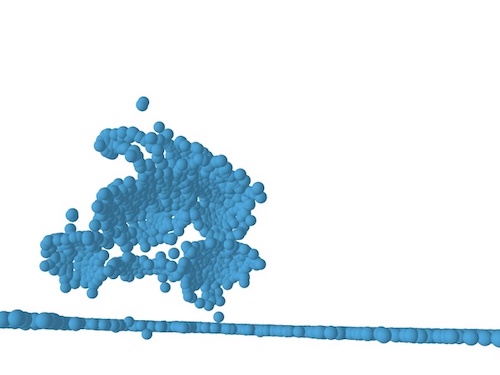} & 
    \includegraphics[width=0.158\linewidth]{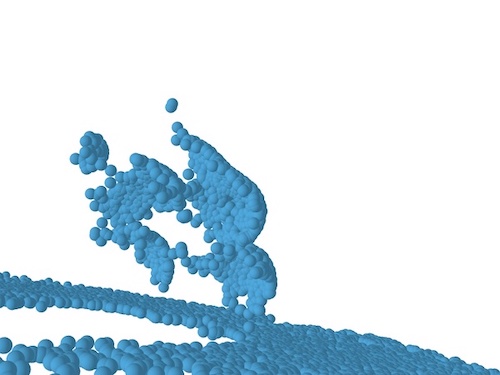} \\

    \includegraphics[width=0.158\linewidth]{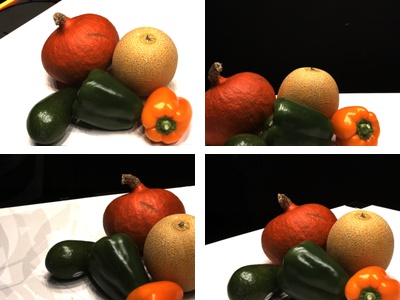} & 
    \includegraphics[width=0.158\linewidth]{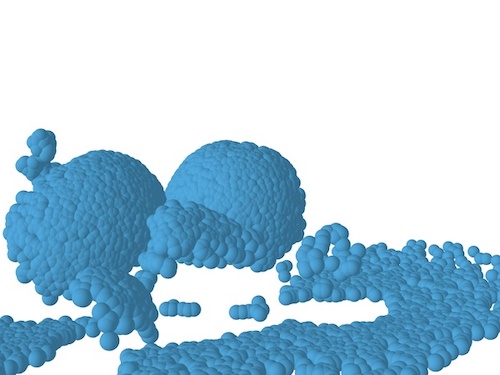} & 
    \includegraphics[width=0.158\linewidth]{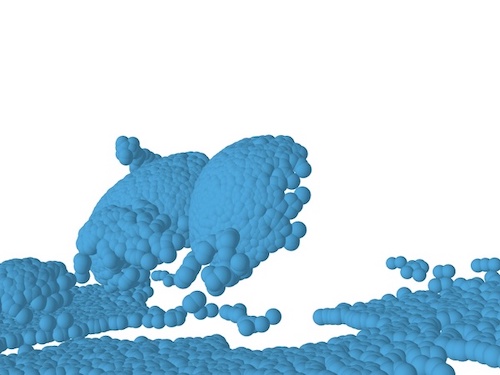} & 
    \includegraphics[width=0.158\linewidth]{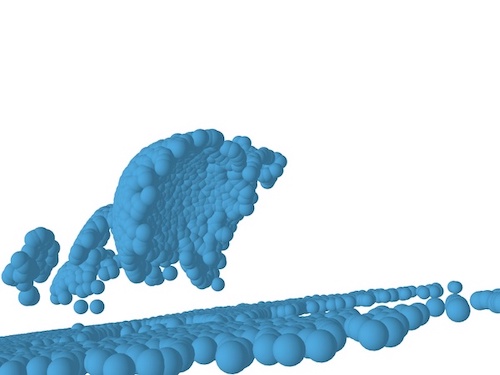} & 
    \includegraphics[width=0.158\linewidth]{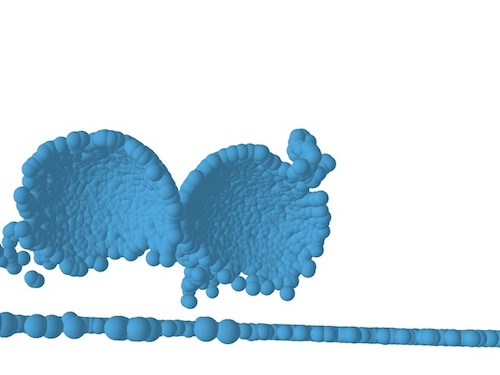} & 
    \includegraphics[width=0.158\linewidth]{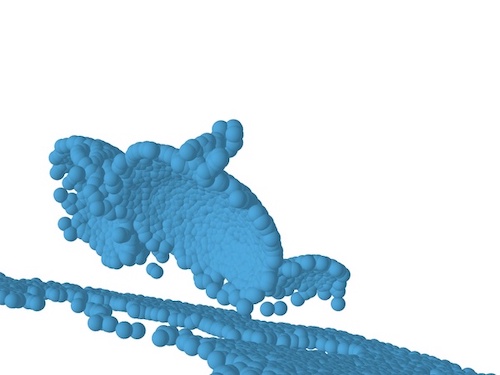} \\

    \includegraphics[width=0.158\linewidth]{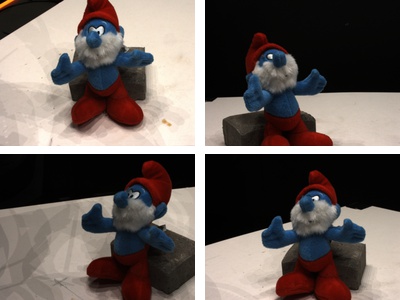} & 
    \includegraphics[width=0.158\linewidth]{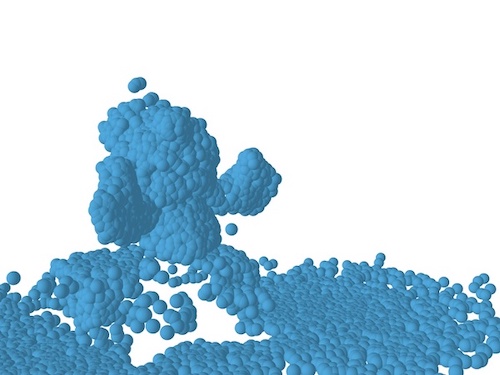} & 
    \includegraphics[width=0.158\linewidth]{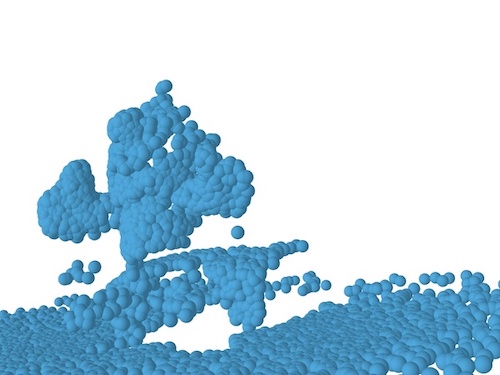} & 
    \includegraphics[width=0.158\linewidth]{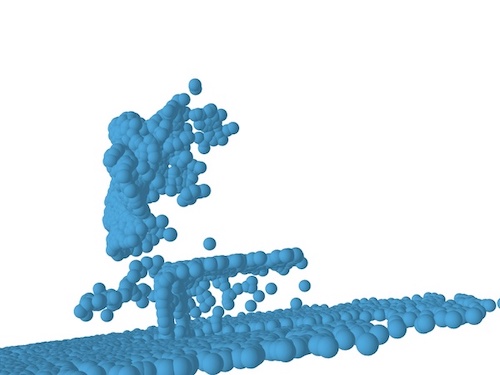} & 
    \includegraphics[width=0.158\linewidth]{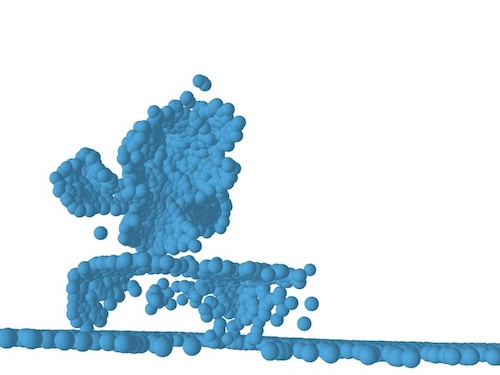} & 
    \includegraphics[width=0.158\linewidth]{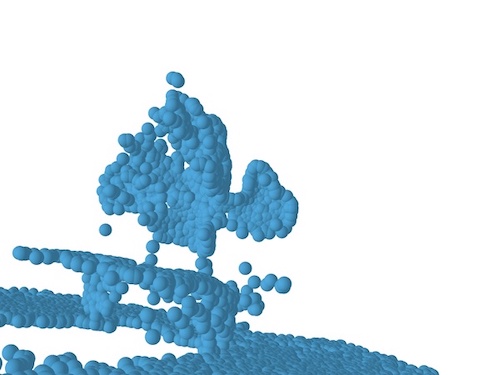} \\

    \includegraphics[width=0.158\linewidth]{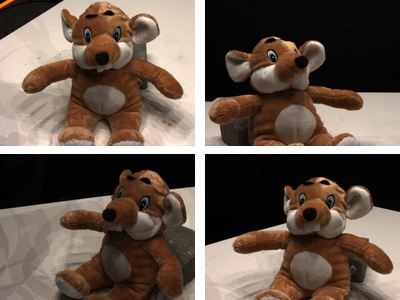} & 
    \includegraphics[width=0.158\linewidth]{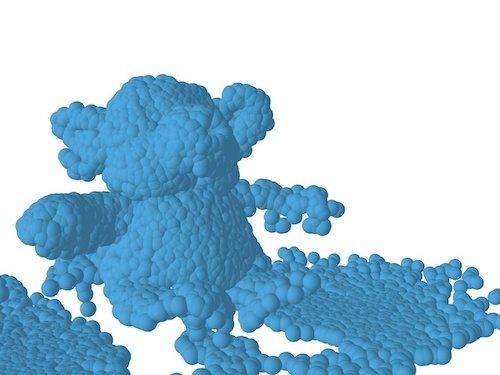} & 
    \includegraphics[width=0.158\linewidth]{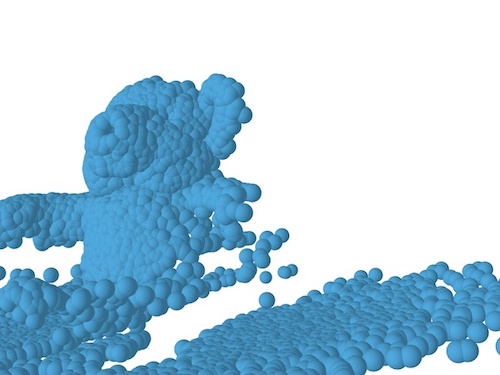} & 
    \includegraphics[width=0.158\linewidth]{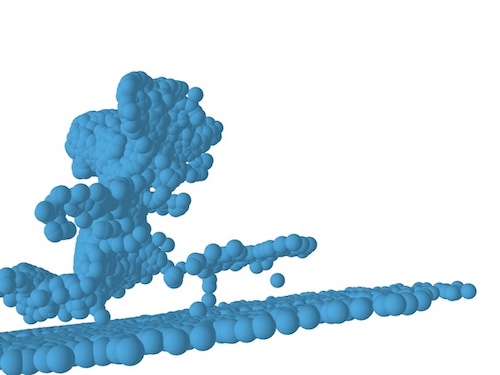} & 
    \includegraphics[width=0.158\linewidth]{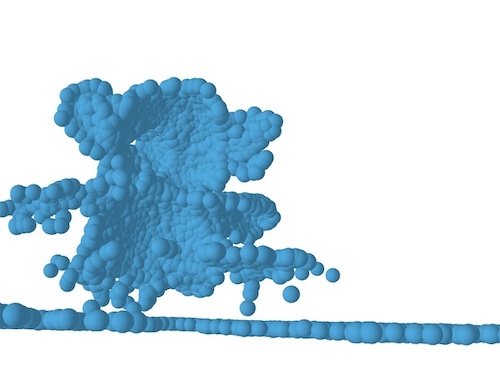} & 
    \includegraphics[width=0.158\linewidth]{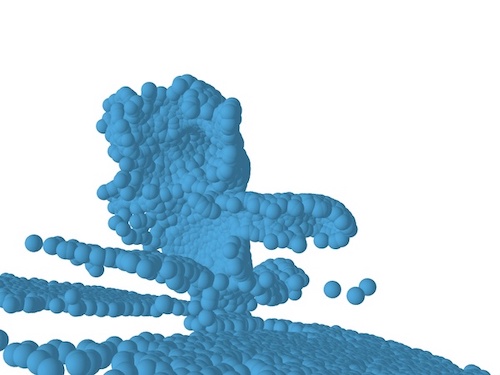} \\

  \end{tabular}
  }
  \caption{\textbf{DTU~\cite{jensen2014large} scenes with ground-truth.} We show a subset of 
  the input images as well as renderings of the GT point clouds.  From top to bottom, scenes 
are: \texttt{S24}, \texttt{S31}, \texttt{S40}, \texttt{S45}, \texttt{S55}, \texttt{S59}, 
\texttt{S63}, \texttt{S75}, \texttt{S83}, \texttt{S105}.}
\label{fig:dtu_gt}
\end{figure}

In~\Cref{fig:dtu_gt}, we show for each scene a subset of the input images as well as 
360$^\circ$ renderings of the GT point clouds obtained through a structured light scanner. To 
compute performances, we use a Python version of the official evaluation: 
\href{https://github.com/jzhangbs/DTUeval-python}{\texttt{https://github.com/jzhangbs/DTUeval-python}}.

\section{Implementation Details}\label{sec:implem}

\begin{figure}[t]
  \centering
  \begin{subfigure}{0.3\textwidth}
    \centering
    \includegraphics[width=0.48\linewidth]{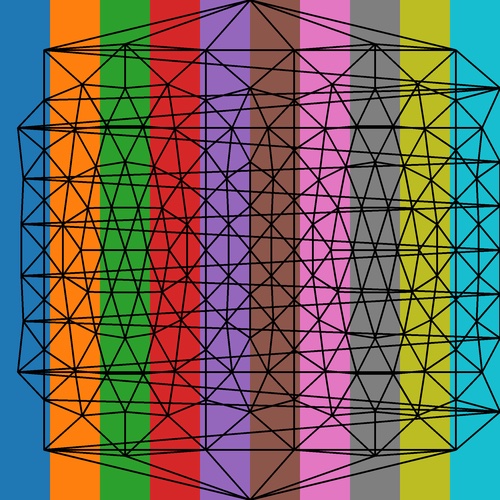}
    \includegraphics[width=0.48\linewidth]{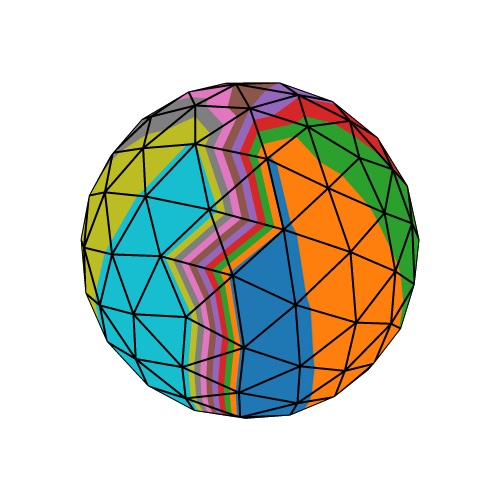}
    \caption{Raw spherical coordinates}
  \end{subfigure}
  \begin{subfigure}{0.34\textwidth}
    \centering
    \includegraphics[width=0.46\linewidth]{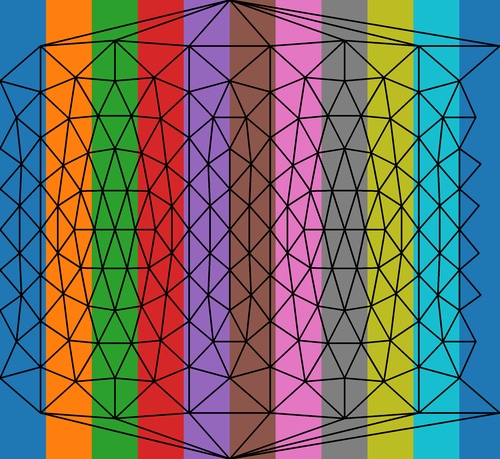}
    \includegraphics[width=0.4235\linewidth]{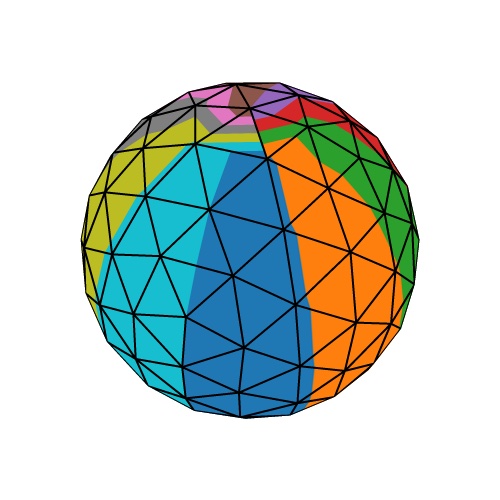}
    \caption{Fixing border discontinuities}
  \end{subfigure}
  \begin{subfigure}{0.34\textwidth}
    \centering
    \includegraphics[width=0.46\linewidth]{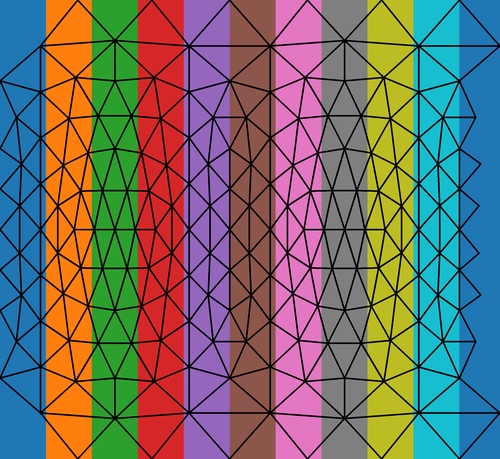}
    \includegraphics[width=0.4235\linewidth]{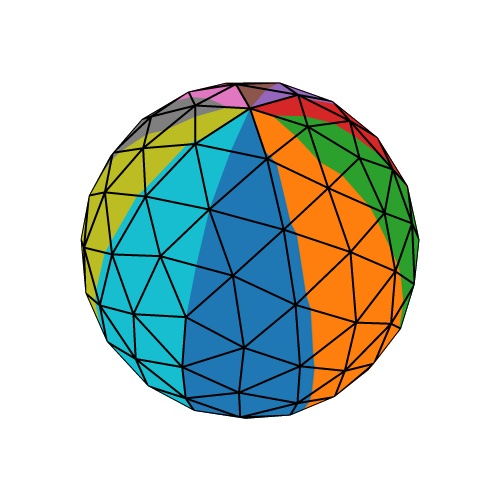}
    \caption{Fixing distortions at the poles}
  \end{subfigure}
  
  \caption{\textbf{Our icosphere UV-mapping.} We illustrate different UV parametrizations 
    using raw spherical coordinates \textbf{(a)} as well as our modified coordinates to fix 
    discontinuities \textbf{(b)} and to prevent distortions at the poles \textbf{(c)}.  For 
    each parametrization, we show the texture image with the face edges representing the 
  UV-mapping as well as a rendering example of the associated icosphere.}
\label{fig:uv_mapping}
\end{figure}

\paragraph{Icosphere and superquadric UV mapping.} We use spherical coordinates that we 
correct to build our texture mapping for the unit icosphere.~\Cref{fig:uv_mapping} shows our 
process with an example. Specifically, we retrieve for each vertex its spherical coordinates 
$\eta \in [-\frac{\pi}{2}, \frac{\pi}{2}]$ and $\omega \in [-\pi, \pi]$ which are linearly 
mapped to the UV space $[0, 1]^2$.  Because such parametrization presents discontinuities and 
strong triangle deformations at the poles, we perform two corrections.  First, we fix 
discontinuities by copying the border pixels involved (using a circular padding on the 
texture image) and introducing new 2D vertices such that triangles do not overlap anymore.  
Second, we avoid distorted triangles at the poles by creating for each triangle, a new 2D 
vertex positioned in the middle of the other two vertices. As detailed in the main paper, we 
derive a superquadric mesh from a unit icosphere in such a way that each vertex of the 
icosphere is continuously mapped to the superquadric vertex. As a result, the texture mapping 
defined for the icosphere is directly transferred to our superquadric meshes without any 
modification.

\paragraph{Design choices.} Except constants related to the world scale, orientation and 
position in the 3D space \wrt to the known cameras, all our experiments share the same design 
choices. Specifically, all the following design choices are defined for a canonical 3D scene 
assumed to be centered and mostly contained in the unit cube, with a y-axis orthogonal to the 
ground and pointing towards the sky. We roughly estimate the scene-specific constants related 
to the world scale and pose (through coarse visual comparisons or using the camera 
locations), and apply them to our final scene model to account for the camera conventions.

The background corresponds to a level-2 icosphere (320 faces), the ground plane is subdivided 
into 128 uniform faces (for visual purposes) and superquadric meshes are derived from level-1 
icospheres (80 faces). The scale for the background and the ground is set to 10. The ground 
is initialized perpendicular to the y-axis and positioned at $[0, -0.9, 0]$.The poses of our 
primitive blocks are initialized using a Gaussian distribution for the 3D translation and a 
random 6D vector for the rotation such that rotations are uniformly distributed on the unit 
sphere. We parametrize their scale with an exponential added to a minimum scale value of 0.2 
to prevent primitives from becoming too small.  These scales are initialized with a uniform 
distribution in $[0.5, 1.5]$ and multiplied by a constant block scale ratio of 0.25 to yield 
primitives smaller than the scene scale. The superquadric shape parameters are implemented 
with a sigmoid linearly mapped in $[0.1, 1.9]$ and are initialized at 1 (thus corresponding 
to a raw icosphere).  Transparency values are parametrized with a sigmoid and initialized at 
0.5.  All texture images have a size of $256 \times 256$, are parametrized using a sigmoid 
and are initialized with small Gaussian noises added to gray images.

\paragraph{Optimization details.} All our experiments share the same optimization details.  
We use Pytorch3D framework~\cite{raviAccelerating3DDeep2020} to build our custom 
differentiable rendering process and use the default hyperparameter $\sigma = 10^{-4}$.  Our 
model is optimized using Adam~\cite{kingma2015adam} with a batch size of 4 for roughly a 
total of 25k iterations. We use learning rates of 0.05 for the texture images and 0.005 for 
all other parameters, and divide them by 10 for the last 2k iterations. Following our 
curriculum learning process, we optimize the model for the first 10k iterations by 
downsampling all texture images by 8. Then, we optimize using the full texture resolution 
during the next 10k iterations. Finally, to further increase the rendering quality, we 
threshold the transparency values at 0.5 to make them binary, remove regularization terms 
related to transparencies (\ie, $\lpars$ and $\lover$), divide the weights for the other 
terms $\lperc$ and $\ltv$ by 10, decrease the smoothness rendering parameter $\sigma$ to 
$5\times10^{-6}$ and finetune our model for the final 5k iterations. In particular, this 
allows the model to output textures that are not darken by non-binary transparencies. During 
the optimization, we systematically kill blocks reaching a transparency lower than 0.01 and 
at inference, we only show blocks with a transparency greater than 0.5. Similar 
to~\cite{paschalidou2021neural}, we use $\lambda = 1.95$ and $\temp = 0.005$ in our overlap 
penalization.

\paragraph{Computational cost.} Optimizing our model on a scene roughly takes 4 hours on a 
single NVIDIA RTX 2080 Ti GPU. Since MBF~\cite{ramamonjisoa2022monteboxfinder} and 
EMS~\cite{liu2022robust} directly operate on the 3D point cloud without computing textures, 
they are much faster and compute primitives in a couple of minutes. To get comparable timings 
however, we have to account for a method that computes 3D point clouds from the calibrated 
images, which is typically longer depending on the method. For example, we report MBF and EMS 
results using the mesh extracted from NeUS~\cite{wang2021neus}, which typically takes 14 
hours to converge on a single DTU scene.

\end{document}